\pdfoutput=1

\documentclass[11pt]{article}

\usepackage[preprint]{acl_mod}

\usepackage{times}
\usepackage{latexsym}

\usepackage{enumitem}

\usepackage[utf8]{inputenc}

\usepackage{placeins}

\usepackage[T1]{fontenc}

\usepackage{pmboxdraw}

\usepackage[figuresright]{rotating}

\usepackage{listings}
\usepackage{dblfloatfix}

\usepackage{microtype}

\usepackage{inconsolata}

\usepackage{graphicx}

\usepackage{geometry}
\usepackage{amsmath}
\usepackage{array}
\usepackage{adjustbox}

\usepackage{tcolorbox}

\usepackage{xcolor}
\definecolor{red}{rgb}{0.99, 0.02, 0.02}
\NewDocumentCommand{\heng}
{ mO{} }{\textcolor{red}{\textsuperscript{\textit{Heng}}\textsf{\textbf{\small[#1]}}}}

\NewDocumentCommand{\carl}
{ mO{} }{\textcolor{blue}{\textsuperscript{\textit{Carl}}\textsf{\textbf{\small[#1]}}}}

\newcolumntype{H}{>{\setbox0=\hbox\bgroup}c<{\egroup}@{}}
\usepackage{booktabs}

\newcommand{\name}{\textsc{MolCap-Arena}}

\title{MolCap-Arena: A Comprehensive Captioning Benchmark on Language-Enhanced Molecular Property Prediction}

\author{Carl Edwards, Ziqing Lu, Ehsan Hajiramezanali, \\\textbf{Tommaso Biancalani, Heng Ji, and Gabriele Scalia} \\
  \texttt{\{cne2, hengji\}@illinois.edu}, \texttt{\{luz21,hajiramm,biancalt,scalia.gabriele\}@gene.com}
}

\usepackage{multicol}

\begin{document}

\maketitle
\vspace{-1cm}
\begin{abstract}
  Bridging biomolecular modeling with natural language information, particularly through large language models (LLMs), has recently emerged as a promising interdisciplinary research area. LLMs, having been trained on large corpora of scientific documents, demonstrate significant potential in understanding and reasoning about biomolecules by providing enriched contextual and domain knowledge. However, the extent to which LLM-driven insights can improve performance on complex predictive tasks (e.g., toxicity) remains unclear. Further, the extent to which \textit{relevant} knowledge can be extracted from LLMs also remains unknown. In this study, we present Molecule Caption Arena: the first comprehensive benchmark of LLM-augmented molecular property prediction. We evaluate over twenty LLMs, including both general-purpose and domain-specific molecule captioners, across diverse prediction tasks. To this goal, we introduce a novel, battle-based rating system. Our findings confirm the ability of LLM-extracted knowledge to enhance state-of-the-art molecular representations, with notable model-, prompt-, and dataset-specific variations. \\
  \begin{center}
\vspace{-1.cm}
      Code, resources, and data are available at \href{https://github.com/Genentech/molcap-arena}{github.com/Genentech/molcap-arena}.
  \end{center}
\end{abstract}

\begin{figure*}[h!]
\vspace{-.4cm}
  \centering
  \includegraphics[width=1.0\textwidth] {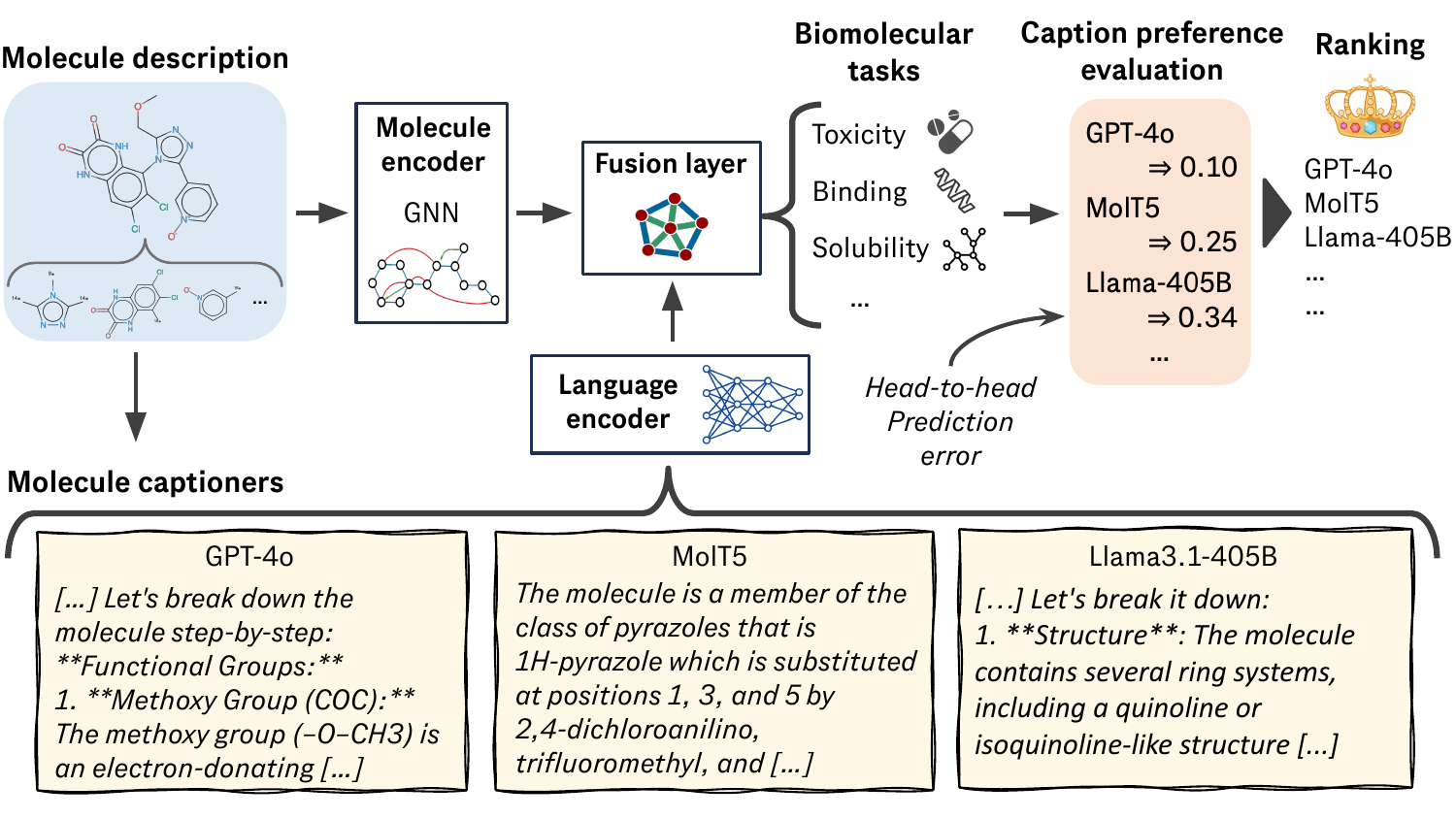}
\label{fig:overview}
\vspace{-.7cm}
\caption{Overview of \name{}}
\vspace{-.4cm}
\end{figure*}

\begin{multicols}{2}
\section{Introduction}
\vspace{-.4cm}
Biological research critically relies on molecular understanding to advance drug discovery. Machine learning methods have gained significant attention in recent years, accelerating and improving diverse tasks such as molecular design~\cite{pang2023deep}, reaction mining~\cite{reactionminer2023}
experiment procedure design~\cite{actionie2024}, 
reaction prediction~\cite{wang2023retrosynthesis}, and quantitative structure-activity modeling (QSAR)~\cite{heid2023chemprop}. In particular, molecular property prediction remains a cornerstone in computational drug discovery~\cite{tropsha2024integrating}, enabling diverse critical tasks such as bioactivity prediction, toxicity characterization, de-novo design, and virtual screening, among others. 

More recently, the intersection of biomolecular modeling and natural language processing has emerged as a promising interdisciplinary frontier in molecular discovery~\cite{zhang2024scientific,pei2024leveraging}. Indeed, natural language can provide tools to enrich, describe, and interact with molecular representations in a multifaceted and information-rich way.
This convergence is particularly evident in the application of large language models (LLMs), where their ability to capture, understand, and reason about biomolecules offers new opportunities to enhance chemistry tasks~\cite{edwards2024tutorial}.

Particularly, \emph{molecule captioning} has emerged as a prominent task, providing a natural interface to bridge biomolecules with human understanding. However, the evaluation of this task has proven challenging due to its ambiguous nature, and there has been limited investigation into domain-specific information captured by the generated captions.

The primary goal of this work is to establish a benchmark for measuring the usefulness of the information extracted from LLMs captioners in real-world molecular modeling tasks. Our primary contributions are as follows:
\begin{itemize}%
    \item We release a large-scale benchmark of molecule captioners aimed at enhancing real-world tasks such as toxicity prediction, moving beyond traditional n-gram metrics. 
    \item We provide molecule captions and a unified codebase including dozens of models, facilitating future comparisons in the field.
    \item By analyzing existing models, including both domain-specific captioners and state-of-the-art LLMs, across hundreds of experiments, we provide insights into the ability of captions to capture relevant biochemical knowledge.
    
\end{itemize}

\section{Why do we need a benchmark?}

With the introduction of molecule-text multimodality \cite{edwards2021text2mol, zeng2022deep}, considerable interest has arisen in the problem of molecule captioning \cite{su2022molecular, edwards2022translation, 
liu-etal-2023-molxpt}. These captions are typically evaluated using traditional machine translation metrics such as BLEU and ROUGE scores. Additionally, metrics have been introduced which use similarity scores between representations \cite{edwards2022translation} or which extract specific text properties \cite{edwards2024_lpm24}. However, these metrics still fundamentally rely on ground truth captions and, therefore, are intrinsically limited by the fact that molecules can be described with far more linguistic variety than traditional image captions \cite{edwards2024tutorial}. For example, a caption on paclitaxel might describe the different synthetic routes for its synthesis, but it could alternatively describe its mechanism of action for treating cancer, or even its pharmacokinetics and tissue distribution within the body. 

To address these limitations, we introduce \name{}. Our objective is to target an impactful, real-world problem: \textit{enchancing} molecular property prediction models.
In particular, we argue that the informativeness of a caption (and, in turn, of an LLM captioner) is closely tied to its ability to capture properties of interest, therefore connecting LLM-driven insights to tangible predictive tasks. 

This supports three key goals: 1) improving existing molecular property prediction models by leveraging LLM-extracted knowledge, 2) grounding the evaluation of molecule captioners into their usefulness for real-world applications, and 3) enhancing explainability in molecule property prediction tasks by providing captions that aid human scientists in interpreting model predictions. 

We note that very recent work has started exploring using LLMs to augment property prediction models~\cite{guo2024moltailor, GLaD2024,xu2024llm}. However, these studies generally rely on a single caption source with different integration strategies, making it hard to isolate the impact of the captions.
In contrast, \name{} offers a comprehensive benchmark across multiple captioners and tasks, with a standardized evaluation protocol for evaluating their contribution. %
This offers insights into the ability of LLMs to enhance molecular representations, providing a critical resource for further research into this promising area.

\section{Ranking Captioners}

\subsection{Augmenting molecular representations with caption information}
\label{ssec:joint-model}

Given that we are interested in evaluating the \emph{additional knowledge} provided by different captions, we propose to augment a molecule-only model with caption information. We consider a simple yet flexible modeling paradigm based on late fusion of the molecule and text information. 
Each dataset is split into \emph{training}, \emph{preference}, \emph{validation} and \emph{test} splits,
following a scaffold splitting strategy to simulate chemical distribution shifts~\cite{wu2018moleculenet}.

We first train a GNN on the molecular structures. This model is based on graph isomorphism networks with edge features~\cite{xu2018how}, such that $y^{\mathrm{G}}_{\text{task}} = \text{GNN}(g(mol))$, where $g$ denotes the graph representation of the molecule $mol$, and $\text{task}$ refers to the specific molecular property (e.g., toxicity, bioactivity, binding, etc.). This model provides a parameter-efficient yet robust baseline of the information that can be extracted solely from the molecular structure. Next, we train a text encoder on the captions using a specialized BERT model, bioLinkBERT-base \cite{yasunaga2022linkbert}, such that  $y^{\mathrm{C}}_{\text{task}} = \text{BERT}(c(mol))$, where $c$ is a captioner describing the molecule $mol$. Both the GNN and the BERT model are trained on the same training set, using the validation set for early stopping.
Finally, caption embeddings $x_c$ and molecule embeddings $x_m$ are extracted from BERT and the GNN, respectively, and concatenated to obtain $x_{concat} = x_c \oplus x_m$. 
A support vector machine model~\cite{scikit-learn} is trained on such joint embeddings, such that $y^{\mathrm{J}}_{\text{task}} = \text{SVM}(x_{concat})$, using the preference set as training data.

We observe how this strategy mitigates overfitting a source of information, as each modality is independently used to train a specialized architecture first, and the joint embedding is used to train a shallow model. This allows us to directly test the impact of a given caption on predictive performance. %
Similar latent fusion strategies have been successfully applied to integrate molecular descriptors~\cite{heid2023chemprop} or to combine phenotypic readouts, such as imaging, with molecular representations~\cite{moshkov2023predicting}. The predictions $y^{\mathrm{J}}_{\text{task}}$ are used to calculate standard metrics for each captioner, such as ROC-AUC or MAE. 

\subsection{Rating captioners}
\label{ssec:rating}
In addition to computing standard metrics for each captioner, we introduce a novel rating system inspired by Chatbot Arena~\cite{chiang2024chatbot}. This strategy allows for a more direct and comprehensive comparison of the impact of different caption sources. In particular, it provides two key benefits: 1) it allows for the seamless integration of multiple datasets, spanning both regression and classification tasks, into a single metric, and 2) it offers a fine-grained, molecule-specific evaluation of captioners relative to each other (whereas standard metrics are calculated for each model \textit{before} comparing different caption sources).

Similar to Chatbot Arena, we consider a ranking based on ``battles'' between different caption sources. However, instead of human preference feedback, we use the prediction error $\mathcal{E}(y^{\mathrm{J}}_{\text{task}}, \hat{y}_{\text{task}})$, where $\hat{y}_{\text{task}}$ is the ground truth label, as feedback. 
Specifically, given two captions $c_i(mol)$, $c_j(mol)$ as input to the model $M$, the caption that yields the lowest error is deemed the ``winner''. In our setup, the prediction error serves as a proxy for the additional information contributed by the caption to the task-specific molecular property. 

This approach requires a model $M$ trained to understand captions and molecule structure; here, a naive approach would be to train on all captions combined. However, this can bias the model towards caption distributions that are overrepresented in the benchmark (e.g., when multiple variations of the same captioner are present). %
To avoid this bias and improve robustness, we instead train a model $M_{i,j}$ for each pair of captioners, constructing a dataset consisting of both $x_{c_i}$ and $x_{c_j}$.
This ``head-to-head'' approach ensures no caption source has an unfair advantage due to related captions not being used in the battle.
As detailed in Subsection~\ref{ssec:joint-model}, we leverage an SVM on top of pre-trained embeddings for $M$. Importantly,  for a given molecule, the same GNN embedding $x_m$ is used for both caption sources. This ensures that, given two captions for a molecule, the error difference \textit{exclusively} comes from the capability of the caption to augment the molecular representation as optimized by the GNN. We further found this strategy to facilitate training.

After constructing these battles, we calculate ratings for each model (as well as model-vs-model win rates). We employ the Bradley-Terry model~\cite{bradley1952rank}, a well-known approach to rank items, %
to produce ratings (see  \citet{chiang2024chatbot}). %
This also allows us to bootstrap confidence intervals for our ratings (10 bootstrap experiments are performed: each is 250,000 battles, uniformly sampled across all datasets).

\end{multicols}

\begin{table*}[h!]
\centering
\small
\adjustbox{max width=\textwidth}{
\begin{tabular}{l|HcHcHcccccccccc}
\hline
model & lower & Rating & upper & 95\% CI & rating & ROC-AUC & BCE Loss & Avg. Error & Average Precision & Pearson R & Spearman R & $R^2$ & MSE & MAE \\
\midrule
BioT5\_plus & 1059.344 & 1061 & 1064.393 & +3 / -2 & 1061.390 & 87.359 & 0.335 & 0.471 & 74.907 & 0.654 & 0.660 & 0.408 & 1.057 & 0.744 \\
BioT5 & 1056.603 & 1059 & 1061.436 & +3 / -2 & 1058.780 & 87.371 & 0.334 & 0.472 & 74.631 & 0.654 & 0.658 & 0.409 & 1.057 & 0.744 \\
LlaSMol & 1036.177 & 1040 & 1042.149 & +2 / -3 & 1039.670 & 86.522 & 0.340 & 0.473 & 73.271 & 0.655 & 0.661 & 0.409 & 1.054 & 0.746 \\
GPT-4o-Generic & 1030.201 & 1032 & 1033.888 & +2 / -2 & 1032.340 & 85.685 & 0.341 & 0.472 & 72.388 & 0.657 & 0.662 & 0.412 & 1.049 & 0.742 \\
TextChemT5 & 1029.816 & 1032 & 1035.797 & +4 / -2 & 1032.120 & 87.074 & 0.341 & 0.475 & 75.295 & 0.652 & 0.657 & 0.407 & 1.057 & 0.746 \\
Gemma2-9B-Frags-Drug & 1027.232 & 1031 & 1033.738 & +3 / -4 & 1030.860 & 85.543 & 0.343 & 0.475 & 72.435 & 0.653 & 0.657 & 0.407 & 1.062 & 0.747 \\
Gemma2-9B-Frags-Chem & 1027.145 & 1030 & 1033.369 & +3 / -4 & 1030.040 & 85.619 & 0.342 & 0.475 & 72.312 & 0.652 & 0.657 & 0.406 & 1.061 & 0.747 \\
Llama3.1-70B-Chem & 1019.179 & 1021 & 1023.051 & +2 / -2 & 1021.280 & 85.431 & 0.344 & 0.475 & 72.529 & 0.653 & 0.656 & 0.408 & 1.056 & 0.746 \\
Llama3.1-70B-Generic & 1018.225 & 1021 & 1022.140 & +1 / -3 & 1020.830 & 85.618 & 0.344 & 0.474 & 72.373 & 0.653 & 0.657 & 0.408 & 1.054 & 0.744 \\
MolT5 & 1017.068 & 1021 & 1022.134 & +1 / -4 & 1020.730 & 86.406 & 0.340 & 0.475 & 75.251 & 0.652 & 0.655 & 0.406 & 1.064 & 0.748 \\
Gemma2-9B-Frags-Generic & 1016.929 & 1020 & 1023.279 & +3 / -3 & 1019.980 & 85.550 & 0.342 & 0.476 & 72.455 & 0.651 & 0.657 & 0.406 & 1.064 & 0.749 \\
Llama3.1-70B-Drug & 1018.127 & 1020 & 1020.978 & +1 / -2 & 1019.780 & 85.594 & 0.343 & 0.471 & 72.430 & 0.655 & 0.660 & 0.412 & 1.042 & 0.740 \\
Llama3.1-405B-Generic & 1018.138 & 1020 & 1022.579 & +3 / -1 & 1019.560 & 85.469 & 0.342 & 0.471 & 72.048 & 0.658 & 0.663 & 0.416 & 1.038 & 0.738 \\
Llama3-70B-Drug & 1014.723 & 1017 & 1019.738 & +2 / -3 & 1017.330 & 85.392 & 0.343 & 0.472 & 72.179 & 0.656 & 0.660 & 0.413 & 1.047 & 0.741 \\
MistralNeMo-12B-Frags-Generic & 1015.567 & 1017 & 1019.185 & +2 / -1 & 1017.020 & 85.376 & 0.345 & 0.473 & 72.022 & 0.653 & 0.658 & 0.408 & 1.051 & 0.742 \\
Llama3.1-70B-Bio & 1013.653 & 1016 & 1018.199 & +2 / -2 & 1016.020 & 85.238 & 0.347 & 0.473 & 72.074 & 0.656 & 0.661 & 0.411 & 1.047 & 0.741 \\
Gemma2-27B-Generic & 1012.850 & 1015 & 1018.055 & +3 / -3 & 1015.420 & 85.254 & 0.343 & 0.471 & 72.152 & 0.654 & 0.661 & 0.412 & 1.043 & 0.739 \\
Llama3-70B-Chem & 1010.570 & 1014 & 1016.581 & +2 / -4 & 1014.370 & 85.355 & 0.344 & 0.473 & 72.181 & 0.655 & 0.659 & 0.409 & 1.053 & 0.742 \\
Gemma2-27B-Chem & 1009.634 & 1013 & 1013.970 & +1 / -3 & 1012.730 & 85.223 & 0.345 & 0.474 & 72.057 & 0.653 & 0.659 & 0.409 & 1.051 & 0.743 \\
MistralNeMo-12B-Frags-Chem & 1008.842 & 1012 & 1014.084 & +2 / -3 & 1011.960 & 85.746 & 0.343 & 0.472 & 72.320 & 0.656 & 0.660 & 0.411 & 1.046 & 0.741 \\
GPT-4o-Frags-Generic & 1009.718 & 1012 & 1014.869 & +3 / -2 & 1011.780 & 85.510 & 0.343 & 0.475 & 72.210 & 0.654 & 0.658 & 0.408 & 1.054 & 0.745 \\
Llama3-8B-Task & 1006.404 & 1009 & 1009.875 & +1 / -2 & 1008.660 & 85.414 & 0.344 & 0.477 & 72.182 & 0.652 & 0.656 & 0.406 & 1.063 & 0.748 \\
MistralNeMo-12B-Frags-Drug & 1006 & 1008 & 1011.182 & +3 / -2 & 1008.410 & 85.731 & 0.344 & 0.473 & 72.436 & 0.654 & 0.659 & 0.410 & 1.046 & 0.742 \\
Llama3-70B-Generic & 1006.451 & 1008 & 1011.691 & +3 / -2 & 1008.290 & 85.108 & 0.345 & 0.475 & 71.920 & 0.653 & 0.656 & 0.406 & 1.061 & 0.746 \\
Gemma2-9B-Frags-Bio & 1003.444 & 1006 & 1009.666 & +3 / -3 & 1006.330 & 85.284 & 0.346 & 0.477 & 72.020 & 0.651 & 0.657 & 0.405 & 1.065 & 0.750 \\
Llama3.1-70B-Quant & 1003.130 & 1006 & 1008.633 & +3 / -3 & 1006.000 & 85.312 & 0.344 & 0.474 & 72.163 & 0.653 & 0.660 & 0.408 & 1.052 & 0.743 \\
Llama3.1-405B-Frags-Generic & 1002.802 & 1005 & 1007.438 & +2 / -2 & 1005.050 & 85.482 & 0.344 & 0.476 & 72.682 & 0.653 & 0.657 & 0.405 & 1.065 & 0.749 \\
Gemma2-9B-Chem & 1002.009 & 1004 & 1005.319 & +1 / -2 & 1004.190 & 85.455 & 0.343 & 0.474 & 72.112 & 0.653 & 0.657 & 0.409 & 1.057 & 0.745 \\
Llama3-70B-Bio & 1001.010 & 1003 & 1004.506 & +1 / -2 & 1003.290 & 85.479 & 0.344 & 0.476 & 72.253 & 0.651 & 0.656 & 0.403 & 1.066 & 0.748 \\
Llama3-70B-Frags-Generic & 1000.327 & 1003 & 1006.045 & +3 / -3 & 1003.050 & 85.780 & 0.342 & 0.478 & 72.693 & 0.650 & 0.655 & 0.403 & 1.072 & 0.751 \\
Llama3.1-8B-Drug & 1000.990 & 1003 & 1006.726 & +4 / -2 & 1002.690 & 85.088 & 0.345 & 0.475 & 72.004 & 0.654 & 0.657 & 0.410 & 1.052 & 0.745 \\
Llama3.1-70B-Frags-Generic & 999.298 & 1001 & 1004.401 & +3 / -2 & 1001.450 & 85.580 & 0.342 & 0.477 & 72.238 & 0.650 & 0.656 & 0.403 & 1.071 & 0.750 \\
Llama3.1-70B-Frags-Chem & 999.164 & 1001 & 1004.037 & +3 / -2 & 1001.290 & 85.512 & 0.343 & 0.477 & 72.306 & 0.651 & 0.657 & 0.404 & 1.070 & 0.749 \\
Gemma2-27B-Frags-Drug & 997.126 & 1001 & 1002.575 & +2 / -4 & 1000.870 & 85.644 & 0.342 & 0.477 & 72.333 & 0.649 & 0.653 & 0.402 & 1.069 & 0.752 \\
MistralNeMo-12B-Bio & 997.934 & 1000 & 1001.403 & +1 / -2 & 1000.250 & 85.301 & 0.344 & 0.476 & 72.144 & 0.650 & 0.655 & 0.403 & 1.066 & 0.748 \\
Llama3-8B-Drug & 997.014 & 999 & 1001.968 & +3 / -2 & 999.240 & 85.375 & 0.344 & 0.474 & 72.148 & 0.654 & 0.658 & 0.408 & 1.057 & 0.745 \\
MistralNeMo-12B-Generic & 997.840 & 999 & 1000.418 & +1 / -1 & 998.990 & 85.421 & 0.343 & 0.477 & 72.113 & 0.650 & 0.654 & 0.403 & 1.069 & 0.750 \\
Gemma2-27B-Bio & 996.521 & 998 & 1000.162 & +2 / -1 & 997.920 & 85.347 & 0.343 & 0.476 & 72.176 & 0.652 & 0.657 & 0.405 & 1.061 & 0.747 \\
Gemma2-9B-Bio & 995.259 & 997 & 999.509 & +3 / -2 & 996.770 & 85.392 & 0.344 & 0.475 & 72.139 & 0.651 & 0.656 & 0.406 & 1.057 & 0.746 \\
Gemma2-9B-Frags-Quant & 994.784 & 997 & 1001.680 & +5 / -2 & 996.650 & 85.305 & 0.346 & 0.475 & 72.035 & 0.652 & 0.656 & 0.406 & 1.060 & 0.746 \\
Gemma2-27B-Quant & 992.329 & 997 & 999.470 & +3 / -4 & 996.630 & 85.348 & 0.346 & 0.476 & 72.120 & 0.651 & 0.656 & 0.405 & 1.061 & 0.746 \\
MistralNeMo-12B-Frags-Bio & 993.067 & 996 & 998.245 & +2 / -3 & 996.300 & 85.364 & 0.345 & 0.475 & 72.082 & 0.652 & 0.658 & 0.406 & 1.060 & 0.746 \\
ChemDFM & 992.820 & 996 & 999.755 & +4 / -3 & 996.200 & 86.052 & 0.344 & 0.477 & 72.614 & 0.648 & 0.653 & 0.401 & 1.073 & 0.751 \\
MistralNeMo-12B-Chem & 991.621 & 996 & 997.068 & +1 / -5 & 996.160 & 85.418 & 0.346 & 0.475 & 72.201 & 0.651 & 0.655 & 0.406 & 1.063 & 0.746 \\
3D-MoLM & 992.736 & 996 & 997.874 & +2 / -3 & 995.970 & 86.032 & 0.341 & 0.475 & 73.161 & 0.651 & 0.655 & 0.406 & 1.062 & 0.748 \\
ChemLLM & 993.082 & 996 & 998.634 & +3 / -3 & 995.950 & 85.810 & 0.342 & 0.475 & 72.849 & 0.650 & 0.654 & 0.405 & 1.063 & 0.747 \\
Llama3-8B-Frags-Bio & 994.737 & 996 & 997.578 & +2 / -1 & 995.810 & 85.338 & 0.344 & 0.477 & 72.102 & 0.652 & 0.657 & 0.405 & 1.071 & 0.750 \\
Llama3-8B-Quant & 991.436 & 995 & 998.111 & +3 / -4 & 995.380 & 85.397 & 0.344 & 0.476 & 72.154 & 0.650 & 0.654 & 0.405 & 1.064 & 0.747 \\
Llama3-8B-Frags-Chem & 992.936 & 995 & 998.163 & +3 / -2 & 995.220 & 85.336 & 0.344 & 0.476 & 72.086 & 0.654 & 0.659 & 0.406 & 1.065 & 0.749 \\
Llama3.1-8B-Generic & 991.820 & 995 & 996.981 & +2 / -3 & 994.850 & 85.331 & 0.343 & 0.476 & 72.231 & 0.652 & 0.657 & 0.406 & 1.062 & 0.748 \\
MistralNeMo-12B-Frags-Quant & 992.347 & 994 & 997.003 & +3 / -2 & 994.460 & 85.194 & 0.344 & 0.476 & 72.143 & 0.651 & 0.655 & 0.403 & 1.065 & 0.749 \\
Llama3-70B-Frags-Drug & 992.068 & 994 & 997.374 & +3 / -2 & 994.310 & 85.590 & 0.342 & 0.478 & 72.270 & 0.650 & 0.655 & 0.402 & 1.073 & 0.751 \\
Gemma2-27B-Drug & 991.308 & 994 & 997.072 & +3 / -3 & 994.080 & 85.332 & 0.344 & 0.473 & 72.067 & 0.653 & 0.659 & 0.409 & 1.045 & 0.741 \\
Llama3-70B-Quant & 991.359 & 993 & 995.475 & +2 / -2 & 993.340 & 85.241 & 0.344 & 0.474 & 72.109 & 0.654 & 0.657 & 0.408 & 1.055 & 0.745 \\
Llama3-8B-Frags-Drug & 990.841 & 993 & 995.669 & +2 / -2 & 993.310 & 85.352 & 0.343 & 0.477 & 72.144 & 0.652 & 0.656 & 0.406 & 1.064 & 0.748 \\
Gemma2-9B-Generic & 989.555 & 993 & 997.394 & +4 / -3 & 992.970 & 85.412 & 0.344 & 0.476 & 72.085 & 0.650 & 0.654 & 0.404 & 1.067 & 0.748 \\
Llama3.1-8B-Chem & 990.477 & 992 & 995.488 & +3 / -2 & 992.130 & 85.425 & 0.344 & 0.475 & 72.214 & 0.652 & 0.658 & 0.406 & 1.059 & 0.746 \\
MolInstructions\_molecule & 987.809 & 992 & 995.939 & +4 / -4 & 992.060 & 85.747 & 0.342 & 0.477 & 72.594 & 0.650 & 0.654 & 0.402 & 1.070 & 0.750 \\
Llama3.1-70B-Frags-Bio & 988.897 & 991 & 994.411 & +4 / -2 & 990.890 & 85.379 & 0.343 & 0.477 & 72.171 & 0.649 & 0.653 & 0.401 & 1.075 & 0.751 \\
Llama3-8B-Generic & 986.197 & 991 & 995.029 & +4 / -5 & 990.820 & 85.349 & 0.344 & 0.476 & 72.227 & 0.650 & 0.656 & 0.405 & 1.062 & 0.747 \\
Llama3-70B-Frags-Chem & 986.967 & 990 & 992.543 & +2 / -4 & 990.480 & 85.295 & 0.345 & 0.478 & 72.278 & 0.651 & 0.656 & 0.403 & 1.070 & 0.750 \\
Llama3-70B-Frags-Quant & 987.812 & 990 & 992.552 & +2 / -3 & 990.340 & 85.355 & 0.344 & 0.475 & 72.139 & 0.654 & 0.658 & 0.406 & 1.065 & 0.748 \\
Llama3-8B-Frags-Quant & 987.419 & 990 & 992.273 & +2 / -3 & 990.220 & 85.375 & 0.344 & 0.476 & 72.171 & 0.655 & 0.660 & 0.407 & 1.064 & 0.748 \\
Llama3.1-8B-Quant & 985.181 & 990 & 991.954 & +2 / -5 & 990.020 & 85.511 & 0.343 & 0.476 & 72.339 & 0.650 & 0.655 & 0.403 & 1.067 & 0.748 \\
Llama3-8B-Frags-Generic & 986.207 & 989 & 990.875 & +2 / -3 & 989.070 & 85.271 & 0.347 & 0.477 & 72.035 & 0.652 & 0.658 & 0.404 & 1.069 & 0.749 \\
Llama3.1-70B-Frags-Drug & 986.595 & 988 & 992.191 & +4 / -1 & 988.050 & 85.332 & 0.344 & 0.477 & 72.132 & 0.650 & 0.654 & 0.403 & 1.073 & 0.750 \\
Llama3-8B-Chem & 984.343 & 988 & 991.095 & +3 / -3 & 987.660 & 85.359 & 0.344 & 0.475 & 72.083 & 0.651 & 0.656 & 0.405 & 1.059 & 0.746 \\
Llama3-70B-Frags-Bio & 980.275 & 986 & 988.781 & +3 / -6 & 985.940 & 85.389 & 0.343 & 0.477 & 72.168 & 0.650 & 0.653 & 0.402 & 1.074 & 0.751 \\
Llama3-8B-Bio & 983.534 & 986 & 988.284 & +3 / -2 & 985.640 & 85.456 & 0.343 & 0.474 & 72.265 & 0.650 & 0.656 & 0.405 & 1.064 & 0.746 \\
Gemma2-9B-Quant & 982.224 & 984 & 986.458 & +2 / -2 & 984.300 & 85.329 & 0.344 & 0.476 & 72.110 & 0.650 & 0.654 & 0.404 & 1.063 & 0.749 \\
Gemma2-9B-Drug & 980.855 & 984 & 986.668 & +3 / -3 & 984.060 & 85.266 & 0.344 & 0.477 & 72.262 & 0.649 & 0.654 & 0.403 & 1.068 & 0.750 \\
Gemma2-27B-Frags-Chem & 980.360 & 983 & 984.500 & +1 / -3 & 983.250 & 85.319 & 0.345 & 0.478 & 72.136 & 0.647 & 0.651 & 0.400 & 1.078 & 0.753 \\
Llama3.1-70B-Frags-Quant & 981.347 & 983 & 987.562 & +5 / -2 & 983.020 & 85.185 & 0.344 & 0.477 & 72.088 & 0.652 & 0.657 & 0.405 & 1.069 & 0.749 \\
Llama3.1-8B-Frags-Bio & 979.837 & 982 & 984.236 & +2 / -2 & 982.030 & 85.322 & 0.343 & 0.479 & 72.117 & 0.647 & 0.651 & 0.399 & 1.079 & 0.753 \\
Llama3.1-8B-Bio & 978.329 & 982 & 982.298 & +1 / -3 & 981.550 & 85.431 & 0.344 & 0.476 & 72.155 & 0.649 & 0.654 & 0.403 & 1.064 & 0.748 \\
MistralNeMo-12B-Quant & 979.582 & 981 & 984.361 & +3 / -2 & 981.310 & 85.404 & 0.343 & 0.478 & 72.092 & 0.649 & 0.654 & 0.403 & 1.068 & 0.751 \\
Llama3.1-8B-Frags-Quant & 977.205 & 980 & 983.143 & +3 / -3 & 980.240 & 85.311 & 0.345 & 0.479 & 72.091 & 0.648 & 0.652 & 0.401 & 1.074 & 0.752 \\
MolT5\_LPM24 & 976.423 & 980 & 981.492 & +1 / -4 & 980.100 & 85.385 & 0.343 & 0.477 & 72.140 & 0.649 & 0.653 & 0.402 & 1.071 & 0.750 \\
Llama3.1-8B-Frags-Chem & 975.452 & 978 & 979.146 & +1 / -2 & 977.820 & 85.340 & 0.344 & 0.478 & 72.132 & 0.647 & 0.652 & 0.400 & 1.078 & 0.752 \\
BlankCaption & 975.538 & 978 & 981.136 & +3 / -2 & 977.790 & 85.336 & 0.344 & 0.478 & 72.127 & 0.647 & 0.651 & 0.399 & 1.079 & 0.753 \\
Gemma2-27B-Frags-Quant & 976.067 & 978 & 980.094 & +3 / -1 & 977.560 & 85.423 & 0.342 & 0.478 & 72.169 & 0.647 & 0.651 & 0.399 & 1.077 & 0.752 \\
Llama3.1-8B-Frags-Generic & 972.455 & 977 & 981.315 & +5 / -4 & 976.690 & 85.326 & 0.346 & 0.478 & 72.101 & 0.648 & 0.653 & 0.401 & 1.074 & 0.751 \\
MistralNeMo-12B-Drug & 972.775 & 975 & 979.960 & +5 / -2 & 974.620 & 85.323 & 0.345 & 0.477 & 72.097 & 0.650 & 0.654 & 0.402 & 1.070 & 0.750 \\
Gemma2-27B-Frags-Bio & 971.442 & 973 & 975.506 & +2 / -2 & 973.350 & 85.431 & 0.343 & 0.478 & 72.274 & 0.647 & 0.651 & 0.399 & 1.079 & 0.753 \\
Gemma2-27B-Frags-Generic & 969.671 & 973 & 973.991 & +1 / -3 & 972.720 & 85.362 & 0.342 & 0.478 & 72.032 & 0.647 & 0.652 & 0.399 & 1.078 & 0.753 \\
Llama3.1-8B-Frags-Drug & 969.250 & 970 & 972.783 & +2 / -1 & 970.290 & 85.353 & 0.345 & 0.479 & 72.122 & 0.647 & 0.651 & 0.399 & 1.079 & 0.753 \\
GNN & NaN & NaN & NaN & NaN & NaN & 85.336 & 0.345 & 0.479 & 72.111 & 0.647 & 0.652 & 0.397 & 1.080 & 0.753 \\
\hline
\end{tabular}
}
\caption{Full Leaderboard and Performance Metrics}
\label{tab:full_model_performance}
\end{table*}

\twocolumn

\section{Experiments}
\paragraph{Molecular datasets.} To fairly rank captions, it is important to evaluate across multiple tasks. We select six datasets from the MoleculeNet benchmark~\cite{wu2018moleculenet} (Appendix~\ref{appendix:datasets}). These datasets represent diverse real-world applications, privileging small/medium datasets, as they can benefit more from multimodal augmentations.

\paragraph{Caption sources.} We consider a wide variety of domain-specific captioners \cite{edwards2022translation, christofidellis2023unifying} and general-purpose LLMs \cite{achiam2023gpt, dubey2024llama, team2024gemma}. For general-purpose LLMs, we consider different ``personas'' for each model: a biologist, chemist, drug researcher, quantum chemist, and generic chemist. We supply the target molecule to each LLM either as the commonly-employed SMILES string \cite{weininger1988smiles, weininger1989smiles}, or as a set of fragments (using the the BRICS algorithm \cite{degen2008art}). %
We also consider two theoretical caption sources as controls: the ``BlankCaption'', which contains no content, and a ``BBBP Oracle'', which states the BBBP ground truth, reporting consistent results (see Appendix \ref{appendix:oracle}).
All results are repeated across 5 folds. 

\paragraph{Captions consistently lead to improved performance, with dataset- and task-specific variations.} Table \ref{tab:full_model_performance} presents results for models. Overall, captions enhance the baseline performance of the GNN across all tasks. For classification tasks, the average ROC-AUC increases from 85.34 to 87.37, while for regression tasks, the average $R^2$ improves from 0.397 to 0.413.
We observe how multiple captioners improve GNN performance on some datasets, such as ClinTox, ESOL, and FreeSolv, while their impact is less pronounced on others, thus underscoring the task-specific nature of LLM-driven enhancements.

\paragraph{Domain-specific captioners lead to best improvements, followed by large-scale, general-purpose LLMs.} Generally, domain-specific captioners such as BioT5\_plus and LLasMol lead to the best results. This is consistent with their specialized fine-tuning datasets.  However, we also observe that strong general-purpose captioners, such as GPT-4o and Llama3.1-405B, rank among the top models, demonstrating their ability to perform competitively even without domain-specific training.

\paragraph{Larger models correlate with score improvements, but model-specific safeguards can lead to reduced performance.} Grouping ratings by model family (Table~\ref{tab:model_model_performance}), we observe that Llama3.1-405B slightly outperforms the 70B version, and both significantly outperform the 8B version. Interestingly, however, Llama3-8B outperforms Llama3.1-8B. Upon further investigation, we discovered that this is due to an overzealous safeguarding mechanism in Llama3.1-8B, which frequently refuses to describe molecular structures (see Appendix \ref{app:caption_examples} for examples). We note this may be a concerning development for the use of general LLMs in molecule-related tasks.

\paragraph{Different personas and molecular representations can benefit different tasks and models.}~Grouping general-purpose LLMs by persona and molecular representation used, we observe both general trends and task-specific variations. On average, the chemist and drug researcher personas obtained the highest rankings (Table~\ref{tab:prompt_model_performance}), which is consistent with their general, while still domain-specific prompt. Additionally, we find that captions of SMILES strings generally outperform captions of BRICS fragments (Table~\ref{tab:frag_model_performance}). However, we observe how the impact of different prompts is task-, representation- and model-specific, thus highlighting the need for future research into the connection between prompts and downstream performance.

\paragraph{Task-specific prompts enhance performance.} In addition to different personas, we evaluate the impact of \emph{task-specific prompts} in general-purpose LLMs, where the model is provided a custom prompt for each different task. For this, we implement the Llama3-8B-Task captioner. Notably, when compared to other Llama-8B variations (Appendix~\ref{appendix:llama3_8b_performance}), the task-specific model leads to the best performance. This highlights the importance of application-specific prompts for molecular modeling tasks and underscores a trade-off between efficiency (i.e., general captions which can support multiple tasks) and accuracy.

\paragraph{Battle-based rating provides a more robust and complementary evaluation metric.}
In addition to the theoretical advantages provided by the proposed rating metric (Subsection~\ref{ssec:rating}), we investigated its empirical robustness. Notably, we observe how our proposed rating system, even if correlated with standard metrics (Table~\ref{tab:correlation}), 
has better agreement across different tasks. For example, between the three classification tasks and three regression tasks, there is a 46\% Spearman correlation between ratings, but only a 12\% Spearman correlation between average error (MAE) (Appendix \ref{appendix:classification_vs_regression}).

\section{Conclusion}

We introduce \name{}, a robust, large-scale benchmark for evaluating the capability of molecule captions to augment real-world property prediction tasks. Further, we propose a rating system for ranking models, which allows performance to be aggregated across different task types. This enables a new way of evaluating molecule captioners, which have primarily been evaluated using string-based metrics. Future work may also consider multi-modality evaluation. We make this benchmark and generated resources available to the community. 

\section*{Limitations}

Although we make considerable effort to produce a fair and accurate ranking between caption sources, there is likely still effects from unforeseen biases in the source models and their training datasets. Future captioners may also try to game the leaderboard. To some extent, this can be alleviated by extending future versions of the leaderboard to new datasets. We note multi-task datasets as an interesting direction for future research. Further, we use a fairly simple architecture which may limit the capabilities of some captioners. We leave the study of more advanced molecule-language fusion architectures to future work. Another possible limitation is that we choose battles based purely on which caption source produces a lower error. It may be beneficial to take into account margin of victory in future work. In our preliminary testing, we found this to produce similar results, so we opted not to introduce additional complexity. 

\section*{Ethical Considerations}
The use of predictive models for biological and chemical modeling unfortunately suffers from a dual use problem \cite{urbina2022dual}, where the technology can be greatly beneficial but also detrimental. We note that our benchmark is created from open datasets where predictive models already exist, so our work introduces no new harm. Further, we believe our benchmark to be useful here: the usage of natural language captions for improving predictive performance increases explainability by allowing domain scientists to read which relevant information improves molecule property prediction. This can potentially mitigate harms in employing this technology by allowing a clearer view into the molecular discovery process. 

\section*{Acknowledements}

ZL, EH, TB, and GS are employees of Genentech, Inc., and shareholders of Roche. 
This research is based in part upon work supported by the Molecule Maker Lab Institute: an AI research institute program supported by NSF under award No. 2019897. The views and conclusions contained herein are those of the authors and should not be interpreted as necessarily representing the official policies, either expressed or implied, of the U.S. Government. The U.S. Government is authorized to reproduce and distribute reprints for governmental purposes notwithstanding any copyright annotation therein.

\newpage

\bibliography{custom_fix}

\newpage 
\clearpage

\appendix

\twocolumn
\section{Datasets}
\label{appendix:datasets}

We select six representative datasets from MoleculeNet \cite{wu2018moleculenet}. These are selected to represent task diversity (3 regression and 3 classification tasks), application diversity, and to be fairly small to allow efficient captioning for community use. Each dataset is split into 60\% \emph{training}, 20\% \emph{preference}, 10\% \emph{validation} and 10\% \emph{test} split following a scaffold splitting strategy to simulate chemical distribution shifts.

\begin{itemize}
    \item BBBP: blood brain barrier penetration prediction (binary classification)
    \item BACE: beta-secretase 1 binding prediction (binary classification)
    \item ClinTox: clinical trial toxicity prediction (binary classification)
    \item FreeSolv: hydration-free energy prediction of small molecules in water (regression)
    \item ESOL: aqueous solubility prediction (regression)
    \item Lipo: lipophilicity prediction (regression)
\end{itemize}

\section{Caption Sources}
\begin{enumerate}
    \item 3D-MoLM \cite{li2024towards}
    \item BioT5 \cite{pei2023biot5}
    \item BioT5\_plus \cite{pei2024biot5+}
    \item BlankCaption 
    \item ChemDFM \cite{zhao2024chemdfm}
    \item ChemLLM \cite{zhang2024chemllm}
    \item GPT-4o \cite{achiam2023gpt}
    \item Gemma2-9B \cite{team2024gemma}
    \item Gemma2-27B \cite{team2024gemma}
    \item LlaSMol \cite{yu2024llasmol}
    \item Llama3-8B \cite{dubey2024llama}
    \item Llama3-70B \cite{dubey2024llama}
    \item Llama3.1-8B \cite{dubey2024llama}
    \item Llama3.1-70B \cite{dubey2024llama}
    \item Llama3.1-405B \cite{dubey2024llama}
    \item MistralNeMo-12B \url{https://mistral.ai/news/mistral-nemo/}
    \item MolInstructions\_molecule \cite{fang2023mol}
    \item MolT5 \cite{edwards2022translation}
    \item MolT5\_LPM24 \cite{edwards2024_lpm24}
    \item TextChemT5 \cite{christofidellis2023unifying}
\end{enumerate}

We note that MolInstructions, 3D-MoLM, ChemDFM, and ChemLLM hyperparameters were ambiguous, so sampled captions may not be perfect. 

\section{Prompt Variations}
We consider the following prompt variations to extract different information from general-purpose LLMs.\\
\textbf{Generic}:
\begin{tcolorbox}
System Prompt: ``You are a helpful expert in chemistry. You are sharing your knowledge of known chemistry with a colleague seeking to better understand how small molecules work.'' \\
Prompt: ``The following molecule has important properties. Tell me about relevant functionality and applications of this molecule.''
\end{tcolorbox}
\textbf{Drug}:
\begin{tcolorbox}
System Prompt: ``You are a helpful expert in medicinal chemistry. In particular, you specialize in the design of small molecule drugs. You are sharing your knowledge of known chemistry with a colleague seeking to better understand drug design.'' \\
Prompt: ``The following molecule might be a drug. Tell me properties of the molecule, such as the mechanism of action, class, and target, which might be relevant.''
\end{tcolorbox}
\textbf{Biology}:
\begin{tcolorbox}
System Prompt: ``You are a helpful expert in biology. You are sharing your knowledge of known biochemistry with a colleague.'' \\
Prompt: ``The following molecule has important biological properties. Tell me biological properties of the molecule and relevant functionality and applications.''
\end{tcolorbox}
\textbf{Chemistry}:
\begin{tcolorbox}
System Prompt: ``You are a helpful expert in chemistry. You are sharing your knowledge of known chemistry with a colleague.'' \\
Prompt: ``The following molecule has important chemical properties. Tell me chemical properties of the molecule and relevant functionality and applications.''
\end{tcolorbox}
\textbf{Quantum Chemistry}:
\begin{tcolorbox}
System Prompt: ``You are a helpful expert in chemistry. You are sharing your knowledge of known chemistry with a colleague seeking to better understand how small molecules work.'' \\
Prompt: ``The following molecule has important quantum properties. Tell me quantum properties of the molecule and relevant functionality and applications.''
\end{tcolorbox}
\textbf{Task}:
\begin{tcolorbox}
System Prompt: ``You are a helpful expert in chemistry. You are sharing your knowledge of known chemistry with a colleague seeking to better understand how small molecules work.'' \\
Prompt: ``The following molecule has important properties. $\backslash$nMolecule: \{SMILES\}$\backslash$n$\backslash$nTell me about relevant properties and functionality of this molecule which indicate whether it can complete the following task.$\backslash$nTask: \{task\_desc\}$\backslash$n$\backslash$n''
\end{tcolorbox}

The following task descriptions are used:
\begin{tcolorbox}
\begin{tiny}
\begin{verbatim}
task_desc = {
    'BBBP' : 'blood brain barrier penetration prediction',
    'ClinTox' : 'clinical trial toxicity prediction',
    'BACE' : 'beta-secretase 1 binding prediction',
    'ESOL' : 'aqueous solubility prediction',
    'FreeSolv' : 'hydration-free energy prediction
                  of small molecules in water',
    'Lipo' : 'lipophilicity prediction',
}
\end{verbatim}
\end{tiny}
\end{tcolorbox}

\textbf{Generic Fragments}:
\begin{tcolorbox}
System Prompt: ``You are a helpful expert in chemistry. You are sharing your knowledge of known chemistry with a colleague seeking to better understand how small molecules work.'' \\
Prompt: ``The following molecular fragments have important properties. Tell me about relevant functionality and applications of these molecular fragments.''
\end{tcolorbox}
\textbf{Drug Fragments}:
\begin{tcolorbox}
System Prompt: ``You are a helpful expert in medicinal chemistry. In particular, you specialize in the design of small molecule drugs. You are sharing your knowledge of known chemistry with a colleague seeking to better understand drug design.'' \\
Prompt: ``The following molecular fragments constitute a possible drug. Tell me properties of these fragments, such as the mechanism of action, class, and target, which might be relevant.''
\end{tcolorbox}
\textbf{Biology Fragments}:
\begin{tcolorbox}
System Prompt: ``You are a helpful expert in quantum chemistry. You are sharing your knowledge of known chemistry with a colleague.'' \\
Prompt: ``The following molecular fragments have important biological properties. Tell me biological properties of these fragments and possible relevant functionality and applications.''
\end{tcolorbox}
\textbf{Chemistry Fragments}:
\begin{tcolorbox}
System Prompt: ``You are a helpful expert in chemistry. You are sharing your knowledge of known chemistry with a colleague.'' \\
Prompt: ``The following molecular fragments have important chemical properties. Tell me chemical properties of these fragments and possible relevant functionality and applications.''
\end{tcolorbox}
\textbf{Quantum Chemistry Fragments}:
\begin{tcolorbox}
System Prompt: ``You are a helpful expert in chemistry. You are sharing your knowledge of known chemistry with a colleague seeking to better understand how small molecules work.'' \\
Prompt: ``The following molecular fragments have important quantum properties. Tell me quantum properties of these fragments and possible relevant functionality and applications.''
\end{tcolorbox}

\newpage
\clearpage

\onecolumn
\section{Metric Correlations}

\begin{table*}[h!]
\centering
\resizebox{\textwidth}{!}{
\begin{tabular}{l|HcHHHHccccccccc}
\hline
 & lower & Rating & upper & error\_y & error\_y\_minus & rating\_rounded & ROC-AUC & BCE Loss & Avg. Error & Average Precision & Pearson R & Spearman R & $R^2$ & MSE & MAE \\
\hline
 & \multicolumn{15}{c}{Pearson Correlations} \\
Rating & 0.999 & 1.000 & 0.999 & -0.084 & -0.054 & 1.000 & 0.649 & -0.570 & -0.734 & 0.585 & 0.696 & 0.686 & 0.703 & -0.641 & -0.624 \\
ROC-AUC & 0.643 & 0.649 & 0.651 & 0.016 & 0.054 & 0.649 & 1.000 & -0.840 & -0.342 & 0.922 & 0.164 & 0.173 & 0.180 & -0.136 & -0.135 \\
BCE Loss & -0.569 & -0.570 & -0.570 & 0.034 & 0.038 & -0.570 & -0.840 & 1.000 & 0.303 & -0.741 & -0.148 & -0.146 & -0.150 & 0.077 & 0.073 \\
Avg. Error & -0.734 & -0.734 & -0.731 & 0.098 & 0.061 & -0.734 & -0.342 & 0.303 & 1.000 & -0.283 & -0.853 & -0.845 & -0.921 & 0.931 & 0.952 \\
AP & 0.580 & 0.585 & 0.587 & -0.002 & 0.067 & 0.585 & 0.922 & -0.741 & -0.283 & 1.000 & 0.132 & 0.119 & 0.150 & -0.104 & -0.105 \\
Pearson R & 0.701 & 0.696 & 0.695 & -0.070 & -0.143 & 0.696 & 0.164 & -0.148 & -0.853 & 0.132 & 1.000 & 0.958 & 0.955 & -0.883 & -0.865 \\
\end{tabular}
}
\caption{Correlations between ratings and other metrics on all datasets}
\label{tab:correlation}
\end{table*}

\newpage
\clearpage

\section{Llama3-8B Comparison}
\label{appendix:llama3_8b_performance}

\begin{table*}[h!]
\centering
\small
\adjustbox{max width=\textwidth}{
\begin{tabular}{l|HcHHHHcccccccccc}
\hline
model & lower & Rating & upper & & 95\% CI & rating & ROC-AUC & BCE Loss & Avg. Error & Average Precision & Pearson R & Spearman R & $R^2$ & MSE & MAE \\
\midrule
Llama3-8B-Task & 1014.877 & 1016.014 & 1016.709 & +1& -1 & 1016.010 & 85.414 & 0.344 & 0.477 & 72.182 & 0.652 & 0.656 & 0.406 & 1.063 & 0.748 \\
Llama3-8B-Quant & 1007.523 & 1008.241 & 1008.986 & +1& -1 & 1008.240 & 85.397 & 0.344 & 0.476 & 72.154 & 0.650 & 0.654 & 0.405 & 1.064 & 0.747 \\
Llama3-8B-Drug & 1005.890 & 1006.850 & 1007.601 & 0.751 & 0.960 & 1006.850 & 85.375 & 0.344 & 0.474 & 72.148 & 0.654 & 0.658 & 0.408 & 1.057 & 0.745 \\
Llama3-8B-Frags-Bio & 1002.799 & 1003.334 & 1004.306 & 0.971 & 0.535 & 1003.330 & 85.338 & 0.344 & 0.477 & 72.102 & 0.652 & 0.657 & 0.405 & 1.071 & 0.750 \\
Llama3-8B-Frags-Quant & 999.117 & 1000.260 & 1000.903 & 0.643 & 1.143 & 1000.260 & 85.375 & 0.344 & 0.476 & 72.171 & 0.655 & 0.660 & 0.407 & 1.064 & 0.748 \\
Llama3-8B-Frags-Chem & 998.362 & 998.747 & 1000.296 & 1.548 & 0.385 & 998.750 & 85.336 & 0.344 & 0.476 & 72.086 & 0.654 & 0.659 & 0.406 & 1.065 & 0.749 \\
Llama3-8B-Generic & 996.386 & 997.181 & 997.714 & 0.534 & 0.795 & 997.180 & 85.349 & 0.344 & 0.476 & 72.227 & 0.650 & 0.656 & 0.405 & 1.062 & 0.747 \\
\bottomrule
\end{tabular}
}
\caption{Llama3-8B Leaderboard and Performance Metrics}
\label{tab:llama3_model_performance}
\end{table*}

\newpage
\clearpage

\section{Grouped Leaderboards}
\label{appendix:grouped_performance}

\FloatBarrier

\begin{table*}[h!]
\centering
\small
\adjustbox{max width=\textwidth}{
\begin{tabular}{l|HcHHHcccccccccc}
\hline
model & lower & Rating & upper & & & rating & ROC-AUC & BCE Loss & Avg. Error & Average Precision & Pearson R & Spearman R & $R^2$ & MSE & MAE \\
\midrule
BioT5\_plus & 1049.056 & 1050.550 & 1052.430 & 1.879 & 1.495 & 1050.550 & 87.359 & 74.907 & 0.335 & 0.471 & 0.654 & 0.660 & 0.408 & 1.057 & 0.744 \\
BioT5 & 1045.109 & 1046.186 & 1049.349 & 3.163 & 1.076 & 1046.190 & 87.371 & 74.631 & 0.334 & 0.472 & 0.654 & 0.658 & 0.409 & 1.057 & 0.744 \\
LlaSMol & 1025.038 & 1029.055 & 1032.914 & 3.859 & 4.017 & 1029.060 & 86.522 & 73.271 & 0.340 & 0.473 & 0.655 & 0.661 & 0.409 & 1.054 & 0.746 \\
TextChemT5 & 1020.317 & 1021.741 & 1024.724 & 2.983 & 1.425 & 1021.740 & 87.074 & 75.295 & 0.341 & 0.475 & 0.652 & 0.657 & 0.407 & 1.057 & 0.746 \\
MolT5 & 1006.964 & 1007.849 & 1011.594 & 3.746 & 0.885 & 1007.850 & 86.406 & 75.251 & 0.340 & 0.475 & 0.652 & 0.655 & 0.406 & 1.064 & 0.748 \\
Llama3-8B-Task & 994.631 & 996.769 & 1000.282 & 3.513 & 2.138 & 996.770 & 85.414 & 72.182 & 0.344 & 0.477 & 0.652 & 0.656 & 0.406 & 1.063 & 0.748 \\
General LLM & 989.009 & 989.696 & 990.652 & 0.956 & 0.687 & 989.700 & 85.375 & 72.173 & 0.344 & 0.475 & 0.652 & 0.657 & 0.407 & 1.058 & 0.745 \\
\bottomrule
\end{tabular}
}
\caption{Leaderboard Grouped by Fragment vs. No Fragment}
\label{tab:frag_model_performance}
\bigskip\bigskip

\adjustbox{max width=\textwidth}{
\begin{tabular}{l|HcHHHcccccccccc}
\hline
model & lower & Rating & upper & 95\% CI & rating & ROC-AUC & BCE Loss & Avg. Error & Average Precision & Pearson R & Spearman R & $R^2$ & MSE & MAE \\
\midrule
BioT5\_plus & 1054.849 & 1056.357 & 1057.592 & 1.235 & 1.508 & 1056.360 & 87.359 & 74.907 & 0.335 & 0.471 & 0.654 & 0.660 & 0.408 & 1.057 & 0.744 \\
BioT5 & 1048.641 & 1050.576 & 1053.647 & 3.071 & 1.935 & 1050.580 & 87.371 & 74.631 & 0.334 & 0.472 & 0.654 & 0.658 & 0.409 & 1.057 & 0.744 \\
LlaSMol & 1029.386 & 1033.410 & 1035.777 & 2.367 & 4.024 & 1033.410 & 86.522 & 73.271 & 0.340 & 0.473 & 0.655 & 0.661 & 0.409 & 1.054 & 0.746 \\
TextChemT5 & 1024.225 & 1026.785 & 1029.894 & 3.108 & 2.560 & 1026.790 & 87.074 & 75.295 & 0.341 & 0.475 & 0.652 & 0.657 & 0.407 & 1.057 & 0.746 \\
GPT-4o-Generic & 1025.020 & 1025.770 & 1027.645 & 1.874 & 0.750 & 1025.770 & 85.685 & 72.388 & 0.341 & 0.472 & 0.657 & 0.662 & 0.412 & 1.049 & 0.742 \\
Llama3.1-405B-Generic & 1011.663 & 1014.464 & 1017.015 & 2.551 & 2.801 & 1014.460 & 85.469 & 72.048 & 0.342 & 0.471 & 0.658 & 0.663 & 0.416 & 1.038 & 0.738 \\
MolT5 & 1013.364 & 1014.026 & 1016.049 & 2.023 & 0.662 & 1014.030 & 86.406 & 75.251 & 0.340 & 0.475 & 0.652 & 0.655 & 0.406 & 1.064 & 0.748 \\
\bottomrule
\end{tabular}
}
\caption{Leaderboard Grouped by Model}
\label{tab:model_model_performance}
\end{table*}

\begin{table*}[h!]
\centering
\small
\adjustbox{max width=\textwidth}{
\begin{tabular}{l|HcHHHcccccccccc}
\hline
model & lower & Rating & upper & 95\% CI & rating & ROC-AUC & BCE Loss & Avg. Error & Average Precision & Pearson R & Spearman R & $R^2$ & MSE & MAE \\
\midrule
BioT5\_plus & 1050.916 & 1052.948 & 1055.798 & 2.850 & 2.032 & 1052.950 & 87.359 & 74.907 & 0.335 & 0.471 & 0.654 & 0.660 & 0.408 & 1.057 & 0.744 \\
BioT5 & 1045.149 & 1048.469 & 1051.445 & 2.976 & 3.320 & 1048.470 & 87.371 & 74.631 & 0.334 & 0.472 & 0.654 & 0.658 & 0.409 & 1.057 & 0.744 \\
LlaSMol & 1028.254 & 1029.885 & 1031.194 & 1.309 & 1.632 & 1029.890 & 86.522 & 73.271 & 0.340 & 0.473 & 0.655 & 0.661 & 0.409 & 1.054 & 0.746 \\
TextChemT5 & 1020.973 & 1023.517 & 1025.285 & 1.768 & 2.544 & 1023.520 & 87.074 & 75.295 & 0.341 & 0.475 & 0.652 & 0.657 & 0.407 & 1.057 & 0.746 \\
GPT-4o-Generic & 1010.372 & 1012.118 & 1013.963 & 1.845 & 1.746 & 1012.120 & 85.597 & 72.299 & 0.342 & 0.473 & 0.656 & 0.660 & 0.410 & 1.051 & 0.743 \\
MolT5 & 1008.340 & 1011.191 & 1014.438 & 3.247 & 2.851 & 1011.190 & 86.406 & 75.251 & 0.340 & 0.475 & 0.652 & 0.655 & 0.406 & 1.064 & 0.748 \\
Llama3.1-405B-Generic & 1000.625 & 1003.365 & 1004.924 & 1.560 & 2.739 & 1003.360 & 85.476 & 72.365 & 0.343 & 0.474 & 0.655 & 0.660 & 0.410 & 1.051 & 0.743 \\
\bottomrule
\end{tabular}
}
\caption{Leaderboard Grouped by Persona}
\label{tab:prompt_model_performance}
\end{table*}

\begin{table*}[h!]
\centering
\small
\adjustbox{max width=\textwidth}{
\begin{tabular}{l|HcHHHcccccccccc}
\hline
model & lower & Rating & upper & 95\% CI & rating & ROC-AUC & BCE Loss & Avg. Error & Average Precision & Pearson R & Spearman R & $R^2$ & MSE & MAE \\
\midrule
BioT5\_plus & 1051.041 & 1053.906 & 1056.527 & 2.621 & 2.865 & 1053.910 & 87.359 & 74.907 & 0.335 & 0.471 & 0.654 & 0.660 & 0.408 & 1.057 & 0.744 \\
BioT5 & 1047.544 & 1050.456 & 1053.271 & 2.815 & 2.912 & 1050.460 & 87.371 & 74.631 & 0.334 & 0.472 & 0.654 & 0.658 & 0.409 & 1.057 & 0.744 \\
LlaSMol & 1030.592 & 1032.487 & 1034.214 & 1.727 & 1.895 & 1032.490 & 86.522 & 73.271 & 0.340 & 0.473 & 0.655 & 0.661 & 0.409 & 1.054 & 0.746 \\
TextChemT5 & 1022.554 & 1025.644 & 1028.227 & 2.583 & 3.091 & 1025.640 & 87.074 & 75.295 & 0.341 & 0.475 & 0.652 & 0.657 & 0.407 & 1.057 & 0.746 \\
GPT-4o & 1011.677 & 1014.434 & 1016.854 & 2.420 & 2.757 & 1014.430 & 85.597 & 72.299 & 0.342 & 0.473 & 0.656 & 0.660 & 0.410 & 1.051 & 0.743 \\
MolT5 & 1010.100 & 1012.590 & 1015.073 & 2.484 & 2.490 & 1012.590 & 86.406 & 75.251 & 0.340 & 0.475 & 0.652 & 0.655 & 0.406 & 1.064 & 0.748 \\
Llama3.1-405B & 1003.000 & 1004.116 & 1006.164 & 2.048 & 1.116 & 1004.120 & 85.476 & 72.365 & 0.343 & 0.474 & 0.655 & 0.660 & 0.410 & 1.051 & 0.743 \\
\bottomrule
\end{tabular}
}
\caption{Leaderboard Grouped by Persona and Fragment}
\label{tab:prompt_frag_model_performance}
\end{table*}

\begin{table*}[h!]
\centering
\small
\adjustbox{max width=\textwidth}{
\begin{tabular}{l|HcHHHcccccccccc}
\hline
model & lower & Rating & upper & 95\% CI & rating & ROC-AUC & BCE Loss & Avg. Error & Average Precision & Pearson R & Spearman R & $R^2$ & MSE & MAE \\
\midrule
BioT5\_plus & 1047.737 & 1052.310 & 1053.674 & 1.364 & 4.573 & 1052.310 & 87.359 & 74.907 & 0.335 & 0.471 & 0.654 & 0.660 & 0.408 & 1.057 & 0.744 \\
BioT5 & 1044.773 & 1046.761 & 1051.139 & 4.378 & 1.988 & 1046.760 & 87.371 & 74.631 & 0.334 & 0.472 & 0.654 & 0.658 & 0.409 & 1.057 & 0.744 \\
LlaSMol & 1027.506 & 1029.626 & 1031.025 & 1.399 & 2.120 & 1029.630 & 86.522 & 73.271 & 0.340 & 0.473 & 0.655 & 0.661 & 0.409 & 1.054 & 0.746 \\
TextChemT5 & 1020.586 & 1022.486 & 1024.672 & 2.186 & 1.900 & 1022.490 & 87.074 & 75.295 & 0.341 & 0.475 & 0.652 & 0.657 & 0.407 & 1.057 & 0.746 \\
MolT5 & 1006.122 & 1009.902 & 1012.877 & 2.976 & 3.779 & 1009.900 & 86.406 & 75.251 & 0.340 & 0.475 & 0.652 & 0.655 & 0.406 & 1.064 & 0.748 \\
General LLM Frontier & 1005.981 & 1006.674 & 1007.723 & 1.050 & 0.693 & 1006.670 & 85.537 & 72.332 & 0.343 & 0.473 & 0.656 & 0.660 & 0.410 & 1.051 & 0.743 \\
Llama3-8B-Task & 995.149 & 997.858 & 1000.061 & 2.203 & 2.709 & 997.860 & 85.414 & 72.182 & 0.344 & 0.477 & 0.652 & 0.656 & 0.406 & 1.063 & 0.748 \\
\bottomrule
\end{tabular}
}
\caption{Leaderboard Grouped by Model Size. Note that Medium only includes Gemma2-27B.}
\label{tab:size_model_performance}
\end{table*}

\newpage
\clearpage

\section{Specific Dataset Benchmarks}

\begin{table*}[h!]
\centering
\tiny
\adjustbox{max width=\textwidth}{
\begin{tabular}{l|HHHHHcccccHHHHH}
\hline
model & lower & Rating & upper & error\_y & error\_y\_minus & Rating & ROC-AUC & BCE Loss & Avg. Error & Average Precision & Pearson R & Spearman R & $R^2$ & MSE & MAE \\
\midrule
BioT5 & 1183.39 & 1189.60 & 1198.00 & 8.40 & 6.22 & 1189.60 & 88.68 & 0.33 & 0.19 & 95.38 & [] & [] & [] & [] & [] \\
BioT5\_plus & 1159.96 & 1164.96 & 1173.87 & 8.91 & 4.99 & 1164.96 & 89.29 & 0.32 & 0.18 & 95.77 & [] & [] & [] & [] & [] \\
Gemma2-9B-Frags-Chem & 1099.56 & 1104.62 & 1108.72 & 4.09 & 5.06 & 1104.62 & 88.82 & 0.33 & 0.19 & 95.66 & [] & [] & [] & [] & [] \\
Gemma2-9B-Frags-Generic & 1073.45 & 1080.14 & 1089.09 & 8.96 & 6.69 & 1080.14 & 88.59 & 0.33 & 0.19 & 95.43 & [] & [] & [] & [] & [] \\
LlaSMol & 1069.29 & 1073.04 & 1078.16 & 5.12 & 3.75 & 1073.04 & 88.17 & 0.33 & 0.19 & 95.41 & [] & [] & [] & [] & [] \\
Llama3.1-70B-Chem & 1064.97 & 1072.72 & 1076.63 & 3.91 & 7.75 & 1072.72 & 88.29 & 0.33 & 0.19 & 95.29 & [] & [] & [] & [] & [] \\
TextChemT5 & 1064.45 & 1070.99 & 1075.36 & 4.37 & 6.54 & 1070.99 & 88.10 & 0.33 & 0.19 & 95.27 & [] & [] & [] & [] & [] \\
\hline
\end{tabular}
}
\caption{BBBP Leaderboard}
\label{tab:bbbp_model_performance}
\end{table*}

\newpage
\clearpage
\begin{table*}[h!]
\centering
\tiny
\adjustbox{max width=\textwidth}{
\begin{tabular}{l|HHHHHcccccHHHHH}
\hline
model & lower & Rating & upper & error\_y & error\_y\_minus & Rating & ROC-AUC & BCE Loss & Avg. Error & Average Precision & Pearson R & Spearman R & $R^2$ & MSE & MAE \\
\midrule
Gemma2-9B-Frags-Drug & 1093.93 & 1099.31 & 1109.01 & 9.70 & 5.38 & 1099.31 & 85.37 & 0.49 & 0.31 & 82.31 & [] & [] & [] & [] & [] \\
LlaSMol & 1040.27 & 1047.54 & 1056.16 & 8.62 & 7.27 & 1047.54 & 85.00 & 0.50 & 0.31 & 81.32 & [] & [] & [] & [] & [] \\
BioT5\_plus & 1035.04 & 1038.29 & 1041.58 & 3.29 & 3.25 & 1038.29 & 84.94 & 0.50 & 0.31 & 81.35 & [] & [] & [] & [] & [] \\
MistralNeMo-12B-Bio & 1030.22 & 1035.63 & 1042.62 & 6.99 & 5.41 & 1035.63 & 84.81 & 0.50 & 0.32 & 81.61 & [] & [] & [] & [] & [] \\
Gemma2-9B-Frags-Generic & 1025.86 & 1030.70 & 1037.96 & 7.27 & 4.84 & 1030.70 & 84.80 & 0.49 & 0.32 & 81.41 & [] & [] & [] & [] & [] \\
Llama3.1-8B-Quant & 1013.17 & 1022.96 & 1031.71 & 8.75 & 9.79 & 1022.96 & 84.74 & 0.50 & 0.31 & 81.37 & [] & [] & [] & [] & [] \\
Llama3.1-70B-Drug & 1017.49 & 1022.08 & 1029.47 & 7.39 & 4.59 & 1022.08 & 84.79 & 0.49 & 0.32 & 81.69 & [] & [] & [] & [] & [] \\
\hline
\end{tabular}
}
\caption{BACE Model Performance Metrics}
\label{tab:bace_model_performance}
\end{table*}

\newpage
\clearpage
\begin{table*}[h!]
\centering
\tiny
\adjustbox{max width=\textwidth}{
\begin{tabular}{l|HHHHHcccccHHHHH}
\hline
model & lower & Rating & upper & error\_y & error\_y\_minus & Rating & ROC-AUC & BCE Loss & Avg. Error & Average Precision & Pearson R & Spearman R & $R^2$ & MSE & MAE \\
\midrule
TextChemT5 & 1137.37 & 1144.91 & 1152.90 & 8.00 & 7.54 & 1144.91 & 88.28 & 0.19 & 0.10 & 49.33 & [] & [] & [] & [] & [] \\
BioT5\_plus & 1131.93 & 1137.96 & 1147.49 & 9.54 & 6.03 & 1137.96 & 87.84 & 0.19 & 0.10 & 47.60 & [] & [] & [] & [] & [] \\
BioT5 & 1100.48 & 1113.64 & 1115.52 & 1.88 & 13.16 & 1113.64 & 88.74 & 0.18 & 0.10 & 47.24 & [] & [] & [] & [] & [] \\
LlaSMol & 1075.68 & 1080.82 & 1087.12 & 6.30 & 5.14 & 1080.82 & 86.40 & 0.19 & 0.10 & 43.08 & [] & [] & [] & [] & [] \\
MolInstructions\_molecule & 1048.72 & 1059.32 & 1063.74 & 4.43 & 10.59 & 1059.32 & 84.15 & 0.20 & 0.11 & 40.89 & [] & [] & [] & [] & [] \\
MolT5 & 1055.39 & 1059.26 & 1069.75 & 10.49 & 3.87 & 1059.26 & 86.38 & 0.19 & 0.10 & 49.12 & [] & [] & [] & [] & [] \\
Llama3-70B-Frags-Generic & 1043.69 & 1056.45 & 1059.28 & 2.83 & 12.76 & 1056.45 & 84.39 & 0.20 & 0.10 & 41.38 & [] & [] & [] & [] & [] \\
\hline
\end{tabular}
}
\caption{ClinTox Leaderboard}
\label{tab:clintox_model_performance}
\end{table*}

\newpage
\clearpage
\begin{table*}[h!]
\centering
\tiny
\adjustbox{max width=\textwidth}{
\begin{tabular}{l|HHHHHcHHcHccccc}
\hline
model & lower & Rating & upper & error\_y & error\_y\_minus & Rating & ROC-AUC & BCE Loss & Avg. Error & Average Precision & Pearson R & Spearman R & $R^2$ & MSE & MAE \\
\midrule
GPT-4o-Generic & 1072.223 & 1077.867 & 1091.582 & 13.715 & 5.645 & 1077.870 & [] & [] & 0.790 & [] & 0.325 & 0.360 & -0.074 & 1.297 & 0.790 \\
Llama3.1-405B-Generic & 1059.000 & 1062.397 & 1071.598 & 9.202 & 3.397 & 1062.400 & [] & [] & 0.790 & [] & 0.324 & 0.361 & -0.072 & 1.296 & 0.790 \\
Gemma2-9B-Chem & 1051.990 & 1058.902 & 1062.669 & 3.767 & 6.912 & 1058.900 & [] & [] & 0.792 & [] & 0.317 & 0.352 & -0.076 & 1.301 & 0.792 \\
BioT5 & 1048.084 & 1056.420 & 1059.834 & 3.414 & 8.336 & 1056.420 & [] & [] & 0.791 & [] & 0.319 & 0.354 & -0.078 & 1.301 & 0.791 \\
BioT5\_plus & 1040.573 & 1048.617 & 1052.589 & 3.973 & 8.044 & 1048.620 & [] & [] & 0.793 & [] & 0.318 & 0.355 & -0.081 & 1.305 & 0.793 \\
Llama3-70B-Drug & 1040.547 & 1045.240 & 1049.483 & 4.243 & 4.692 & 1045.240 & [] & [] & 0.792 & [] & 0.320 & 0.355 & -0.074 & 1.298 & 0.792 \\
Llama3-8B-Frags-Generic & 1038.821 & 1043.637 & 1049.508 & 5.871 & 4.816 & 1043.640 & [] & [] & 0.794 & [] & 0.316 & 0.356 & -0.082 & 1.305 & 0.794 \\
\hline
\end{tabular}
}
\caption{FreeSolv Leaderboard}
\label{tab:freesolv_model_performance}
\end{table*}

\newpage
\clearpage
\begin{table*}[h!]
\centering
\tiny
\adjustbox{max width=\textwidth}{
\begin{tabular}{l|HHHHHcHHcHccccc}
\hline
model & lower & Rating & upper & error\_y & error\_y\_minus & Rating & ROC-AUC & BCE Loss & Avg. Error & Average Precision & Pearson R & Spearman R & $R^2$ & MSE & MAE \\
\midrule
Gemma2-27B-Generic & 1033.220 & 1038.013 & 1044.431 & 6.419 & 4.792 & 1038.010 & [] & [] & 0.882 & [] & 0.819 & 0.815 & 0.637 & 1.328 & 0.882 \\
MistralNeMo-12B-Frags-Drug & 1030.413 & 1037.158 & 1043.010 & 5.852 & 6.745 & 1037.160 & [] & [] & 0.888 & [] & 0.819 & 0.810 & 0.635 & 1.333 & 0.888 \\
Llama3.1-405B-Generic & 1027.947 & 1034.548 & 1039.399 & 4.851 & 6.601 & 1034.550 & [] & [] & 0.885 & [] & 0.821 & 0.810 & 0.638 & 1.322 & 0.885 \\
Llama3.1-70B-Drug & 1023.158 & 1030.198 & 1032.348 & 2.150 & 7.040 & 1030.200 & [] & [] & 0.886 & [] & 0.821 & 0.809 & 0.638 & 1.326 & 0.886 \\
Llama3.1-8B-Chem & 1024.447 & 1028.275 & 1033.865 & 5.590 & 3.827 & 1028.270 & [] & [] & 0.901 & [] & 0.814 & 0.804 & 0.626 & 1.369 & 0.901 \\
GPT-4o-Generic & 1021.733 & 1028.082 & 1032.197 & 4.115 & 6.350 & 1028.080 & [] & [] & 0.894 & [] & 0.817 & 0.808 & 0.630 & 1.353 & 0.894 \\
Gemma2-9B-Frags-Quant & 1021.322 & 1027.687 & 1037.590 & 9.903 & 6.365 & 1027.690 & [] & [] & 0.902 & [] & 0.813 & 0.802 & 0.624 & 1.374 & 0.902 \\
\hline
\end{tabular}
}
\caption{ESOL Leaderboard}
\label{tab:esol_model_performance}
\end{table*}

\newpage
\clearpage
\begin{table*}[h!]
\centering
\tiny
\adjustbox{max width=\textwidth}{
\begin{tabular}{l|HHHHHcHHcHccccc}
\hline
model & lower & Rating & upper & error\_y & error\_y\_minus & Rating & ROC-AUC & BCE Loss & Avg. Error & Average Precision & Pearson R & Spearman R & $R^2$ & MSE & MAE \\
\midrule
MistralNeMo-12B-Frags-Drug & 1009.341 & 1014.057 & 1017.395 & 3.338 & 4.717 & 1014.060 & [] & [] & 0.539 & [] & 0.830 & 0.819 & 0.681 & 0.495 & 0.539 \\
TextChemT5 & 1006.502 & 1013.871 & 1018.904 & 5.033 & 7.369 & 1013.870 & [] & [] & 0.540 & [] & 0.830 & 0.819 & 0.681 & 0.495 & 0.540 \\
Llama3-70B-Generic & 999.710 & 1012.213 & 1016.913 & 4.700 & 12.503 & 1012.210 & [] & [] & 0.539 & [] & 0.830 & 0.820 & 0.681 & 0.495 & 0.539 \\
BioT5 & 1002.621 & 1011.696 & 1016.166 & 4.471 & 9.075 & 1011.700 & [] & [] & 0.540 & [] & 0.830 & 0.819 & 0.681 & 0.496 & 0.540 \\
Gemma2-27B-Generic & 1007.490 & 1011.212 & 1017.027 & 5.815 & 3.722 & 1011.210 & [] & [] & 0.538 & [] & 0.832 & 0.821 & 0.684 & 0.491 & 0.538 \\
MistralNeMo-12B-Bio & 1002.481 & 1011.154 & 1014.392 & 3.238 & 8.673 & 1011.150 & [] & [] & 0.539 & [] & 0.830 & 0.820 & 0.681 & 0.495 & 0.539 \\
Gemma2-9B-Frags-Drug & 1004.839 & 1009.968 & 1016.931 & 6.963 & 5.129 & 1009.970 & [] & [] & 0.536 & [] & 0.832 & 0.821 & 0.684 & 0.491 & 0.536 \\
\hline
\end{tabular}
}
\caption{Lipo Leaderboard}
\label{tab:lipo_model_performance}
\end{table*}

\newpage
\clearpage

\twocolumn
\section{Testing with ``Oracle'' captions.}
\label{appendix:oracle}

We also consider two theoretical caption sources: the ``BlankCaption'', which contains no content, and a ``BBBP Oracle'', which states the BBBP ground truth. Essentially, a molecule is given the caption True if it can penetrate the BBB, and False otherwise. We find that this achieves perfect results and easily beats all other caption sources, as expected. Results show that BlankCaption doesn't improve over the GNN, as expected. Certain models, especially Llama3.1-8B, show worse ratings, where the captions actively cause performance due to misinformation.

\onecolumn
\begin{table*}[h!]
\centering
\tiny
\adjustbox{max width=\textwidth}{
\begin{tabular}{l|HHHHHcccccHHHHH}
\hline
model & lower & Rating & upper & error\_y & error\_y\_minus & Rating & ROC-AUC & BCE Loss & Avg. Error & Average Precision & Pearson R & Spearman R & $R^2$ & MSE & MAE \\
\midrule
BBBP Oracle & 3541.90 & 3769.78 & 3946.53 & 176.76 & 227.88 & 3769.78 & 100.00 & 0.01 & 0.01 & 100.00 & [] & [] & [] & [] & [] \\
BioT5 & 1148.75 & 1157.70 & 1163.83 & 6.12 & 8.95 & 1157.70 & 88.68 & 0.33 & 0.19 & 95.38 & [] & [] & [] & [] & [] \\
BioT5\_plus & 1128.78 & 1134.59 & 1143.37 & 8.78 & 5.81 & 1134.59 & 89.29 & 0.32 & 0.18 & 95.77 & [] & [] & [] & [] & [] \\
Gemma2-9B-Frags-Chem & 1069.87 & 1072.80 & 1080.75 & 7.96 & 2.92 & 1072.80 & 88.82 & 0.33 & 0.19 & 95.66 & [] & [] & [] & [] & [] \\
Gemma2-9B-Frags-Generic & 1043.34 & 1047.95 & 1053.96 & 6.01 & 4.61 & 1047.95 & 88.59 & 0.33 & 0.19 & 95.43 & [] & [] & [] & [] & [] \\
LlaSMol & 1033.86 & 1041.02 & 1047.83 & 6.82 & 7.16 & 1041.02 & 88.17 & 0.33 & 0.19 & 95.41 & [] & [] & [] & [] & [] \\
Llama3.1-70B-Chem & 1035.33 & 1039.52 & 1047.65 & 8.13 & 4.19 & 1039.52 & 88.29 & 0.33 & 0.19 & 95.29 & [] & [] & [] & [] & [] \\
TextChemT5 & 1033.96 & 1038.46 & 1043.07 & 4.62 & 4.50 & 1038.46 & 88.10 & 0.33 & 0.19 & 95.27 & [] & [] & [] & [] & [] \\
Gemma2-9B-Frags-Drug & 1028.91 & 1038.37 & 1042.44 & 4.06 & 9.46 & 1038.37 & 88.11 & 0.33 & 0.19 & 95.43 & [] & [] & [] & [] & [] \\
GPT-4o-Generic & 1029.13 & 1037.28 & 1044.11 & 6.83 & 8.15 & 1037.28 & 88.68 & 0.33 & 0.19 & 95.55 & [] & [] & [] & [] & [] \\
MistralNeMo-12B-Frags-Generic & 1026.95 & 1034.10 & 1040.01 & 5.91 & 7.15 & 1034.10 & 88.74 & 0.33 & 0.19 & 95.50 & [] & [] & [] & [] & [] \\
Llama3.1-70B-Generic & 1023.43 & 1030.20 & 1038.50 & 8.30 & 6.77 & 1030.20 & 88.56 & 0.33 & 0.19 & 95.46 & [] & [] & [] & [] & [] \\
GPT-4o-Frags-Generic & 1011.42 & 1024.75 & 1029.80 & 5.05 & 13.34 & 1024.75 & 88.63 & 0.33 & 0.19 & 95.64 & [] & [] & [] & [] & [] \\
Gemma2-27B-Generic & 1015.03 & 1021.62 & 1032.32 & 10.71 & 6.59 & 1021.62 & 88.28 & 0.33 & 0.19 & 95.42 & [] & [] & [] & [] & [] \\
MistralNeMo-12B-Frags-Chem & 1010.77 & 1019.68 & 1025.03 & 5.35 & 8.91 & 1019.68 & 88.22 & 0.33 & 0.19 & 95.26 & [] & [] & [] & [] & [] \\
Llama3-70B-Chem & 1010.54 & 1018.65 & 1021.91 & 3.26 & 8.11 & 1018.65 & 88.72 & 0.33 & 0.19 & 95.65 & [] & [] & [] & [] & [] \\
Llama3-70B-Drug & 1010.49 & 1018.58 & 1027.01 & 8.43 & 8.09 & 1018.58 & 88.24 & 0.33 & 0.19 & 95.32 & [] & [] & [] & [] & [] \\
Gemma2-27B-Chem & 1010.80 & 1014.91 & 1022.65 & 7.74 & 4.11 & 1014.91 & 88.34 & 0.34 & 0.19 & 95.34 & [] & [] & [] & [] & [] \\
Llama3.1-70B-Bio & 1011.48 & 1014.71 & 1022.25 & 7.54 & 3.23 & 1014.71 & 88.07 & 0.34 & 0.19 & 95.24 & [] & [] & [] & [] & [] \\
Llama3-70B-Generic & 1008.61 & 1011.00 & 1018.50 & 7.49 & 2.39 & 1011.00 & 88.45 & 0.33 & 0.19 & 95.44 & [] & [] & [] & [] & [] \\
Llama3.1-405B-Generic & 1006.82 & 1010.83 & 1021.50 & 10.67 & 4.01 & 1010.83 & 88.58 & 0.33 & 0.19 & 95.48 & [] & [] & [] & [] & [] \\
Gemma2-27B-Drug & 1000.68 & 1008.53 & 1016.22 & 7.69 & 7.84 & 1008.53 & 88.32 & 0.33 & 0.19 & 95.43 & [] & [] & [] & [] & [] \\
Llama3.1-70B-Drug & 1001.69 & 1007.57 & 1016.12 & 8.55 & 5.88 & 1007.57 & 88.56 & 0.33 & 0.19 & 95.52 & [] & [] & [] & [] & [] \\
ChemDFM & 998.13 & 1002.34 & 1012.06 & 9.72 & 4.20 & 1002.34 & 88.91 & 0.33 & 0.19 & 95.68 & [] & [] & [] & [] & [] \\
Gemma2-27B-Quant & 991.23 & 1000.42 & 1005.97 & 5.55 & 9.19 & 1000.42 & 88.49 & 0.33 & 0.19 & 95.47 & [] & [] & [] & [] & [] \\
Llama3-70B-Bio & 992.98 & 998.75 & 1009.61 & 10.86 & 5.77 & 998.75 & 88.54 & 0.33 & 0.19 & 95.56 & [] & [] & [] & [] & [] \\
Llama3.1-70B-Quant & 993.28 & 997.34 & 1004.83 & 7.49 & 4.06 & 997.34 & 88.37 & 0.33 & 0.19 & 95.35 & [] & [] & [] & [] & [] \\
MolT5 & 991.55 & 995.85 & 1005.38 & 9.53 & 4.30 & 995.85 & 88.13 & 0.33 & 0.19 & 95.35 & [] & [] & [] & [] & [] \\
Llama3.1-8B-Drug & 982.58 & 991.11 & 1001.70 & 10.59 & 8.53 & 991.11 & 88.39 & 0.33 & 0.19 & 95.50 & [] & [] & [] & [] & [] \\
Gemma2-27B-Bio & 981.25 & 988.04 & 996.86 & 8.81 & 6.79 & 988.04 & 88.34 & 0.33 & 0.19 & 95.45 & [] & [] & [] & [] & [] \\
Gemma2-27B-Frags-Drug & 980.63 & 986.19 & 993.07 & 6.89 & 5.55 & 986.19 & 88.31 & 0.33 & 0.19 & 95.32 & [] & [] & [] & [] & [] \\
Gemma2-9B-Chem & 974.41 & 982.57 & 993.13 & 10.56 & 8.17 & 982.57 & 88.57 & 0.33 & 0.19 & 95.52 & [] & [] & [] & [] & [] \\
Gemma2-27B-Frags-Bio & 971.34 & 978.56 & 986.42 & 7.86 & 7.22 & 978.56 & 88.49 & 0.33 & 0.19 & 95.50 & [] & [] & [] & [] & [] \\
Gemma2-27B-Frags-Chem & 969.19 & 975.88 & 985.36 & 9.48 & 6.69 & 975.88 & 88.38 & 0.33 & 0.19 & 95.43 & [] & [] & [] & [] & [] \\
Gemma2-9B-Frags-Quant & 966.63 & 975.65 & 985.06 & 9.41 & 9.02 & 975.65 & 88.42 & 0.33 & 0.19 & 95.43 & [] & [] & [] & [] & [] \\
Llama3-8B-Task & 969.64 & 975.48 & 983.70 & 8.22 & 5.84 & 975.48 & 88.28 & 0.33 & 0.19 & 95.40 & [] & [] & [] & [] & [] \\
MistralNeMo-12B-Frags-Drug & 961.92 & 967.26 & 980.51 & 13.25 & 5.34 & 967.26 & 88.48 & 0.33 & 0.19 & 95.45 & [] & [] & [] & [] & [] \\
Llama3-70B-Frags-Generic & 959.27 & 966.93 & 974.17 & 7.24 & 7.66 & 966.93 & 88.23 & 0.33 & 0.19 & 95.39 & [] & [] & [] & [] & [] \\
MistralNeMo-12B-Bio & 958.10 & 966.36 & 970.95 & 4.59 & 8.26 & 966.36 & 88.38 & 0.33 & 0.19 & 95.47 & [] & [] & [] & [] & [] \\
MistralNeMo-12B-Generic & 958.89 & 965.71 & 971.71 & 6.00 & 6.83 & 965.71 & 88.44 & 0.33 & 0.19 & 95.44 & [] & [] & [] & [] & [] \\
MistralNeMo-12B-Frags-Bio & 951.30 & 961.22 & 968.12 & 6.90 & 9.92 & 961.22 & 88.30 & 0.33 & 0.19 & 95.41 & [] & [] & [] & [] & [] \\
MistralNeMo-12B-Quant & 954.18 & 960.12 & 970.84 & 10.71 & 5.95 & 960.12 & 88.27 & 0.33 & 0.19 & 95.41 & [] & [] & [] & [] & [] \\
MistralNeMo-12B-Chem & 948.69 & 959.82 & 967.56 & 7.74 & 11.12 & 959.82 & 88.37 & 0.33 & 0.19 & 95.47 & [] & [] & [] & [] & [] \\
Gemma2-9B-Drug & 955.22 & 958.72 & 969.79 & 11.06 & 3.50 & 958.72 & 88.23 & 0.33 & 0.19 & 95.37 & [] & [] & [] & [] & [] \\
Llama3-8B-Chem & 952.55 & 958.63 & 966.47 & 7.84 & 6.08 & 958.63 & 88.48 & 0.33 & 0.19 & 95.47 & [] & [] & [] & [] & [] \\
Llama3-8B-Drug & 945.11 & 957.06 & 963.17 & 6.11 & 11.95 & 957.06 & 88.40 & 0.33 & 0.19 & 95.47 & [] & [] & [] & [] & [] \\
Gemma2-9B-Frags-Bio & 953.52 & 956.24 & 962.70 & 6.46 & 2.72 & 956.24 & 88.13 & 0.33 & 0.19 & 95.31 & [] & [] & [] & [] & [] \\
3D-MoLM & 947.99 & 956.09 & 963.65 & 7.56 & 8.11 & 956.09 & 88.22 & 0.33 & 0.19 & 95.41 & [] & [] & [] & [] & [] \\
Llama3.1-70B-Frags-Generic & 953.82 & 956.05 & 966.39 & 10.34 & 2.23 & 956.05 & 88.30 & 0.33 & 0.19 & 95.42 & [] & [] & [] & [] & [] \\
Llama3.1-70B-Frags-Chem & 944.18 & 953.21 & 959.31 & 6.10 & 9.03 & 953.21 & 88.33 & 0.33 & 0.19 & 95.41 & [] & [] & [] & [] & [] \\
Llama3-70B-Quant & 942.44 & 952.52 & 956.93 & 4.41 & 10.09 & 952.52 & 88.20 & 0.33 & 0.19 & 95.35 & [] & [] & [] & [] & [] \\
Gemma2-27B-Frags-Quant & 941.80 & 951.48 & 958.95 & 7.47 & 9.68 & 951.48 & 88.55 & 0.33 & 0.19 & 95.48 & [] & [] & [] & [] & [] \\
Gemma2-9B-Generic & 944.39 & 951.03 & 954.96 & 3.92 & 6.64 & 951.03 & 88.30 & 0.33 & 0.19 & 95.40 & [] & [] & [] & [] & [] \\
Llama3.1-405B-Frags-Generic & 944.43 & 949.76 & 956.65 & 6.89 & 5.33 & 949.76 & 88.26 & 0.33 & 0.19 & 95.35 & [] & [] & [] & [] & [] \\
Gemma2-27B-Frags-Generic & 936.10 & 947.31 & 951.63 & 4.32 & 11.21 & 947.31 & 88.22 & 0.33 & 0.19 & 95.31 & [] & [] & [] & [] & [] \\
Llama3.1-70B-Frags-Quant & 937.31 & 946.91 & 951.34 & 4.43 & 9.60 & 946.91 & 88.28 & 0.33 & 0.19 & 95.39 & [] & [] & [] & [] & [] \\
Llama3.1-70B-Frags-Bio & 941.12 & 946.34 & 952.73 & 6.39 & 5.22 & 946.34 & 88.25 & 0.33 & 0.19 & 95.41 & [] & [] & [] & [] & [] \\
Llama3.1-70B-Frags-Drug & 927.96 & 934.60 & 940.47 & 5.86 & 6.64 & 934.60 & 88.31 & 0.33 & 0.19 & 95.43 & [] & [] & [] & [] & [] \\
Gemma2-9B-Bio & 928.68 & 934.04 & 939.83 & 5.79 & 5.36 & 934.04 & 88.23 & 0.33 & 0.19 & 95.38 & [] & [] & [] & [] & [] \\
Llama3.1-8B-Quant & 922.33 & 931.97 & 936.40 & 4.42 & 9.64 & 931.97 & 88.51 & 0.33 & 0.19 & 95.54 & [] & [] & [] & [] & [] \\
Llama3-8B-Frags-Drug & 921.32 & 929.72 & 940.30 & 10.57 & 8.40 & 929.72 & 88.26 & 0.33 & 0.19 & 95.41 & [] & [] & [] & [] & [] \\
Llama3-8B-Frags-Bio & 920.09 & 928.48 & 933.69 & 5.21 & 8.39 & 928.48 & 88.28 & 0.33 & 0.19 & 95.41 & [] & [] & [] & [] & [] \\
Llama3-70B-Frags-Quant & 925.28 & 928.47 & 935.98 & 7.52 & 3.19 & 928.47 & 88.29 & 0.33 & 0.19 & 95.42 & [] & [] & [] & [] & [] \\
Llama3-8B-Generic & 917.73 & 927.56 & 936.23 & 8.67 & 9.83 & 927.56 & 88.43 & 0.33 & 0.19 & 95.48 & [] & [] & [] & [] & [] \\
Llama3-70B-Frags-Bio & 920.59 & 927.24 & 938.50 & 11.27 & 6.65 & 927.24 & 88.37 & 0.33 & 0.19 & 95.44 & [] & [] & [] & [] & [] \\
MistralNeMo-12B-Frags-Quant & 919.44 & 926.87 & 933.21 & 6.34 & 7.43 & 926.87 & 88.37 & 0.33 & 0.19 & 95.49 & [] & [] & [] & [] & [] \\
Llama3-8B-Quant & 917.59 & 922.94 & 927.35 & 4.41 & 5.34 & 922.94 & 88.37 & 0.33 & 0.19 & 95.43 & [] & [] & [] & [] & [] \\
Llama3-70B-Frags-Drug & 915.21 & 922.64 & 931.35 & 8.72 & 7.42 & 922.64 & 88.15 & 0.33 & 0.19 & 95.31 & [] & [] & [] & [] & [] \\
MolT5\_LPM24 & 918.09 & 922.32 & 930.24 & 7.91 & 4.23 & 922.32 & 88.29 & 0.33 & 0.19 & 95.42 & [] & [] & [] & [] & [] \\
Llama3-8B-Frags-Chem & 911.26 & 918.85 & 922.01 & 3.16 & 7.59 & 918.85 & 88.27 & 0.33 & 0.19 & 95.42 & [] & [] & [] & [] & [] \\
Gemma2-9B-Quant & 906.36 & 916.59 & 919.86 & 3.28 & 10.23 & 916.59 & 88.26 & 0.33 & 0.19 & 95.39 & [] & [] & [] & [] & [] \\
ChemLLM & 903.32 & 916.09 & 920.70 & 4.62 & 12.77 & 916.09 & 88.33 & 0.33 & 0.19 & 95.45 & [] & [] & [] & [] & [] \\
Llama3.1-8B-Generic & 910.16 & 914.88 & 928.09 & 13.21 & 4.73 & 914.88 & 88.27 & 0.34 & 0.19 & 95.46 & [] & [] & [] & [] & [] \\
Llama3.1-8B-Chem & 904.30 & 914.88 & 925.16 & 10.27 & 10.59 & 914.88 & 88.50 & 0.33 & 0.19 & 95.53 & [] & [] & [] & [] & [] \\
Llama3.1-8B-Frags-Bio & 910.11 & 913.37 & 918.70 & 5.33 & 3.26 & 913.37 & 88.30 & 0.33 & 0.19 & 95.43 & [] & [] & [] & [] & [] \\
MistralNeMo-12B-Drug & 902.91 & 912.37 & 920.97 & 8.60 & 9.46 & 912.37 & 88.24 & 0.33 & 0.19 & 95.36 & [] & [] & [] & [] & [] \\
MolInstructions\_molecule & 907.46 & 911.98 & 918.35 & 6.37 & 4.52 & 911.98 & 88.27 & 0.33 & 0.19 & 95.42 & [] & [] & [] & [] & [] \\
Llama3-8B-Bio & 907.97 & 911.45 & 916.16 & 4.71 & 3.47 & 911.45 & 88.14 & 0.33 & 0.19 & 95.36 & [] & [] & [] & [] & [] \\
BlankCaption & 903.25 & 910.43 & 919.89 & 9.45 & 7.19 & 910.43 & 88.27 & 0.33 & 0.19 & 95.41 & [] & [] & [] & [] & [] \\
Llama3-8B-Frags-Quant & 905.60 & 908.92 & 913.90 & 4.98 & 3.32 & 908.92 & 88.43 & 0.33 & 0.19 & 95.52 & [] & [] & [] & [] & [] \\
Llama3.1-8B-Frags-Quant & 901.39 & 905.37 & 912.73 & 7.36 & 3.98 & 905.37 & 88.22 & 0.33 & 0.19 & 95.36 & [] & [] & [] & [] & [] \\
Llama3.1-8B-Frags-Drug & 897.85 & 904.38 & 907.72 & 3.34 & 6.53 & 904.38 & 88.32 & 0.33 & 0.19 & 95.43 & [] & [] & [] & [] & [] \\
Llama3.1-8B-Bio & 892.79 & 903.00 & 911.03 & 8.03 & 10.21 & 903.00 & 88.26 & 0.33 & 0.19 & 95.40 & [] & [] & [] & [] & [] \\
Llama3.1-8B-Frags-Chem & 895.96 & 899.23 & 904.89 & 5.65 & 3.27 & 899.23 & 88.27 & 0.33 & 0.19 & 95.40 & [] & [] & [] & [] & [] \\
Llama3-70B-Frags-Chem & 881.35 & 894.18 & 900.65 & 6.48 & 12.83 & 894.18 & 88.35 & 0.33 & 0.19 & 95.45 & [] & [] & [] & [] & [] \\
Llama3-8B-Frags-Generic & 889.04 & 893.49 & 900.57 & 7.09 & 4.44 & 893.49 & 88.26 & 0.33 & 0.19 & 95.39 & [] & [] & [] & [] & [] \\
Llama3.1-8B-Frags-Generic & 886.30 & 892.95 & 902.27 & 9.31 & 6.66 & 892.95 & 88.22 & 0.33 & 0.19 & 95.38 & [] & [] & [] & [] & [] \\
GNN & NaN & NaN & NaN & NaN & NaN & NaN & 88.28 & 0.33 & 0.19 & 95.40 & [] & [] & [] & [] & [] \\
\hline
\end{tabular}
}
\caption{BBBP Leaderboard with Oracle Model}
\label{tab:bbbp_oracle_model_performance}
\end{table*}

\newpage
\clearpage

\section{Correlation Between Classification and Regression}
\label{appendix:classification_vs_regression}

\begin{table*}[h!]
\centering
\small
\adjustbox{max width=\textwidth}{
\begin{tabular}{llHrHHHHrrrr|HrHHHHrrrrrr}
\toprule
 &  & \multicolumn{10}{r}{Classification} & \multicolumn{12}{r}{Regression} \\
 &  & lower & Rating & upper & error\_y & error\_y\_minus & Rating\_rounded & ROC-AUC & BCE Loss & Avg. Error & AP & lower & Rating & upper & error\_y & error\_y\_minus & Rating\_rounded & Avg. Error & Pearson R & Spearman R & $R^2$ & MSE & MAE \\
\midrule
Classification & Rating & 0.996 & 1.000 & 0.995 & 0.075 & -0.044 & 1.000 & 0.434 & -0.299 & -0.338 & 0.450 & 0.467 & 0.464 & 0.469 & 0.116 & -0.053 & 0.464 & -0.422 & 0.424 & 0.397 & 0.469 & -0.463 & -0.422 \\
 & ROC-AUC & 0.436 & 0.434 & 0.451 & 0.296 & -0.007 & 0.434 & 1.000 & -0.706 & -0.445 & 0.770 & 0.181 & 0.172 & 0.151 & -0.155 & -0.077 & 0.172 & -0.111 & 0.077 & 0.074 & 0.128 & -0.110 & -0.111 \\
 & BCE Loss & -0.312 & -0.299 & -0.315 & -0.120 & 0.051 & -0.299 & -0.706 & 1.000 & 0.547 & -0.577 & -0.056 & -0.048 & -0.046 & 0.027 & 0.075 & -0.048 & -0.041 & -0.013 & -0.015 & -0.044 & -0.010 & -0.041 \\
 & Avg. Error & -0.341 & -0.338 & -0.357 & -0.169 & -0.022 & -0.338 & -0.445 & 0.547 & 1.000 & -0.484 & -0.196 & -0.197 & -0.191 & 0.056 & 0.080 & -0.197 & 0.127 & -0.207 & -0.169 & -0.203 & 0.145 & 0.127 \\
 & AP & 0.448 & 0.450 & 0.468 & 0.336 & 0.010 & 0.450 & 0.770 & -0.577 & -0.484 & 1.000 & 0.160 & 0.149 & 0.129 & -0.140 & -0.120 & 0.149 & -0.051 & 0.063 & 0.062 & 0.080 & -0.068 & -0.051 \\
\cline{1-24}
Regression & Rating & 0.458 & 0.464 & 0.465 & 0.020 & -0.093 & 0.464 & 0.172 & -0.048 & -0.197 & 0.149 & 0.993 & 1.000 & 0.993 & 0.030 & -0.032 & 1.000 & -0.752 & 0.908 & 0.866 & 0.863 & -0.755 & -0.752 \\
 & Avg. Error & -0.428 & -0.422 & -0.431 & -0.102 & 0.136 & -0.422 & -0.111 & -0.041 & 0.127 & -0.051 & -0.750 & -0.752 & -0.737 & 0.021 & 0.066 & -0.752 & 1.000 & -0.843 & -0.811 & -0.931 & 0.972 & 1.000 \\
 & Pearson R & 0.422 & 0.424 & 0.428 & 0.013 & -0.089 & 0.424 & 0.077 & -0.013 & -0.207 & 0.063 & 0.903 & 0.908 & 0.902 & 0.034 & -0.036 & 0.908 & -0.843 & 1.000 & 0.950 & 0.950 & -0.854 & -0.843 \\
 & Spearman R & 0.394 & 0.397 & 0.401 & 0.004 & -0.074 & 0.397 & 0.074 & -0.015 & -0.169 & 0.062 & 0.857 & 0.866 & 0.862 & 0.044 & 0.020 & 0.866 & -0.811 & 0.950 & 1.000 & 0.905 & -0.826 & -0.811 \\
 & $R^2$ & 0.471 & 0.469 & 0.475 & 0.035 & -0.099 & 0.469 & 0.128 & -0.044 & -0.203 & 0.080 & 0.861 & 0.863 & 0.857 & 0.049 & -0.076 & 0.863 & -0.931 & 0.950 & 0.905 & 1.000 & -0.943 & -0.931 \\
 & MSE & -0.468 & -0.463 & -0.473 & -0.100 & 0.132 & -0.463 & -0.110 & -0.010 & 0.145 & -0.068 & -0.753 & -0.755 & -0.742 & 0.007 & 0.067 & -0.755 & 0.972 & -0.854 & -0.826 & -0.943 & 1.000 & 0.972 \\
 & MAE & -0.428 & -0.422 & -0.431 & -0.102 & 0.136 & -0.422 & -0.111 & -0.041 & 0.127 & -0.051 & -0.750 & -0.752 & -0.737 & 0.021 & 0.066 & -0.752 & 1.000 & -0.843 & -0.811 & -0.931 & 0.972 & 1.000 \\
\cline{1-24}
\bottomrule
\end{tabular}
}
\caption{Spearman Correlations Between Classification and Regression Metrics}
\label{tab:class_vs_regression_spearman}
\end{table*}

\newpage
\clearpage

\twocolumn
\section{Correlation Between BBBP Dataset Splits}

Average correlations between folds for different metrics are calculated using values in Tables \ref{tab:bbbp_splits} and \ref{tab:spearman_bbbp_splits}:\\
\begin{enumerate}
    \item Average Pearson Correlation between folds:
    \begin{enumerate}
        \item Rating: 44.5\%
        \item AP: 2.5\%
        \item ROC-AUC: 4.9\%
    \end{enumerate}
    \item Average Spearman Correlation between folds:
    \begin{enumerate}
        \item Rating: 39.1\%
        \item AP: 0.5\%
        \item ROC-AUC: 1.4\%
    \end{enumerate}

\end{enumerate}

\newpage
\clearpage

\begin{sidewaystable*}

\centering
\adjustbox{max width=24cm}{
\begin{tabular}{llHrHHHHrrrrHrHHHHrrrrHrHHHHrrrrHrHHHHrrrrHrHHHHrrrr}
\toprule
 &  & \multicolumn{10}{r}{Fold 0} & \multicolumn{10}{r}{Fold 1} & \multicolumn{10}{r}{Fold 2} & \multicolumn{10}{r}{Fold 3} & \multicolumn{10}{r}{Fold 4} \\
 &  & lower & Rating & upper & error\_y & error\_y\_minus & Rating\_rounded & ROC-AUC & BCE Loss & Avg. Error & AP & lower & Rating & upper & error\_y & error\_y\_minus & Rating\_rounded & ROC-AUC & BCE Loss & Avg. Error & AP & lower & Rating & upper & error\_y & error\_y\_minus & Rating\_rounded & ROC-AUC & BCE Loss & Avg. Error & AP & lower & Rating & upper & error\_y & error\_y\_minus & Rating\_rounded & ROC-AUC & BCE Loss & Avg. Error & AP & lower & Rating & upper & error\_y & error\_y\_minus & Rating\_rounded & ROC-AUC & BCE Loss & Avg. Error & AP \\
\midrule
Fold 0 & Rating & 0.999 & 1.000 & 0.999 & 0.022 & 0.089 & 1.000 & 0.784 & -0.449 & -0.537 & 0.645 & 0.475 & 0.477 & 0.477 & 0.095 & -0.041 & 0.477 & -0.124 & -0.148 & -0.187 & -0.201 & 0.550 & 0.547 & 0.548 & -0.001 & -0.113 & 0.547 & 0.309 & -0.225 & -0.382 & 0.305 & 0.447 & 0.442 & 0.443 & 0.118 & -0.280 & 0.442 & 0.045 & -0.108 & 0.020 & 0.028 & 0.392 & 0.395 & 0.395 & 0.022 & 0.029 & 0.395 & -0.102 & -0.094 & -0.285 & -0.189 \\
 & ROC-AUC & 0.783 & 0.784 & 0.784 & 0.046 & 0.089 & 0.784 & 1.000 & -0.683 & -0.691 & 0.887 & 0.434 & 0.432 & 0.434 & 0.135 & -0.098 & 0.432 & -0.013 & -0.228 & -0.262 & -0.094 & 0.467 & 0.467 & 0.465 & -0.081 & -0.049 & 0.467 & 0.296 & -0.341 & -0.430 & 0.257 & 0.431 & 0.428 & 0.428 & 0.103 & -0.236 & 0.428 & 0.211 & -0.225 & -0.004 & 0.317 & 0.407 & 0.409 & 0.408 & 0.003 & 0.003 & 0.409 & -0.016 & -0.128 & -0.407 & -0.098 \\
 & BCE Loss & -0.452 & -0.449 & -0.453 & -0.128 & 0.070 & -0.449 & -0.683 & 1.000 & 0.389 & -0.520 & -0.385 & -0.383 & -0.388 & -0.190 & 0.091 & -0.383 & -0.221 & 0.366 & 0.277 & -0.160 & -0.379 & -0.381 & -0.383 & -0.035 & -0.024 & -0.381 & -0.140 & 0.341 & 0.412 & -0.128 & -0.389 & -0.389 & -0.392 & -0.171 & 0.060 & -0.389 & -0.179 & 0.104 & -0.037 & -0.316 & -0.374 & -0.375 & -0.371 & 0.074 & 0.029 & -0.375 & -0.069 & 0.217 & 0.303 & 0.015 \\
 & Avg. Error & -0.533 & -0.537 & -0.536 & 0.019 & -0.150 & -0.537 & -0.691 & 0.389 & 1.000 & -0.469 & -0.481 & -0.483 & -0.484 & -0.128 & 0.048 & -0.483 & -0.101 & 0.293 & 0.271 & -0.032 & -0.259 & -0.258 & -0.260 & -0.055 & 0.059 & -0.258 & -0.128 & 0.289 & 0.441 & -0.043 & -0.365 & -0.361 & -0.359 & 0.010 & 0.223 & -0.361 & -0.320 & 0.279 & 0.085 & -0.455 & -0.425 & -0.427 & -0.431 & -0.123 & -0.017 & -0.427 & -0.071 & 0.269 & 0.382 & 0.017 \\
 & AP & 0.644 & 0.645 & 0.646 & 0.055 & 0.077 & 0.645 & 0.887 & -0.520 & -0.469 & 1.000 & 0.219 & 0.217 & 0.215 & -0.005 & -0.074 & 0.217 & 0.028 & -0.180 & -0.199 & -0.004 & 0.305 & 0.304 & 0.300 & -0.152 & -0.052 & 0.304 & 0.264 & -0.165 & -0.206 & 0.255 & 0.270 & 0.266 & 0.266 & 0.049 & -0.213 & 0.266 & 0.060 & -0.130 & -0.015 & 0.106 & 0.200 & 0.202 & 0.200 & -0.011 & 0.016 & 0.202 & -0.075 & 0.051 & -0.315 & -0.111 \\
\cline{1-52}
Fold 1 & Rating & 0.472 & 0.477 & 0.475 & -0.056 & 0.183 & 0.477 & 0.432 & -0.383 & -0.483 & 0.217 & 0.999 & 1.000 & 0.999 & 0.198 & -0.150 & 1.000 & 0.008 & -0.299 & -0.368 & -0.087 & 0.193 & 0.191 & 0.188 & -0.108 & -0.052 & 0.191 & 0.054 & -0.124 & -0.228 & -0.030 & 0.434 & 0.433 & 0.432 & 0.054 & -0.122 & 0.433 & 0.109 & -0.249 & -0.160 & 0.145 & 0.431 & 0.432 & 0.440 & 0.216 & -0.023 & 0.432 & 0.139 & -0.242 & -0.082 & 0.090 \\
 & ROC-AUC & -0.123 & -0.124 & -0.123 & 0.017 & -0.028 & -0.124 & -0.013 & -0.221 & -0.101 & 0.028 & 0.007 & 0.008 & 0.005 & -0.051 & 0.017 & 0.008 & 1.000 & -0.807 & -0.187 & 0.869 & -0.130 & -0.138 & -0.135 & 0.114 & -0.238 & -0.138 & -0.097 & 0.016 & 0.067 & -0.085 & 0.043 & 0.040 & 0.039 & -0.007 & -0.126 & 0.040 & 0.021 & -0.067 & -0.217 & 0.087 & -0.112 & -0.115 & -0.113 & 0.034 & -0.057 & -0.115 & -0.003 & 0.074 & 0.042 & 0.050 \\
 & BCE Loss & -0.149 & -0.148 & -0.150 & -0.056 & 0.027 & -0.148 & -0.228 & 0.366 & 0.293 & -0.180 & -0.294 & -0.299 & -0.299 & -0.069 & -0.031 & -0.299 & -0.807 & 1.000 & 0.476 & -0.591 & -0.024 & -0.018 & -0.020 & -0.057 & 0.195 & -0.018 & 0.127 & -0.003 & 0.038 & 0.124 & -0.234 & -0.232 & -0.232 & -0.055 & 0.169 & -0.232 & -0.020 & 0.132 & 0.218 & -0.078 & -0.065 & -0.063 & -0.068 & -0.120 & 0.051 & -0.063 & -0.018 & 0.033 & -0.033 & -0.037 \\
 & Avg. Error & -0.191 & -0.187 & -0.190 & -0.105 & 0.110 & -0.187 & -0.262 & 0.277 & 0.271 & -0.199 & -0.366 & -0.368 & -0.369 & -0.106 & 0.037 & -0.368 & -0.187 & 0.476 & 1.000 & -0.108 & -0.040 & -0.041 & -0.040 & 0.040 & -0.033 & -0.041 & 0.050 & -0.040 & 0.053 & 0.038 & -0.160 & -0.162 & -0.164 & -0.099 & -0.059 & -0.162 & -0.018 & 0.097 & 0.121 & -0.176 & -0.227 & -0.227 & -0.232 & -0.130 & 0.017 & -0.227 & -0.013 & 0.048 & 0.054 & -0.023 \\
 & AP & -0.206 & -0.201 & -0.202 & -0.040 & 0.123 & -0.201 & -0.094 & -0.160 & -0.032 & -0.004 & -0.090 & -0.087 & -0.090 & -0.082 & 0.070 & -0.087 & 0.869 & -0.591 & -0.108 & 1.000 & -0.207 & -0.216 & -0.214 & 0.069 & -0.259 & -0.216 & -0.138 & 0.103 & 0.126 & -0.099 & -0.054 & -0.057 & -0.058 & -0.060 & -0.107 & -0.057 & 0.079 & -0.070 & -0.221 & 0.045 & -0.115 & -0.118 & -0.118 & -0.020 & -0.060 & -0.118 & -0.089 & 0.113 & 0.049 & -0.041 \\
\cline{1-52}
Fold 2 & Rating & 0.552 & 0.547 & 0.543 & -0.093 & -0.118 & 0.547 & 0.467 & -0.381 & -0.258 & 0.304 & 0.193 & 0.191 & 0.198 & 0.191 & -0.063 & 0.191 & -0.138 & -0.018 & -0.041 & -0.216 & 0.999 & 1.000 & 0.999 & -0.066 & -0.043 & 1.000 & 0.332 & -0.261 & -0.325 & 0.337 & 0.482 & 0.480 & 0.481 & 0.122 & -0.164 & 0.480 & -0.035 & -0.009 & 0.057 & 0.011 & 0.515 & 0.519 & 0.513 & -0.082 & 0.038 & 0.519 & 0.177 & -0.393 & -0.167 & 0.057 \\
 & ROC-AUC & 0.313 & 0.309 & 0.304 & -0.134 & -0.076 & 0.309 & 0.296 & -0.140 & -0.128 & 0.264 & 0.058 & 0.054 & 0.064 & 0.224 & -0.074 & 0.054 & -0.097 & 0.127 & 0.050 & -0.138 & 0.338 & 0.332 & 0.329 & -0.116 & -0.204 & 0.332 & 1.000 & -0.611 & -0.360 & 0.923 & 0.143 & 0.147 & 0.147 & 0.044 & 0.130 & 0.147 & 0.194 & -0.047 & 0.125 & 0.216 & 0.192 & 0.198 & 0.197 & -0.020 & 0.137 & 0.198 & 0.029 & -0.150 & -0.173 & -0.067 \\
 & BCE Loss & -0.230 & -0.225 & -0.223 & 0.053 & 0.105 & -0.225 & -0.341 & 0.341 & 0.289 & -0.165 & -0.130 & -0.124 & -0.135 & -0.279 & 0.140 & -0.124 & 0.016 & -0.003 & -0.040 & 0.103 & -0.263 & -0.261 & -0.260 & 0.044 & 0.079 & -0.261 & -0.611 & 1.000 & 0.775 & -0.439 & -0.252 & -0.256 & -0.258 & -0.134 & -0.118 & -0.256 & -0.367 & 0.105 & -0.185 & -0.428 & -0.210 & -0.216 & -0.207 & 0.187 & -0.154 & -0.216 & 0.026 & 0.108 & 0.359 & 0.105 \\
 & Avg. Error & -0.384 & -0.382 & -0.383 & -0.028 & 0.015 & -0.382 & -0.430 & 0.412 & 0.441 & -0.206 & -0.230 & -0.228 & -0.237 & -0.235 & 0.068 & -0.228 & 0.067 & 0.038 & 0.053 & 0.126 & -0.326 & -0.325 & -0.327 & -0.057 & 0.061 & -0.325 & -0.360 & 0.775 & 1.000 & -0.252 & -0.380 & -0.381 & -0.380 & -0.033 & 0.020 & -0.381 & -0.273 & 0.100 & -0.086 & -0.406 & -0.353 & -0.360 & -0.353 & 0.129 & -0.133 & -0.360 & -0.001 & 0.212 & 0.313 & 0.082 \\
 & AP & 0.309 & 0.305 & 0.301 & -0.098 & -0.095 & 0.305 & 0.257 & -0.128 & -0.043 & 0.255 & -0.030 & -0.030 & -0.022 & 0.185 & 0.004 & -0.030 & -0.085 & 0.124 & 0.038 & -0.099 & 0.343 & 0.337 & 0.336 & -0.043 & -0.192 & 0.337 & 0.923 & -0.439 & -0.252 & 1.000 & 0.072 & 0.075 & 0.075 & 0.026 & 0.111 & 0.075 & 0.125 & 0.003 & 0.163 & 0.176 & 0.145 & 0.151 & 0.149 & -0.022 & 0.152 & 0.151 & 0.041 & -0.155 & -0.119 & -0.050 \\
\cline{1-52}
Fold 3 & Rating & 0.446 & 0.442 & 0.441 & -0.005 & -0.079 & 0.442 & 0.428 & -0.389 & -0.361 & 0.266 & 0.426 & 0.433 & 0.432 & 0.072 & 0.057 & 0.433 & 0.040 & -0.232 & -0.162 & -0.057 & 0.483 & 0.480 & 0.481 & -0.008 & -0.098 & 0.480 & 0.147 & -0.256 & -0.381 & 0.075 & 1.000 & 1.000 & 1.000 & 0.182 & -0.213 & 1.000 & -0.007 & -0.314 & -0.375 & 0.163 & 0.529 & 0.534 & 0.533 & 0.016 & 0.064 & 0.534 & 0.038 & -0.191 & -0.240 & -0.029 \\
 & ROC-AUC & 0.044 & 0.045 & 0.048 & 0.081 & 0.037 & 0.045 & 0.211 & -0.179 & -0.320 & 0.060 & 0.113 & 0.109 & 0.120 & 0.268 & -0.088 & 0.109 & 0.021 & -0.020 & -0.018 & 0.079 & -0.032 & -0.035 & -0.032 & 0.112 & -0.098 & -0.035 & 0.194 & -0.367 & -0.273 & 0.125 & -0.007 & -0.007 & -0.006 & 0.068 & -0.014 & -0.007 & 1.000 & -0.622 & -0.066 & 0.731 & 0.038 & 0.041 & 0.037 & -0.088 & 0.057 & 0.041 & -0.133 & 0.037 & -0.168 & -0.157 \\
 & BCE Loss & -0.104 & -0.108 & -0.107 & 0.002 & -0.110 & -0.108 & -0.225 & 0.104 & 0.279 & -0.130 & -0.243 & -0.249 & -0.249 & -0.053 & -0.074 & -0.249 & -0.067 & 0.132 & 0.097 & -0.070 & -0.011 & -0.009 & -0.010 & -0.044 & 0.056 & -0.009 & -0.047 & 0.105 & 0.100 & 0.003 & -0.314 & -0.314 & -0.312 & 0.016 & 0.074 & -0.314 & -0.622 & 1.000 & 0.670 & -0.488 & -0.141 & -0.143 & -0.141 & 0.021 & -0.047 & -0.143 & 0.067 & 0.040 & 0.079 & 0.086 \\
 & Avg. Error & 0.022 & 0.020 & 0.024 & 0.118 & -0.046 & 0.020 & -0.004 & -0.037 & 0.085 & -0.015 & -0.150 & -0.160 & -0.152 & 0.149 & -0.160 & -0.160 & -0.217 & 0.218 & 0.121 & -0.221 & 0.051 & 0.057 & 0.054 & -0.117 & 0.180 & 0.057 & 0.125 & -0.185 & -0.086 & 0.163 & -0.377 & -0.375 & -0.372 & 0.046 & 0.147 & -0.375 & -0.066 & 0.670 & 1.000 & -0.138 & -0.096 & -0.095 & -0.100 & -0.112 & 0.024 & -0.095 & 0.040 & -0.021 & -0.052 & 0.033 \\
 & AP & 0.032 & 0.028 & 0.028 & 0.008 & -0.119 & 0.028 & 0.317 & -0.316 & -0.455 & 0.106 & 0.146 & 0.145 & 0.152 & 0.209 & -0.051 & 0.145 & 0.087 & -0.078 & -0.176 & 0.045 & 0.013 & 0.011 & 0.016 & 0.140 & -0.054 & 0.011 & 0.216 & -0.428 & -0.406 & 0.176 & 0.163 & 0.163 & 0.167 & 0.165 & -0.031 & 0.163 & 0.731 & -0.488 & -0.138 & 1.000 & 0.233 & 0.235 & 0.230 & -0.084 & 0.006 & 0.235 & 0.012 & -0.126 & -0.252 & -0.031 \\
\cline{1-52}
Fold 4 & Rating & 0.395 & 0.395 & 0.393 & -0.059 & 0.035 & 0.395 & 0.409 & -0.375 & -0.427 & 0.202 & 0.430 & 0.432 & 0.438 & 0.220 & -0.033 & 0.432 & -0.115 & -0.063 & -0.227 & -0.118 & 0.517 & 0.519 & 0.516 & -0.097 & 0.029 & 0.519 & 0.198 & -0.216 & -0.360 & 0.151 & 0.532 & 0.534 & 0.538 & 0.230 & -0.032 & 0.534 & 0.041 & -0.143 & -0.095 & 0.235 & 0.999 & 1.000 & 0.999 & 0.070 & -0.114 & 1.000 & 0.235 & -0.457 & -0.341 & 0.115 \\
 & ROC-AUC & -0.098 & -0.102 & -0.110 & -0.219 & -0.144 & -0.102 & -0.016 & -0.069 & -0.071 & -0.075 & 0.137 & 0.139 & 0.139 & 0.018 & 0.014 & 0.139 & -0.003 & -0.018 & -0.013 & -0.089 & 0.171 & 0.177 & 0.177 & -0.029 & 0.183 & 0.177 & 0.029 & 0.026 & -0.001 & 0.041 & 0.038 & 0.038 & 0.033 & -0.174 & -0.044 & 0.038 & -0.133 & 0.067 & 0.040 & 0.012 & 0.236 & 0.235 & 0.246 & 0.273 & -0.060 & 0.235 & 1.000 & -0.729 & -0.087 & 0.955 \\
 & BCE Loss & -0.097 & -0.094 & -0.089 & 0.147 & 0.094 & -0.094 & -0.128 & 0.217 & 0.269 & 0.051 & -0.247 & -0.242 & -0.243 & -0.068 & 0.128 & -0.242 & 0.074 & 0.033 & 0.048 & 0.113 & -0.386 & -0.393 & -0.394 & -0.031 & -0.185 & -0.393 & -0.150 & 0.108 & 0.212 & -0.155 & -0.190 & -0.191 & -0.189 & 0.019 & 0.035 & -0.191 & 0.037 & 0.040 & -0.021 & -0.126 & -0.456 & -0.457 & -0.463 & -0.175 & 0.039 & -0.457 & -0.729 & 1.000 & 0.004 & -0.604 \\
 & Avg. Error & -0.290 & -0.285 & -0.286 & -0.060 & 0.123 & -0.285 & -0.407 & 0.303 & 0.382 & -0.315 & -0.077 & -0.082 & -0.085 & -0.090 & -0.073 & -0.082 & 0.042 & -0.033 & 0.054 & 0.049 & -0.163 & -0.167 & -0.166 & 0.056 & -0.107 & -0.167 & -0.173 & 0.359 & 0.313 & -0.119 & -0.239 & -0.240 & -0.238 & 0.008 & 0.043 & -0.240 & -0.168 & 0.079 & -0.052 & -0.252 & -0.336 & -0.341 & -0.341 & -0.051 & -0.085 & -0.341 & -0.087 & 0.004 & 1.000 & 0.019 \\
 & AP & -0.184 & -0.189 & -0.194 & -0.150 & -0.153 & -0.189 & -0.098 & 0.015 & 0.017 & -0.111 & 0.090 & 0.090 & 0.088 & -0.016 & -0.026 & 0.090 & 0.050 & -0.037 & -0.023 & -0.041 & 0.052 & 0.057 & 0.056 & -0.030 & 0.153 & 0.057 & -0.067 & 0.105 & 0.082 & -0.050 & -0.028 & -0.029 & -0.034 & -0.209 & -0.041 & -0.029 & -0.157 & 0.086 & 0.033 & -0.031 & 0.118 & 0.115 & 0.125 & 0.244 & -0.095 & 0.115 & 0.955 & -0.604 & 0.019 & 1.000 \\
\cline{1-52}
\bottomrule
\end{tabular}
}
\caption{Pearson Correlations Between BBBP Dataset Splits}
\label{tab:bbbp_splits}

\bigskip\bigskip

\centering
\adjustbox{max width=24cm}{
\begin{tabular}{llHrHHHHrrrrHrHHHHrrrrHrHHHHrrrrHrHHHHrrrrHrHHHHrrrr}
\toprule
 &  & \multicolumn{10}{r}{Fold 0} & \multicolumn{10}{r}{Fold 1} & \multicolumn{10}{r}{Fold 2} & \multicolumn{10}{r}{Fold 3} & \multicolumn{10}{r}{Fold 4} \\
 &  & lower & Rating & upper & error\_y & error\_y\_minus & Rating\_rounded & ROC-AUC & BCE Loss & Avg. Error & AP & lower & Rating & upper & error\_y & error\_y\_minus & Rating\_rounded & ROC-AUC & BCE Loss & Avg. Error & AP & lower & Rating & upper & error\_y & error\_y\_minus & Rating\_rounded & ROC-AUC & BCE Loss & Avg. Error & AP & lower & Rating & upper & error\_y & error\_y\_minus & Rating\_rounded & ROC-AUC & BCE Loss & Avg. Error & AP & lower & Rating & upper & error\_y & error\_y\_minus & Rating\_rounded & ROC-AUC & BCE Loss & Avg. Error & AP \\
\midrule
Fold 0 & Rating & 0.997 & 1.000 & 0.998 & 0.006 & 0.012 & 1.000 & 0.706 & -0.231 & -0.377 & 0.622 & 0.401 & 0.408 & 0.404 & 0.008 & -0.025 & 0.408 & -0.129 & -0.069 & -0.096 & -0.198 & 0.526 & 0.521 & 0.527 & -0.001 & -0.077 & 0.521 & 0.192 & -0.054 & -0.235 & 0.243 & 0.442 & 0.435 & 0.440 & 0.156 & -0.319 & 0.435 & -0.117 & -0.106 & -0.003 & -0.091 & 0.353 & 0.359 & 0.368 & 0.023 & 0.057 & 0.359 & -0.173 & -0.053 & -0.230 & -0.179 \\
 & ROC-AUC & 0.701 & 0.706 & 0.700 & 0.002 & 0.092 & 0.706 & 1.000 & -0.367 & -0.390 & 0.949 & 0.242 & 0.245 & 0.239 & 0.052 & -0.063 & 0.245 & -0.046 & -0.057 & -0.066 & -0.062 & 0.381 & 0.376 & 0.370 & -0.109 & -0.063 & 0.376 & 0.206 & -0.103 & -0.166 & 0.257 & 0.273 & 0.266 & 0.270 & 0.092 & -0.214 & 0.266 & 0.043 & -0.162 & 0.024 & 0.034 & 0.176 & 0.180 & 0.189 & 0.025 & 0.036 & 0.180 & -0.176 & 0.037 & -0.254 & -0.183 \\
 & BCE Loss & -0.231 & -0.231 & -0.234 & -0.118 & 0.014 & -0.231 & -0.367 & 1.000 & -0.273 & -0.315 & -0.128 & -0.139 & -0.137 & -0.090 & -0.054 & -0.139 & -0.019 & 0.118 & 0.216 & -0.123 & -0.243 & -0.242 & -0.240 & 0.070 & -0.014 & -0.242 & -0.095 & 0.087 & 0.118 & -0.103 & -0.186 & -0.188 & -0.197 & -0.168 & -0.029 & -0.188 & -0.083 & 0.145 & 0.008 & -0.085 & -0.163 & -0.163 & -0.162 & 0.150 & 0.085 & -0.163 & 0.122 & -0.002 & 0.062 & 0.163 \\
 & Avg. Error & -0.378 & -0.377 & -0.373 & 0.130 & -0.103 & -0.377 & -0.390 & -0.273 & 1.000 & -0.334 & -0.296 & -0.306 & -0.303 & 0.031 & 0.045 & -0.306 & -0.073 & 0.139 & 0.080 & 0.082 & -0.135 & -0.139 & -0.139 & -0.045 & 0.032 & -0.139 & 0.037 & -0.046 & 0.145 & 0.030 & -0.181 & -0.180 & -0.170 & 0.042 & 0.185 & -0.180 & -0.172 & 0.193 & 0.078 & -0.250 & -0.151 & -0.147 & -0.162 & -0.145 & 0.000 & -0.147 & -0.069 & 0.238 & 0.158 & -0.046 \\
 & AP & 0.618 & 0.622 & 0.617 & 0.003 & 0.069 & 0.622 & 0.949 & -0.315 & -0.334 & 1.000 & 0.164 & 0.160 & 0.153 & -0.002 & -0.089 & 0.160 & 0.020 & -0.073 & -0.108 & 0.013 & 0.283 & 0.277 & 0.271 & -0.088 & -0.084 & 0.277 & 0.245 & -0.124 & -0.156 & 0.304 & 0.183 & 0.175 & 0.178 & 0.053 & -0.181 & 0.175 & 0.067 & -0.164 & -0.027 & 0.045 & 0.113 & 0.119 & 0.125 & 0.038 & 0.085 & 0.119 & -0.184 & 0.098 & -0.257 & -0.178 \\
\cline{1-52}
Fold 1 & Rating & 0.406 & 0.408 & 0.409 & -0.079 & 0.101 & 0.408 & 0.245 & -0.139 & -0.306 & 0.160 & 0.995 & 1.000 & 0.997 & 0.163 & -0.119 & 1.000 & -0.042 & -0.235 & -0.350 & -0.188 & 0.132 & 0.125 & 0.131 & -0.176 & -0.005 & 0.125 & 0.042 & 0.068 & -0.043 & -0.035 & 0.315 & 0.318 & 0.318 & 0.017 & -0.042 & 0.318 & 0.014 & -0.172 & -0.090 & -0.102 & 0.301 & 0.303 & 0.309 & 0.220 & -0.065 & 0.303 & 0.113 & -0.201 & 0.026 & 0.080 \\
 & ROC-AUC & -0.129 & -0.129 & -0.143 & -0.048 & 0.040 & -0.129 & -0.046 & -0.019 & -0.073 & 0.020 & -0.052 & -0.042 & -0.047 & -0.024 & 0.034 & -0.042 & 1.000 & -0.753 & -0.193 & 0.823 & -0.125 & -0.133 & -0.131 & 0.082 & -0.251 & -0.133 & -0.075 & -0.024 & 0.153 & -0.080 & -0.004 & -0.003 & -0.001 & -0.029 & -0.141 & -0.003 & 0.161 & -0.109 & -0.217 & -0.002 & -0.194 & -0.209 & -0.192 & 0.062 & -0.124 & -0.209 & -0.016 & 0.041 & 0.139 & 0.015 \\
 & BCE Loss & -0.075 & -0.069 & -0.068 & -0.039 & 0.073 & -0.069 & -0.057 & 0.118 & 0.139 & -0.073 & -0.222 & -0.235 & -0.232 & -0.040 & -0.067 & -0.235 & -0.753 & 1.000 & 0.363 & -0.504 & 0.016 & 0.025 & 0.022 & -0.012 & 0.159 & 0.025 & 0.098 & -0.045 & -0.109 & 0.119 & -0.172 & -0.172 & -0.174 & -0.024 & 0.168 & -0.172 & -0.093 & 0.121 & 0.211 & 0.014 & 0.073 & 0.084 & 0.067 & -0.100 & 0.127 & 0.084 & 0.086 & 0.002 & -0.114 & 0.111 \\
 & Avg. Error & -0.106 & -0.096 & -0.112 & -0.104 & 0.127 & -0.096 & -0.066 & 0.216 & 0.080 & -0.108 & -0.344 & -0.350 & -0.344 & -0.066 & 0.007 & -0.350 & -0.193 & 0.363 & 1.000 & -0.180 & -0.006 & -0.002 & -0.004 & 0.051 & -0.108 & -0.002 & 0.051 & -0.174 & -0.016 & 0.021 & -0.077 & -0.080 & -0.087 & -0.057 & -0.053 & -0.080 & -0.086 & 0.095 & 0.140 & -0.057 & -0.074 & -0.073 & -0.090 & -0.119 & -0.015 & -0.073 & 0.094 & 0.037 & -0.002 & 0.074 \\
 & AP & -0.205 & -0.198 & -0.214 & -0.098 & 0.163 & -0.198 & -0.062 & -0.123 & 0.082 & 0.013 & -0.201 & -0.188 & -0.191 & -0.011 & 0.075 & -0.188 & 0.823 & -0.504 & -0.180 & 1.000 & -0.215 & -0.225 & -0.225 & 0.049 & -0.242 & -0.225 & -0.095 & 0.045 & 0.166 & -0.042 & -0.085 & -0.085 & -0.081 & -0.002 & -0.083 & -0.085 & 0.145 & -0.113 & -0.211 & 0.005 & -0.239 & -0.248 & -0.236 & 0.001 & -0.040 & -0.248 & -0.072 & 0.136 & 0.169 & -0.051 \\
\cline{1-52}
Fold 2 & Rating & 0.526 & 0.521 & 0.519 & -0.127 & -0.144 & 0.521 & 0.376 & -0.242 & -0.139 & 0.277 & 0.129 & 0.125 & 0.130 & 0.131 & -0.084 & 0.125 & -0.133 & 0.025 & -0.002 & -0.225 & 0.999 & 1.000 & 0.999 & -0.087 & -0.093 & 1.000 & 0.300 & -0.113 & -0.163 & 0.280 & 0.480 & 0.481 & 0.484 & 0.151 & -0.180 & 0.481 & -0.218 & 0.005 & 0.024 & -0.139 & 0.463 & 0.464 & 0.467 & -0.063 & 0.069 & 0.464 & 0.083 & -0.365 & -0.070 & 0.029 \\
 & ROC-AUC & 0.198 & 0.192 & 0.192 & -0.063 & -0.058 & 0.192 & 0.206 & -0.095 & 0.037 & 0.245 & 0.047 & 0.042 & 0.040 & 0.257 & -0.083 & 0.042 & -0.075 & 0.098 & 0.051 & -0.095 & 0.310 & 0.300 & 0.291 & -0.144 & -0.205 & 0.300 & 1.000 & -0.540 & -0.214 & 0.887 & 0.074 & 0.077 & 0.073 & 0.061 & 0.217 & 0.077 & 0.150 & 0.002 & 0.166 & 0.142 & 0.233 & 0.244 & 0.240 & 0.004 & 0.122 & 0.244 & -0.069 & -0.020 & -0.137 & -0.163 \\
 & BCE Loss & -0.068 & -0.054 & -0.046 & 0.086 & 0.171 & -0.054 & -0.103 & 0.087 & -0.046 & -0.124 & 0.057 & 0.068 & 0.072 & -0.120 & 0.104 & 0.068 & -0.024 & -0.045 & -0.174 & 0.045 & -0.121 & -0.113 & -0.108 & 0.065 & 0.159 & -0.113 & -0.540 & 1.000 & 0.566 & -0.366 & -0.078 & -0.088 & -0.086 & -0.146 & -0.217 & -0.088 & -0.186 & 0.001 & -0.145 & -0.138 & -0.016 & -0.019 & 0.001 & 0.165 & -0.063 & -0.019 & 0.053 & -0.069 & 0.150 & 0.123 \\
 & Avg. Error & -0.244 & -0.235 & -0.236 & 0.020 & 0.082 & -0.235 & -0.166 & 0.118 & 0.145 & -0.156 & -0.045 & -0.043 & -0.045 & -0.074 & 0.065 & -0.043 & 0.153 & -0.109 & -0.016 & 0.166 & -0.167 & -0.163 & -0.165 & -0.064 & 0.093 & -0.163 & -0.214 & 0.566 & 1.000 & -0.104 & -0.239 & -0.247 & -0.242 & 0.021 & -0.045 & -0.247 & 0.031 & -0.029 & -0.045 & -0.099 & -0.095 & -0.100 & -0.087 & 0.073 & -0.020 & -0.100 & -0.001 & 0.041 & 0.047 & 0.006 \\
 & AP & 0.247 & 0.243 & 0.243 & -0.067 & -0.098 & 0.243 & 0.257 & -0.103 & 0.030 & 0.304 & -0.032 & -0.035 & -0.038 & 0.224 & -0.073 & -0.035 & -0.080 & 0.119 & 0.021 & -0.042 & 0.290 & 0.280 & 0.273 & -0.036 & -0.205 & 0.280 & 0.887 & -0.366 & -0.104 & 1.000 & -0.015 & -0.015 & -0.016 & 0.038 & 0.164 & -0.015 & 0.148 & 0.015 & 0.188 & 0.168 & 0.109 & 0.120 & 0.122 & -0.027 & 0.186 & 0.120 & -0.121 & -0.003 & -0.133 & -0.163 \\
\cline{1-52}
Fold 3 & Rating & 0.440 & 0.435 & 0.431 & -0.030 & -0.083 & 0.435 & 0.266 & -0.188 & -0.180 & 0.175 & 0.325 & 0.318 & 0.320 & 0.037 & 0.068 & 0.318 & -0.003 & -0.172 & -0.080 & -0.085 & 0.480 & 0.481 & 0.484 & -0.081 & -0.116 & 0.481 & 0.077 & -0.088 & -0.247 & -0.015 & 0.999 & 1.000 & 0.999 & 0.191 & -0.224 & 1.000 & -0.083 & -0.288 & -0.377 & 0.046 & 0.484 & 0.491 & 0.496 & 0.040 & 0.053 & 0.491 & -0.003 & -0.199 & -0.096 & -0.052 \\
 & ROC-AUC & -0.118 & -0.117 & -0.127 & 0.019 & -0.022 & -0.117 & 0.043 & -0.083 & -0.172 & 0.067 & 0.006 & 0.014 & 0.011 & 0.100 & -0.045 & 0.014 & 0.161 & -0.093 & -0.086 & 0.145 & -0.209 & -0.218 & -0.217 & 0.101 & -0.195 & -0.218 & 0.150 & -0.186 & 0.031 & 0.148 & -0.085 & -0.083 & -0.086 & 0.111 & 0.088 & -0.083 & 1.000 & -0.460 & -0.073 & 0.710 & -0.100 & -0.104 & -0.090 & 0.039 & 0.015 & -0.104 & -0.038 & 0.043 & -0.035 & -0.041 \\
 & BCE Loss & -0.103 & -0.106 & -0.101 & 0.039 & -0.151 & -0.106 & -0.162 & 0.145 & 0.193 & -0.164 & -0.164 & -0.172 & -0.172 & 0.017 & -0.100 & -0.172 & -0.109 & 0.121 & 0.095 & -0.113 & 0.001 & 0.005 & 0.002 & -0.054 & 0.057 & 0.005 & 0.002 & 0.001 & -0.029 & 0.015 & -0.288 & -0.288 & -0.282 & 0.083 & 0.061 & -0.288 & -0.460 & 1.000 & 0.686 & -0.364 & -0.126 & -0.120 & -0.124 & -0.041 & -0.016 & -0.120 & 0.104 & -0.011 & 0.061 & 0.114 \\
 & Avg. Error & -0.006 & -0.003 & 0.004 & 0.163 & -0.053 & -0.003 & 0.024 & 0.008 & 0.078 & -0.027 & -0.085 & -0.090 & -0.098 & 0.141 & -0.139 & -0.090 & -0.217 & 0.211 & 0.140 & -0.211 & 0.021 & 0.024 & 0.023 & -0.094 & 0.187 & 0.024 & 0.166 & -0.145 & -0.045 & 0.188 & -0.380 & -0.377 & -0.373 & 0.062 & 0.137 & -0.377 & -0.073 & 0.686 & 1.000 & -0.136 & -0.085 & -0.079 & -0.084 & -0.158 & 0.032 & -0.079 & 0.050 & -0.041 & -0.063 & 0.043 \\
 & AP & -0.082 & -0.091 & -0.094 & -0.071 & -0.160 & -0.091 & 0.034 & -0.085 & -0.250 & 0.045 & -0.108 & -0.102 & -0.106 & 0.059 & 0.026 & -0.102 & -0.002 & 0.014 & -0.057 & 0.005 & -0.137 & -0.139 & -0.137 & 0.194 & -0.108 & -0.139 & 0.142 & -0.138 & -0.099 & 0.168 & 0.049 & 0.046 & 0.051 & 0.205 & 0.045 & 0.046 & 0.710 & -0.364 & -0.136 & 1.000 & -0.042 & -0.039 & -0.032 & -0.069 & -0.010 & -0.039 & -0.043 & -0.037 & -0.049 & -0.053 \\
\cline{1-52}
Fold 4 & Rating & 0.369 & 0.359 & 0.368 & -0.031 & -0.033 & 0.359 & 0.180 & -0.163 & -0.147 & 0.119 & 0.309 & 0.303 & 0.309 & 0.169 & -0.004 & 0.303 & -0.209 & 0.084 & -0.073 & -0.248 & 0.462 & 0.464 & 0.460 & -0.178 & 0.057 & 0.464 & 0.244 & -0.019 & -0.100 & 0.120 & 0.487 & 0.491 & 0.489 & 0.187 & 0.008 & 0.491 & -0.104 & -0.120 & -0.079 & -0.039 & 0.998 & 1.000 & 0.995 & 0.064 & -0.143 & 1.000 & 0.108 & -0.404 & -0.162 & 0.017 \\
 & ROC-AUC & -0.165 & -0.173 & -0.174 & -0.196 & -0.097 & -0.173 & -0.176 & 0.122 & -0.069 & -0.184 & 0.104 & 0.113 & 0.116 & 0.041 & 0.043 & 0.113 & -0.016 & 0.086 & 0.094 & -0.072 & 0.084 & 0.083 & 0.092 & 0.010 & 0.082 & 0.083 & -0.069 & 0.053 & -0.001 & -0.121 & -0.007 & -0.003 & -0.002 & -0.092 & 0.030 & -0.003 & -0.038 & 0.104 & 0.050 & -0.043 & 0.105 & 0.108 & 0.125 & 0.269 & 0.017 & 0.108 & 1.000 & -0.626 & 0.041 & 0.926 \\
 & BCE Loss & -0.061 & -0.053 & -0.065 & 0.069 & 0.145 & -0.053 & 0.037 & -0.002 & 0.238 & 0.098 & -0.211 & -0.201 & -0.200 & -0.084 & 0.179 & -0.201 & 0.041 & 0.002 & 0.037 & 0.136 & -0.360 & -0.365 & -0.372 & -0.006 & -0.132 & -0.365 & -0.020 & -0.069 & 0.041 & -0.003 & -0.198 & -0.199 & -0.205 & -0.077 & 0.044 & -0.199 & 0.043 & -0.011 & -0.041 & -0.037 & -0.403 & -0.404 & -0.418 & -0.157 & 0.006 & -0.404 & -0.626 & 1.000 & -0.179 & -0.559 \\
 & Avg. Error & -0.231 & -0.230 & -0.234 & -0.075 & 0.119 & -0.230 & -0.254 & 0.062 & 0.158 & -0.257 & 0.030 & 0.026 & 0.025 & -0.050 & -0.097 & 0.026 & 0.139 & -0.114 & -0.002 & 0.169 & -0.064 & -0.070 & -0.066 & 0.095 & -0.067 & -0.070 & -0.137 & 0.150 & 0.047 & -0.133 & -0.101 & -0.096 & -0.090 & 0.049 & -0.024 & -0.096 & -0.035 & 0.061 & -0.063 & -0.049 & -0.161 & -0.162 & -0.160 & -0.072 & -0.031 & -0.162 & 0.041 & -0.179 & 1.000 & 0.166 \\
 & AP & -0.177 & -0.179 & -0.182 & -0.171 & -0.077 & -0.179 & -0.183 & 0.163 & -0.046 & -0.178 & 0.075 & 0.080 & 0.082 & -0.059 & 0.038 & 0.080 & 0.015 & 0.111 & 0.074 & -0.051 & 0.030 & 0.029 & 0.041 & 0.058 & 0.070 & 0.029 & -0.163 & 0.123 & 0.006 & -0.163 & -0.055 & -0.052 & -0.050 & -0.123 & -0.014 & -0.052 & -0.041 & 0.114 & 0.043 & -0.053 & 0.017 & 0.017 & 0.031 & 0.180 & -0.014 & 0.017 & 0.926 & -0.559 & 0.166 & 1.000 \\
\cline{1-52}
\bottomrule
\end{tabular}
}
\caption{Spearman Correlations Between BBBP Dataset Splits}
\label{tab:spearman_bbbp_splits}

\end{sidewaystable*}

\newpage
\clearpage

\section{Winrate Plots}
The following figures are winrates between different models on different datasets. Note that the aggregate winrate is a macro average. Llama3-8B-MolCap-0 is the same as Llama3-8B-Task. 

\begin{figure*}[h!]
  \centering
  \includegraphics[width=1.0\textwidth]{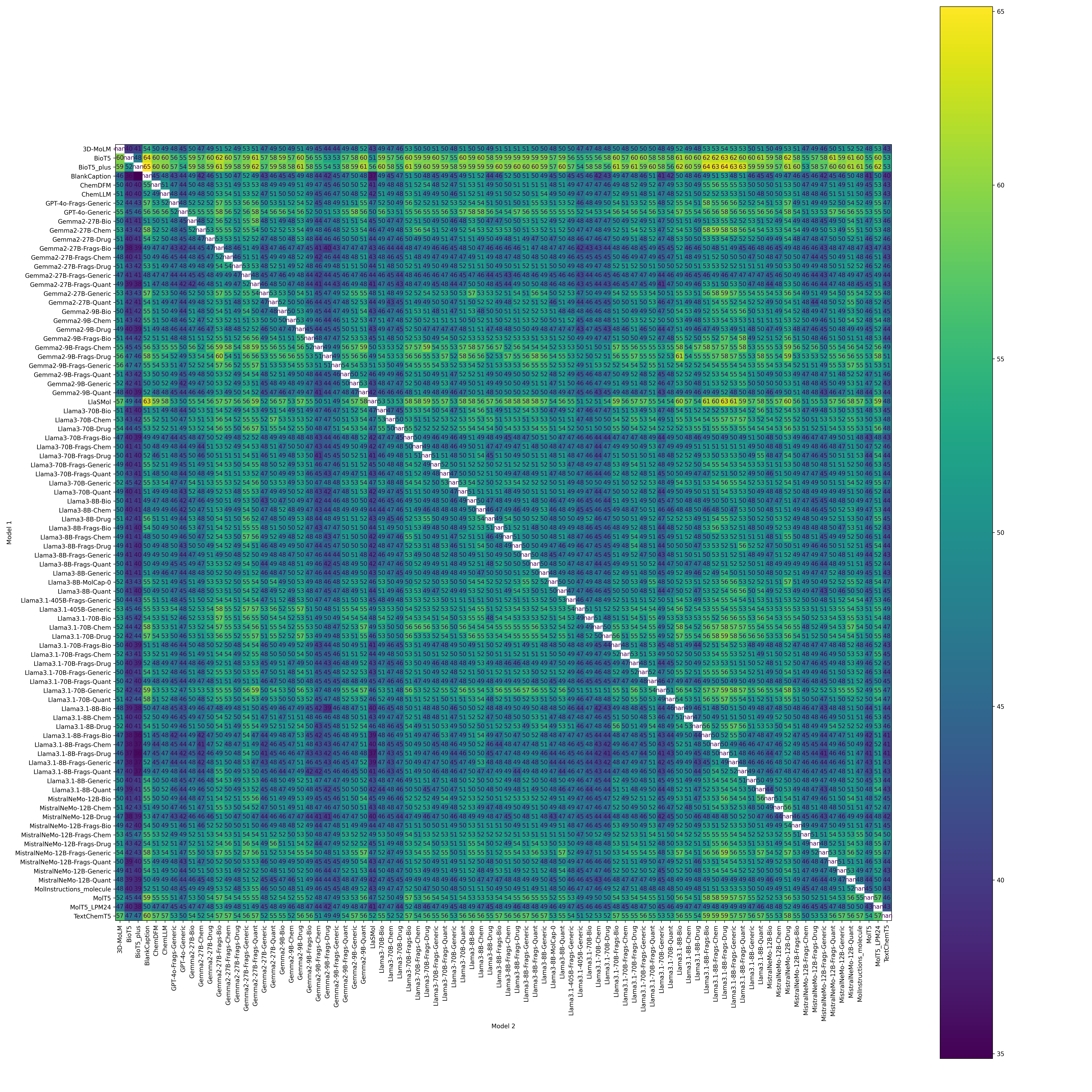}
\caption{Aggregate Winrates}
\label{fig:aggregate_winrate}
\end{figure*}

\newpage
\clearpage

\begin{figure*}[h!]
  \centering
  \includegraphics[width=1.0\textwidth]{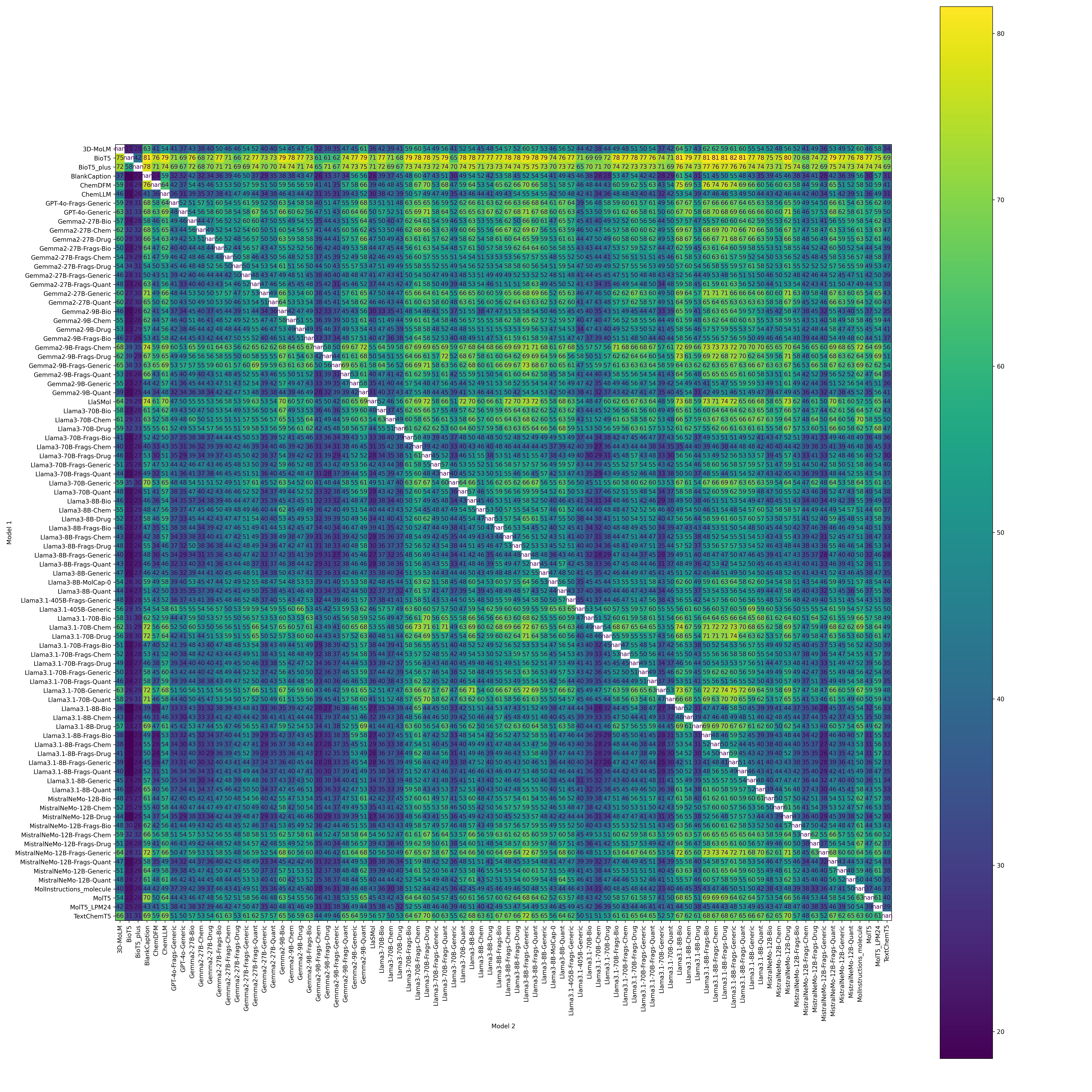}
\caption{BBBP Winrates}
\label{fig:bbbp_winrate}
\end{figure*}

\newpage
\clearpage
\begin{figure*}[h!]
  \centering
  \includegraphics[width=1.0\textwidth]{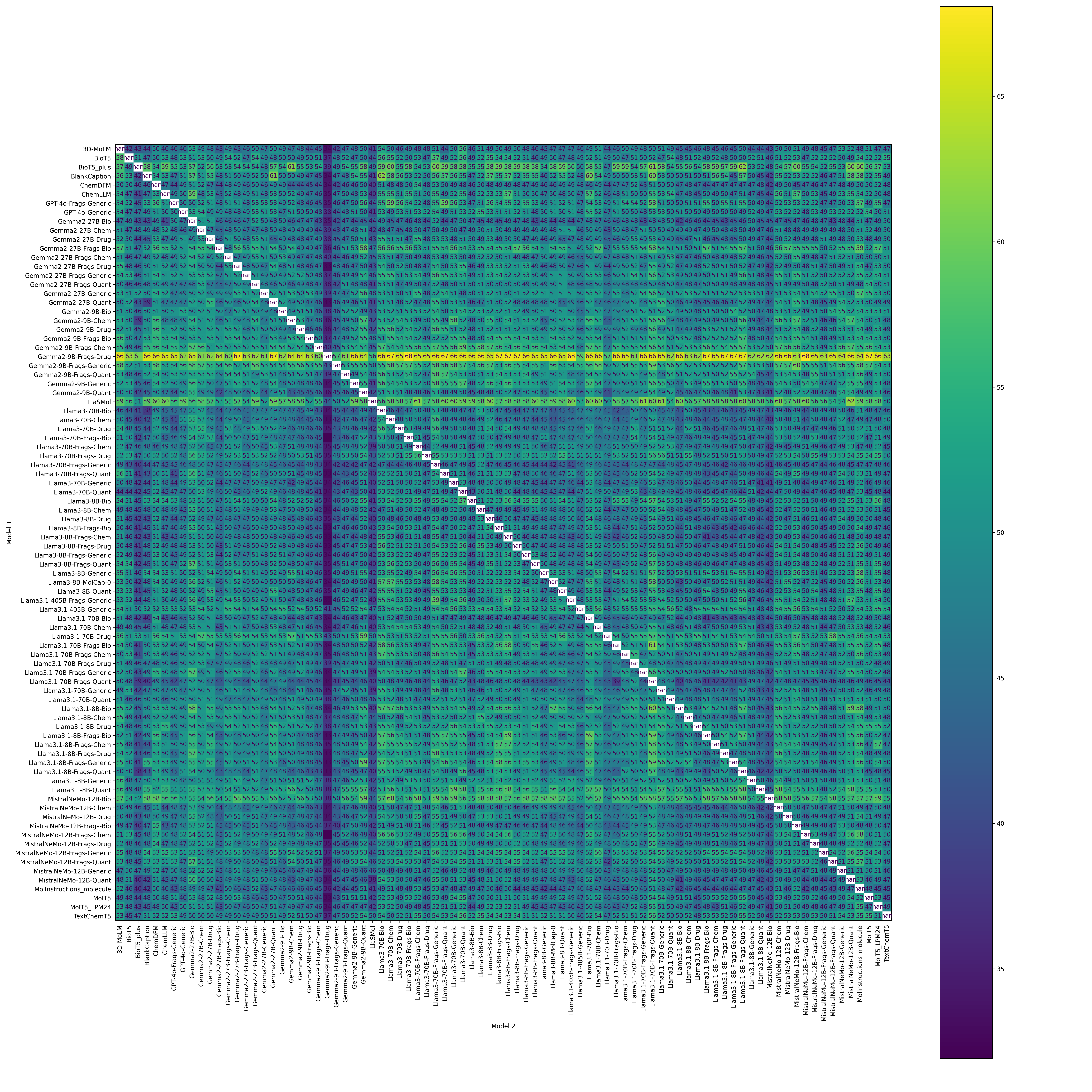}
\caption{BACE Winrates}
\label{fig:bace_winrate}
\end{figure*}

\newpage
\clearpage
\begin{figure*}[h!]
  \centering
  \includegraphics[width=1.0\textwidth]{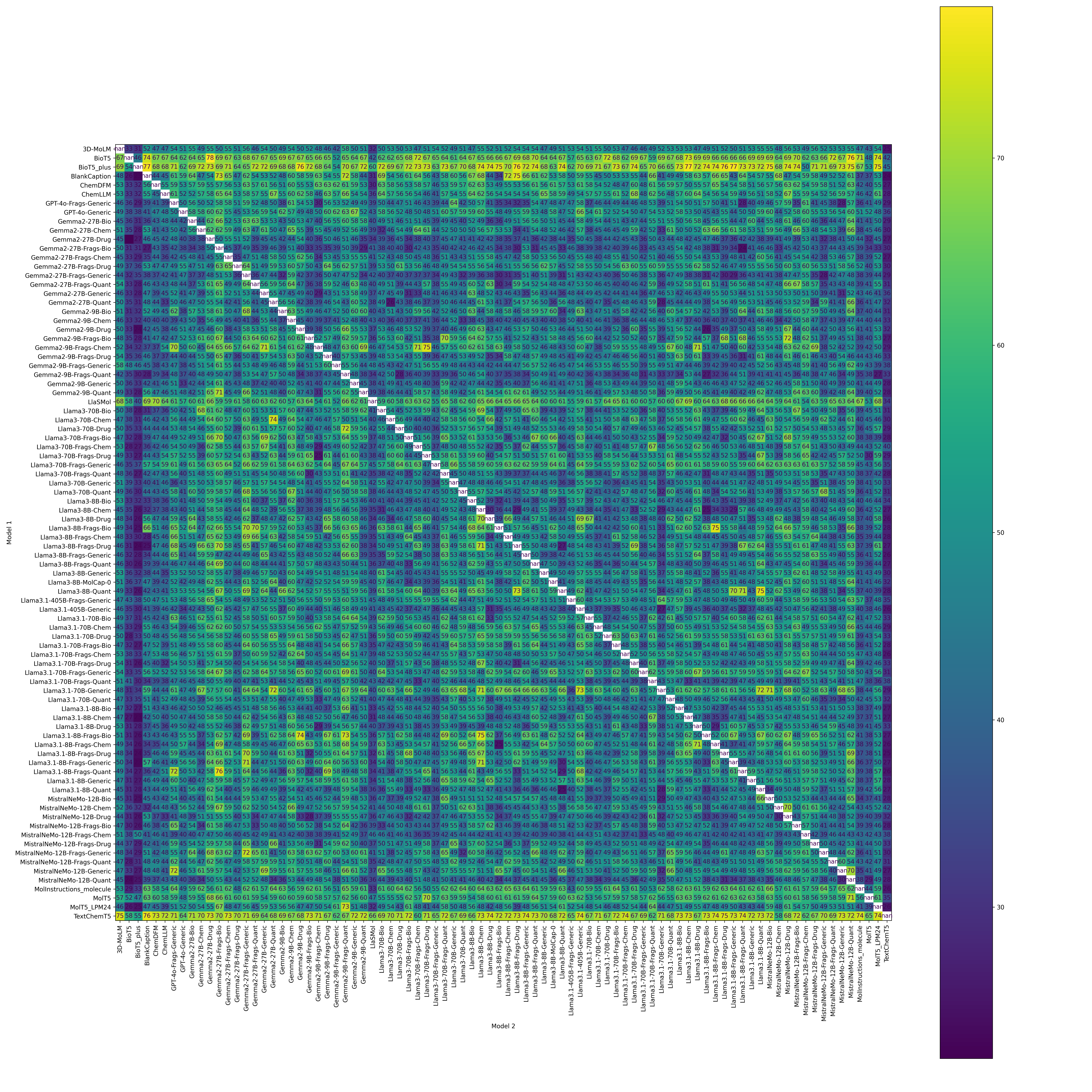}
\caption{ClinTox Winrates}
\label{fig:clintox_winrate}
\end{figure*}

\newpage
\clearpage
\begin{figure*}[h!]
  \centering
  \includegraphics[width=1.0\textwidth]{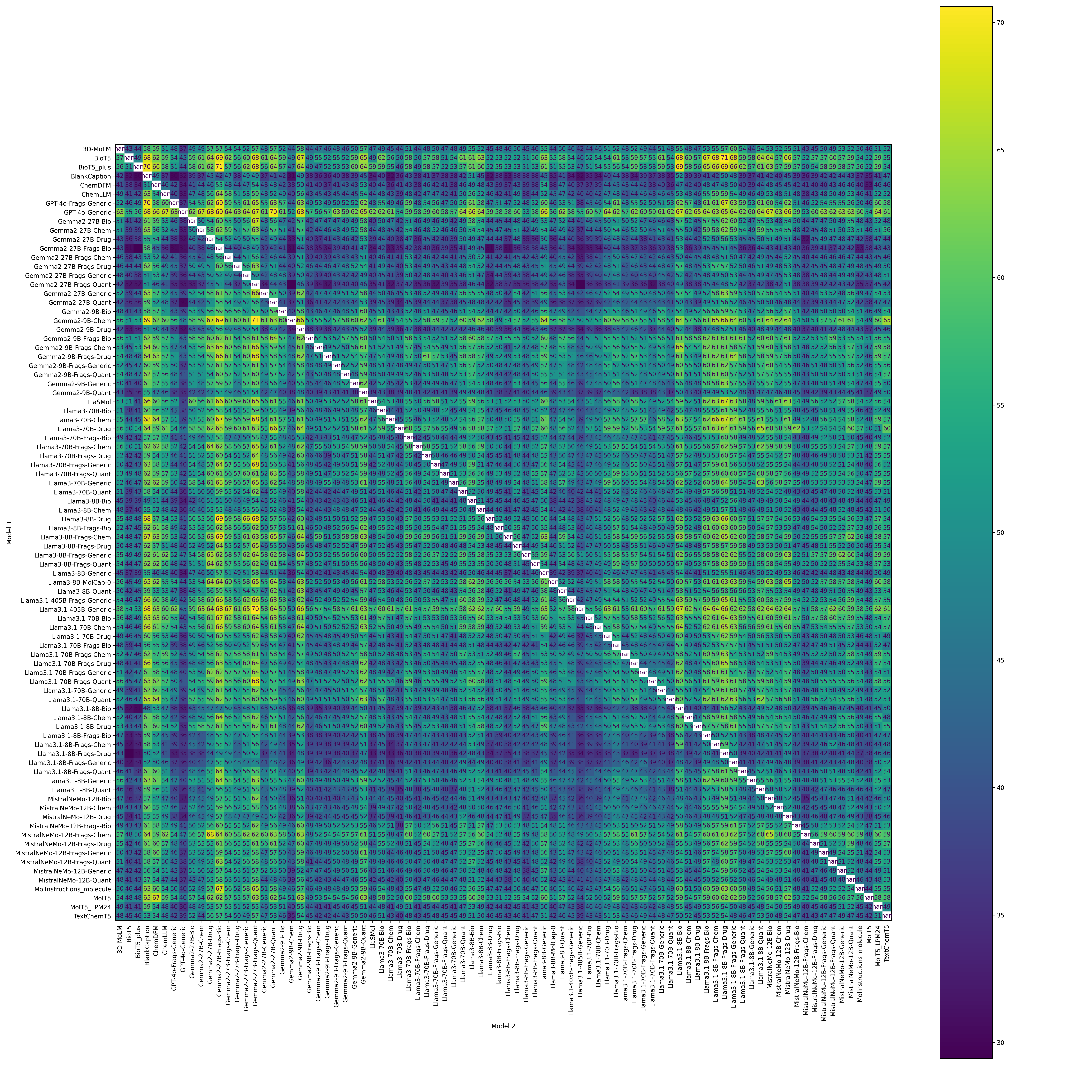}
\caption{FreeSolv Winrates}
\label{fig:freesolv_winrate}
\end{figure*}

\newpage
\clearpage
\begin{figure*}[h!]
  \centering
  \includegraphics[width=1.0\textwidth]{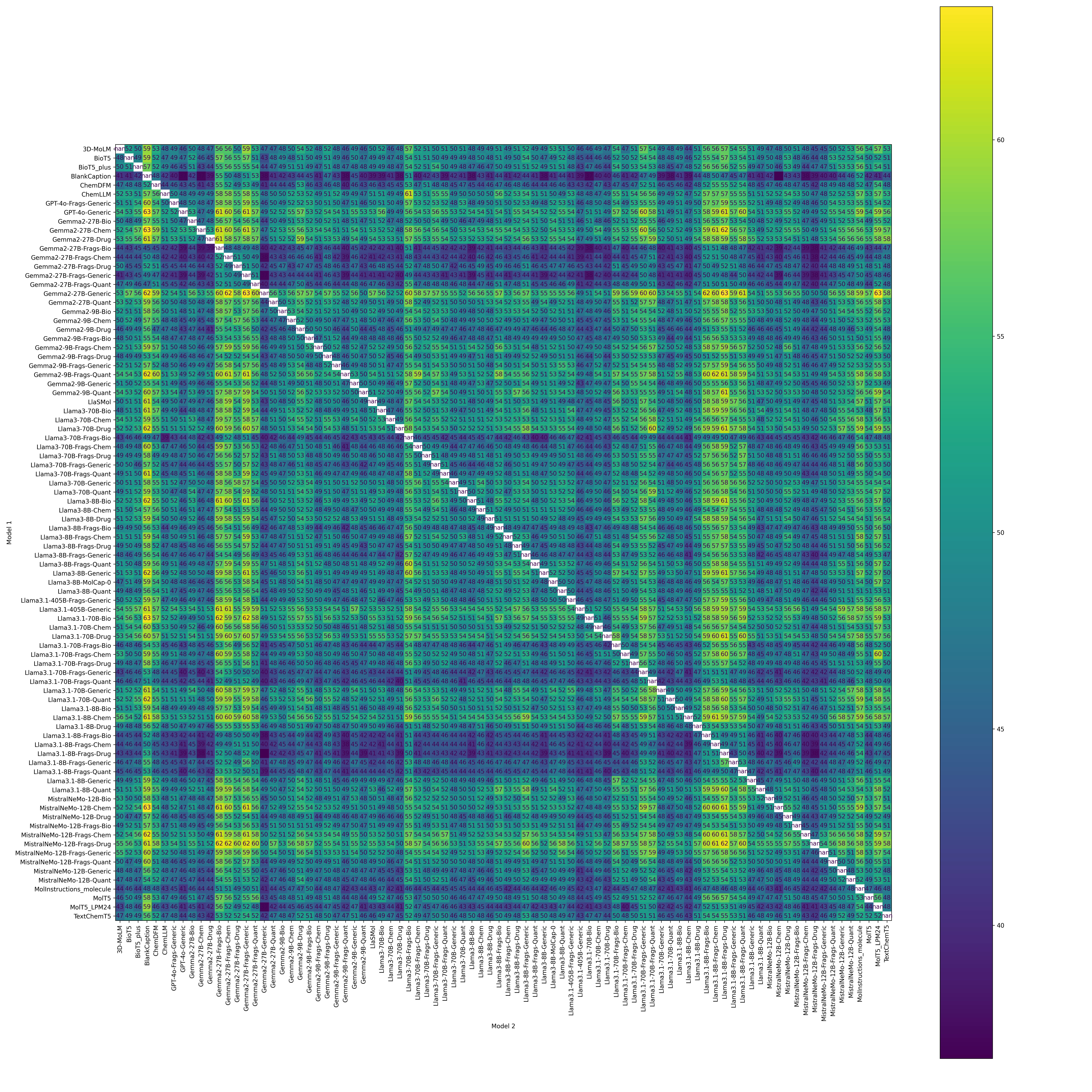}
\caption{ESOL Winrates}
\label{fig:esol_winrate}
\end{figure*}

\newpage
\clearpage
\begin{figure*}[h!]
  \centering
  \includegraphics[width=1.0\textwidth]{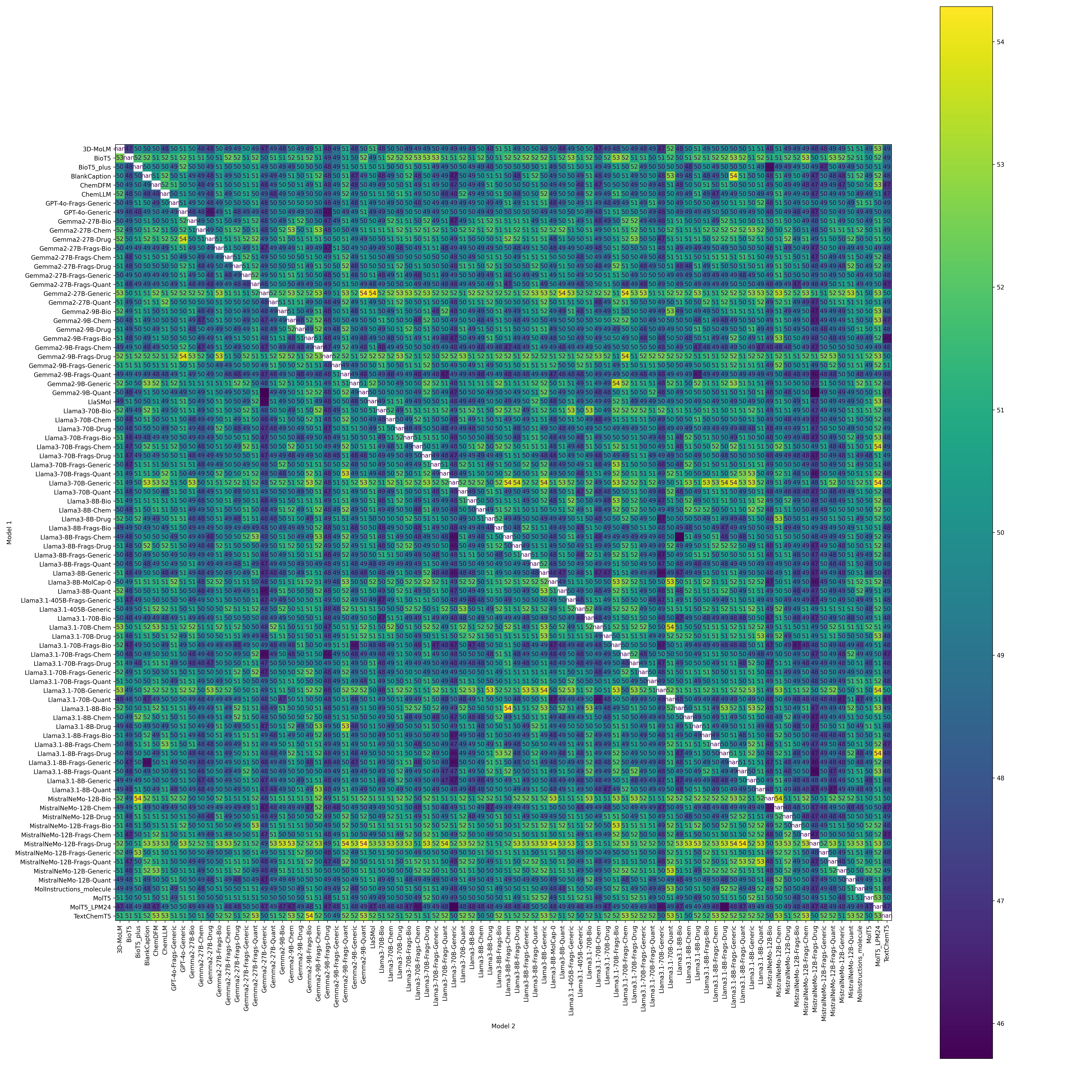}
\caption{Lipo Winrates}
\label{fig:lipo_winrate}
\end{figure*}

\newpage
\clearpage

\section{Caption Examples}
\label{app:caption_examples}
The following are captions generated by different models for the molecule trichlormethiazide.
\begin{figure}[h!]
  \centering
  \includegraphics[width=1.0\columnwidth]{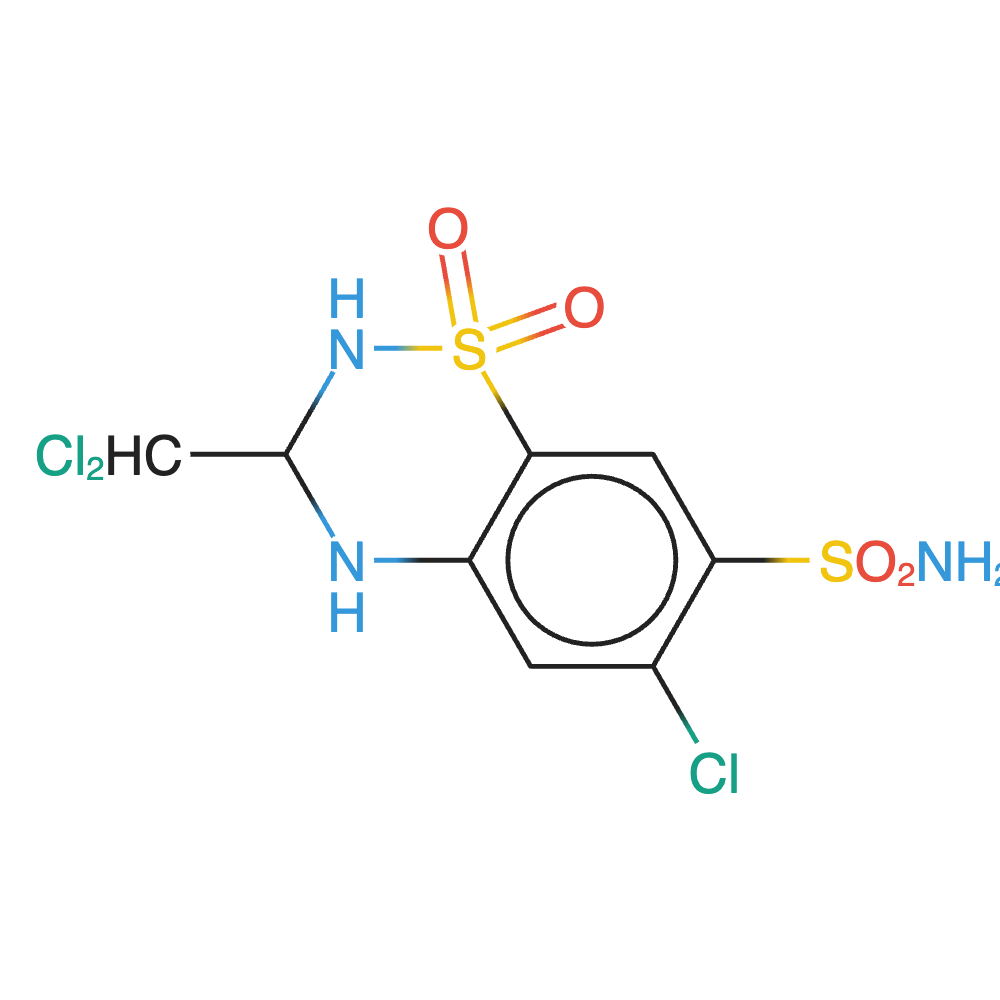}
\label{fig:trichlormethiazide}
\end{figure}

\UseRawInputEncoding

\begin{tcolorbox}
\textbf{MolT5:}
\begin{lstlisting}
The molecule is a benzothiadiazine, hydrogenated at positions 2, 3 and 4 and substituted with an oxo group at C-3, a chloro substituent at C-6 and a dichloromethyl group at C-3 and with S-1 as an S,S-dioxide. A sulfonamide antibiotic, it is used as a diuretic to treat oedema (including that associated with heart failure) and hypertension. It has a role as a diuretic and an antihypertensive agent. It is a benzothiadiazine, a sulfonamide antibiotic and an organochlorine compound.
\end{lstlisting}
\end{tcolorbox}

\begin{tcolorbox}
\textbf{MolT5\_LPM24:}
\begin{lstlisting}
It belongs to the anti viral class of molecules.
\end{lstlisting}
\end{tcolorbox}

\begin{tcolorbox}
\textbf{TextChemT5:}
\begin{lstlisting}
The molecule is a member of the class of sulfamides that is sulfamic acid in which one of the amino groups is substituted by a 2-chloroethyl group, while the other is substituted by a 2-chloroethyl group. It is a metabolite of the herbicide sulfamic acid. It has a role as a marine xenobiotic metabolite. It is a member of sulfamides, an organochlorine compound and a member of ureas. It derives from a 2-chloroethyl and a sulfamic acid.
\end{lstlisting}
\end{tcolorbox}

\begin{tcolorbox}
\textbf{3D-MoLM:}
\begin{lstlisting}
2-[3,4-dichloro-2-(dichloromethyl)-6,7-dihydro-5H-1,2,4-triazolo[4,3-b]thiophen-5-yl]sulfonyl-1,3-dihydro-2-thiolane-1-sulfonate is an organochlorine compound and an organosulfonic acid. It has a role as an anticoronaviral agent.
2-[3,4-dichloro-
\end{lstlisting}
\end{tcolorbox}

\begin{tcolorbox}
\textbf{LlaSMol:}
\begin{lstlisting}
The molecule is a sulfamoylbenzodiazepine that is 1,4-benzodiazepin-2-one substituted by a chloro group at position 5, a sulfamoyl group at position 3 and two chloro groups at position 7. It is a sulfamoylbenzodiazepine, a member of monochlorobenzenes and a dichlorobenzene.
\end{lstlisting}
\end{tcolorbox}

\begin{tcolorbox}
\textbf{BioT5:}
\begin{lstlisting}
The molecule is a benzothiadiazine, hydrogenated at positions 2, 3 and 4 and substituted with an aminosulfonyl group at C-7, a chloro substituent at C-6 and a dichloromethyl group at C-3 and with S-1 as an S,S-dioxide. A sulfonamide antibiotic, it is used as a diuretic to treat oedema (including that associated with heart failure) and hypertension. It has a role as a diuretic and an antihypertensive agent. It is a benzothiadiazine and a sulfonamide antibiotic.
\end{lstlisting}
\end{tcolorbox}

\begin{tcolorbox}
\textbf{BioT5\_plus:}
\begin{lstlisting}
The molecule is a benzothiadiazole, hydrogenated at positions ,  and  and substituted with an aminosulfonyl group at C-, a chloro substituent at C- and a dichloromethyl group at C- and with S- as an S,S-dioxide A sulfonamide antibiotic, it is used as a diuretic to treat oedema (including that associated with heart failure) and hypertension It has a role as a diuretic and an antihypertensive agent It is a benzothiadiazole and a sulfonamide antibiotic
\end{lstlisting}
\end{tcolorbox}

\begin{tcolorbox}
\textbf{ChemDFM:}
\begin{lstlisting}
1,3-Dichloro-5-(methanesulfonyl)-7NS(=O)(=O)indazane is a sulfonamide.
\end{lstlisting}
\end{tcolorbox}

\begin{tcolorbox}
\textbf{ChemLLM:}
\begin{lstlisting}
An antimalarial drug used in the treatment and prevention (PROPHYLAXIS, DRUG ) OF MALARIA. It has also been given orally as an anti-amoebic agent for amo ebiasis." - [N-(4-chlorophenyl)-5-nitroso-N'-methylbenzimidazolone] is not expected under standard conditions to give riseto such reaction with O=[Cr](+3)[F].[Br][Cu+].Cl[Hg]>CO>meclozinil.[I]. – The mechanism by which chloroguanide inhibits plasmodial enzymes was investigated using purified fumarate reductase from Plasmomonadoides malariae strain FCR39/IBM7A6E11B10C13D17J22K25L29P31R34S37T40V43W46X49Y52Z55AA56AB57
\end{lstlisting}
\end{tcolorbox}

\begin{tcolorbox}
\textbf{MolInstructions\_molecule:}
\begin{lstlisting}
An antiprotozoal agent used in the treatment of giardiasis and amebiasis." It has a role as an antiprotozoal drug and an antiparasitic agent. It is a member of benzothiadiazines and an organochlorine compound. It is functionally related to a 1,1'-[(3,5-dichloro-4-hydroxyphenyl)imino]diethylsulfanilamide. It is a conjugate acid of a metronidazole(1-).  It derives from a hydride of a 1,3,4-thiadiazole.  It is an enantiomer of a (R)-metronidazole.  It is a tautomer of a metronidazole zwitterion.  It is a conjugate base of a (S)-metronidazole.  It is a conjugate acid of a (S)-metronidazole(1-).  It is a tautomer of a (S)-metronidazole zwitterion.  It is a conjugate base of a (R)-metronidazole.  It is a conjugate acid of a (R)-metronidazole(1-).  It is a tautomer of a (R)-metronidazole zwitterion.  It is a conjugate base of a (S)-metronidazole.  It is a conjugate acid of a (S)-metronidazole(1-).  It is a tautomer of a (S)-metronidazole zwitterion.  It is a conjugate base of a (R)-metronidazole.  It is a conjugate acid of a (R)-metronidazole(1-).  It is a tautomer of a (R)-metronidazole zwitterion.  It is a conjugate base of a (S)-metronidazole.  It is a conjugate acid of a (S)-metronidazole(1-).  It is a tautomer of a (S)-metronidazole zwitterion.  It is a conjugate base of a (R)-metronidazole.  It is a conjugate acid of a (R)-metronidazole(1-).  It is a tautomer of a (R)-metronidazole zwitterion.
\end{lstlisting}
\end{tcolorbox}

\begin{tcolorbox}
\textbf{BlankCaption:}
\begin{lstlisting}

\end{lstlisting}
\end{tcolorbox}

\begin{tcolorbox}
\textbf{BBBP Oracle:}
\begin{lstlisting}
False
\end{lstlisting}
\end{tcolorbox}

\begin{tcolorbox}
\textbf{GPT-4o-Frags-Generic:}
\begin{lstlisting}
Let's break down the provided molecular fragments and discuss their functionalities and relevant applications.

### Fragment 1: `[13*]C1Nc2cc(Cl)c(S(N)(=O)=O)cc2S(=O)(=O)N1`

This fragment is quite complex, and it contains several important functional groups:

1. **Aromatic Ring (Benzene Ring)**: The central part of the fragment is a benzene ring. Aromatic systems like benzene rings are known for their stability and play a crucial role in many chemical applications.
2. **Chlorine Substituent (Cl)**: The chlorine atom attached to the benzene ring increases the electrophilicity of the ring due to its electron-withdrawing nature. This can make the ring more reactive toward nucleophilic aromatic substitution reactions.
3. **Sulfonamide Group (S(N)(=O)=O)**: The sulfonamide group is a combination of a sulfonyl group (–SO2) and an amine (–NH2). Sulfonamides are known for their medicinal properties and can act as antibiotics, antidiabetics, or in other therapeutic roles.
4. **Sulfonyl Group (S(=O)(=O)N1)**: This group is another sulfonamide moiety, suggesting a linkage to another nitrogen atom, contributing to the molecule's potential biological activity or chemical reactivity.

**Applications**: This type of structure can be common in the pharmaceutical industry, where the multiple functional groups allow for a range of biological activities. The presence of sulfonamides suggests potential use as antibiotic agents or other medicinal compounds.

### Fragment 2: `[8*]C(Cl)Cl`

This is a simpler fragment characterized as follows:

1. **Alkyl Chloride (C(Cl)Cl)**: This fragment is a simple dichloromethane (CH2Cl2), but with an indication that it is labeled with some form of isotopic or special designation at carbon **\([8*]\)**.

**Applications**:
- **Solvent**: Dichloromethane is a well-known solvent used in many chemical reactions and industrial processes due to its ability to dissolve a wide range of compounds.
- **Intermediate in Synthesis**: It can be used as an intermediate chemical in the synthesis of more complex compounds.
- **Isotopic Labeling**: The notation `[*]` might suggest specific isotopic labeling (such as Carbon-13 or Carbon-14), which is often used in studying reaction mechanisms through tracing atoms or in NMR (nuclear magnetic resonance) spectroscopy for detailed molecular analysis.

### Summary:
The first molecular fragment contains multiple functional groups, including aromatic systems and sulfonamides, making it relevant in pharmaceutical applications. The second fragment is dichloromethane, commonly used as a versatile solvent or reagent in chemical synthesis. If the notation `[13*]` and `[8*]` refers to isotopic labeling, this can be utilized for tracing reactions or in detailed spectroscopic studies.
\end{lstlisting}
\end{tcolorbox}

\begin{tcolorbox}
\textbf{GPT-4o-Generic:}
\begin{lstlisting}
The molecule you provided is:

NS(=O)(=O)c1cc2c(cc1Cl)NC(C(Cl)Cl)NS2(=O)=O

To better understand this molecule, let's break down its structure and functionality:

1. **Sulfonamide Groups (NS(=O)(=O))**:
   - The molecule contains two sulfonamide groups, evident from the "NS(=O)(=O)" notation.
   - Sulfonamides are known for their role in antibiotics and other pharmaceutical applications due to their ability to mimic the structure of para-aminobenzoic acid (PABA) and inhibit bacterial growth by targeting dihydropteroate synthase.

2. **Chlorinated Benzene Ring**:
   - The presence of a benzene ring with a chlorine substituent (Cl) indicates aromaticity and potential for electronic interactions.
   - Chlorine atoms can contribute to the molecule's electrophilic properties and reactivity, making it relevant in medicinal chemistry and agrochemical applications.

3. **Additional Chlorine Substituents**:
   - Beyond the benzene ring, the molecule has two additional chlorine atoms attached to a carbon (C(Cl)Cl).
   - These groups can influence the molecule’s lipophilicity, stability, and its ability to interact with biological macromolecules.

4. **Heterocyclic Structure**:
   - The “NC(C(Cl)Cl)NS2(=O)=O” part of the molecule suggests the presence of a heterocyclic system, possibly a bicyclic structure.
   - Heterocyclic compounds are crucial in medicinal chemistry as they often exhibit biological activity and can serve as scaffolds for drug development.

### Relevant Applications:

Given the functionalities present in this molecule, here are some potential applications:

- **Pharmaceuticals**: 
  - The sulfonamide functionality suggests potential use in antimicrobial or antibacterial agents. 
  - The chlorinated aromatic ring and additional chlorine atoms could help in designing drugs that target specific receptors or enzymes in the body.

- **Agrochemicals**:
  - Sulfonamides are also present in some herbicides and fungicides. The chlorinated structure could enhance its effectiveness and stability in agricultural applications.

- **Chemical Synthesis**:
  - The molecule's reactivity due to chlorine and sulfonamide groups might render it a useful intermediate in synthetic chemistry for the development of more complex molecules.

### Summary:

This molecule’s distinct features, including the sulfonamide groups, chlorinated benzene ring, and the heterocyclic structure, suggest that it could have significant uses in pharmaceuticals and agrochemicals. The chlorinated substituents enhance its stability and biological interaction, making it a versatile compound for further research and development in these fields.
\end{lstlisting}
\end{tcolorbox}

\begin{tcolorbox}
\textbf{Gemma2-27B-Bio:}
\begin{lstlisting}
Sulfonamide

**Biological Properties and Functionality:**

The molecule you provided is a sulfonamide. Sulfonamides are a class of synthetic antimicrobial agents that have been widely used to treat bacterial infections.

**Mechanism of Action:**

Sulfonamides work by inhibiting the enzyme dihydropteroate synthase (DHPS), which is essential for the synthesis of folic acid in bacteria. Folic acid is a vital nutrient that bacteria need for growth and reproduction. By blocking DHPS, sulfonamides prevent bacteria from producing folic acid, leading to their death.

**Biological Properties:**

* **Antibacterial activity:** Sulfonamides are effective against a wide range of gram-positive and gram-negative bacteria.
* **Bacteriostatic:** Sulfonamides are bacteriostatic, meaning they inhibit bacterial growth rather than killing them directly.

**Applications:**

* **Treatment of bacterial infections:** Sulfonamides have been used to treat a variety of bacterial infections, including urinary tract infections, respiratory infections, skin infections, and eye infections.
* **Prophylaxis:** Sulfonamides can be used to prevent bacterial infections in individuals at high risk, such as those undergoing surgery or with weakened immune systems.

**Limitations:**

* **Resistance:** Bacteria can develop resistance to sulfonamides, limiting their effectiveness.
* **Side effects:** Sulfonamides can cause side effects such as allergic reactions, nausea, vomiting, and diarrhea.

**Note:**

The specific biological properties and applications of a sulfonamide can vary depending on its chemical structure.

**Important:**

This information is for educational purposes only and should not be considered medical advice. Always consult with a healthcare professional for diagnosis and treatment.

\end{lstlisting}
\end{tcolorbox}

\begin{tcolorbox}
\textbf{Gemma2-27B-Chem:}
\begin{lstlisting}
**

Let's break down the molecule and discuss its properties and potential applications.

**Understanding the Structure**

The molecule you provided is a complex organic compound containing several functional groups:

* **Sulfonamide:** The -S(=O)(=O)N- group is a sulfonamide, known for its biological activity and use in pharmaceuticals.
* **Aromatic Ring:** The "c1cc2c(cc1Cl)..." part indicates an aromatic ring with a chlorine substituent. Aromatic rings contribute to stability and can participate in various reactions.
* **Chlorine Substituents:** The presence of chlorine atoms (Cl) can influence the molecule's reactivity and lipophilicity (its ability to dissolve in fats).
* **Amide:** The -NC(C(Cl)Cl)NS2(=O)=O group contains an amide linkage (-CON-) and another sulfonamide group.

**Potential Chemical Properties**

* **Polarity:** The molecule is likely polar due to the presence of electronegative atoms like oxygen, nitrogen, and chlorine. This polarity can influence its solubility in water and other solvents.
* **Hydrogen Bonding:** The sulfonamide and amide groups can participate in hydrogen bonding, both as donors and acceptors. This property is important for interactions with biological targets.
* **Reactivity:** The sulfonamide and amide groups can undergo various reactions, such as nucleophilic substitution, hydrolysis, and condensation reactions.
* **Stability:** The aromatic ring and sulfonamide groups contribute to the molecule's overall stability.

**Potential Applications**

Given its structure and functional groups, this molecule could have potential applications in:

* **Pharmaceuticals:** Sulfonamides are widely used as antibiotics and have other medicinal properties. This molecule could be investigated for antibacterial, antiviral, or anti-inflammatory activity.
* **Agrochemicals:** Sulfonamides are also used as herbicides and fungicides. This molecule might have potential as a crop protection agent.
* **Materials Science:** The molecule's polarity and potential for hydrogen bonding could make it useful in the development of new materials, such as polymers or coatings.

**Important Note:**

It's crucial to remember that these are just potential applications based on the molecule's structure. Further research and experimentation are necessary to determine its actual properties and suitability for specific uses.

**Additional Information**

To gain a more complete understanding of this molecule, it would be helpful to know:

* **Its full name and IUPAC
\end{lstlisting}
\end{tcolorbox}

\begin{tcolorbox}
\textbf{Gemma2-27B-Drug:}
\begin{lstlisting}
<h2>

Let's break down this molecule and explore its potential as a drug.

**1. Structural Analysis:**

* **Scaffold:** The molecule features a sulfonamide group (-SO2NH-) which is a common pharmacophore found in many drugs.
* **Chlorine Atoms:** The presence of three chlorine atoms suggests potential for lipophilicity (ability to dissolve in fats) and possible interactions with specific protein targets.
* **Sulfonylurea Moiety:** The -SO2-NH-C(=O)- group resembles a sulfonylurea, a class of drugs known for their hypoglycemic (blood sugar lowering) effects.

**2. Potential Mechanism of Action:**

* **Sulfonylurea-like Activity:** Given the structural similarity to sulfonylureas, this molecule *might* act by binding to and inhibiting ATP-sensitive potassium channels (KATP channels) in pancreatic beta cells. This inhibition would lead to increased insulin secretion, potentially lowering blood glucose levels.

* **Other Possibilities:** The molecule's structure could also allow for interactions with other targets, such as enzymes or receptors involved in inflammatory pathways, depending on the specific arrangement of its functional groups.

**3. Potential Class and Target:**

* **Class:** Based on the sulfonylurea-like structure, this molecule *could* be classified as a potential antidiabetic agent.
* **Target:**  The primary target *might* be the sulfonylurea receptor (SUR1) subunit of KATP channels in pancreatic beta cells. However, further investigation is needed to confirm this.

**4. Important Considerations:**

* **Toxicity:** The presence of chlorine atoms raises concerns about potential toxicity. Thorough toxicological studies would be essential.
* **Specificity:** It's crucial to determine if this molecule selectively targets the desired pathway or if it interacts with other targets, leading to unwanted side effects.
* **Pharmacokinetics:** Understanding how the molecule is absorbed, distributed, metabolized, and excreted in the body is crucial for drug development.

**Disclaimer:** This analysis is purely speculative based on the molecule's structure. Experimental validation is necessary to confirm its actual properties, mechanism of action, and potential as a drug.

\end{lstlisting}
\end{tcolorbox}

\begin{tcolorbox}
\textbf{Gemma2-27B-Frags-Bio:}
\begin{lstlisting}

\end{lstlisting}
\end{tcolorbox}

\begin{tcolorbox}
\textbf{Gemma2-27B-Frags-Chem:}
\begin{lstlisting}

\end{lstlisting}
\end{tcolorbox}

\begin{tcolorbox}
\textbf{Gemma2-27B-Frags-Drug:}
\begin{lstlisting}

\end{lstlisting}
\end{tcolorbox}

\begin{tcolorbox}
\textbf{Gemma2-27B-Frags-Generic:}
\begin{lstlisting}

\end{lstlisting}
\end{tcolorbox}

\begin{tcolorbox}
\textbf{Gemma2-27B-Frags-Quant:}
\begin{lstlisting}
* **[13*]C1Nc2cc(Cl)c(S(N)(=O)=O)cc2S(=O)(=O)N1**: This fragment contains a carbon-13 isotope, a nitrogen atom, chlorine atoms, sulfur atoms, and oxygen atoms.

* **[8*]C(Cl)Cl**: This fragment contains a carbon-8 isotope, chlorine atoms.

Let's break down the quantum properties and potential applications of each fragment:

**Fragment 1: [13*]C1Nc2cc(Cl)c(S(N)(=O)=O)cc2S(=O)(=O)N1**

* **Quantum Properties:**
    * **Nuclear Spin:** The carbon-13 isotope (¹³C) has a nuclear spin of 1/2, making it NMR active. This allows for the use of ¹³C NMR spectroscopy to study the structure and dynamics of molecules containing this fragment.
    * **Electronic Structure:** The presence of nitrogen, sulfur, and oxygen atoms, along with the conjugated system of double bonds, creates a complex electronic structure with potential for interesting optical and electronic properties.

* **Possible Functionality and Applications:**
    * **Pharmaceuticals:** The fragment's structure suggests potential as a building block for pharmaceuticals, particularly those targeting enzymes or biological pathways involving sulfur-containing molecules.
    * **Materials Science:** The conjugated system and heteroatoms could be incorporated into materials with unique optical or electronic properties, such as organic semiconductors or fluorescent probes.

**Fragment 2: [8*]C(Cl)Cl**

* **Quantum Properties:**
    * **Nuclear Spin:** Carbon-8 (⁸C) is a radioactive isotope with a very short half-life. Its nuclear properties are not typically exploited in conventional applications.

* **Possible Functionality and Applications:**
    * **Radioactive Tracing:** Due to its radioactivity, carbon-8 could potentially be used as a tracer in very specific research applications, such as studying metabolic pathways or tracking the movement of molecules in biological systems. However, its short half-life limits its practical use.

**Important Considerations:**

* **Isotope Effects:** The presence of isotopes like ¹³C and ⁸C can lead to subtle but measurable differences in chemical and physical properties compared to the naturally occurring isotopes. These isotope effects can be exploited in various analytical and research techniques.
* **Chemical Reactivity:** The specific chemical reactivity of these fragments will
\end{lstlisting}
\end{tcolorbox}

\begin{tcolorbox}
\textbf{Gemma2-27B-Generic:}
\begin{lstlisting}
**

Let's break down this molecule and discuss its potential properties and applications.

**Understanding the Structure**

The molecule you provided, NS(=O)(=O)c1cc2c(cc1Cl)NC(C(Cl)Cl)NS2(=O)=O, is a complex organic compound with several functional groups:

* **Sulfonamide (NS(=O)(=O)):** This group is characterized by a sulfur atom bonded to a nitrogen atom, which is further bonded to two oxygen atoms (forming a sulfonyl group). Sulfonamides are known for their biological activity and are often found in pharmaceuticals.
* **Aromatic Ring:** The "c1cc2c(cc1Cl) " part indicates a benzene ring with a chlorine atom attached. Aromatic rings contribute to stability and can participate in various chemical reactions.
* **Amine (N):** The nitrogen atom connected to the aromatic ring and the sulfonamide group. Amines are basic and can form hydrogen bonds.
* **Chlorine Atoms (Cl):** The presence of chlorine atoms can influence the molecule's reactivity and lipophilicity (its ability to dissolve in fats).

**Potential Functionality and Applications**

Based on the structure, here are some potential functionalities and applications of this molecule:

* **Pharmaceutical Applications:**

    * **Antibacterial/Antimicrobial:** Sulfonamides are well-known for their antibacterial properties. This molecule could potentially be investigated as a new antibiotic or antimicrobial agent.
    * **Enzyme Inhibition:** The sulfonamide group can act as a pharmacophore, meaning it can bind to specific enzymes and inhibit their activity. This could be useful in developing drugs targeting specific diseases.
* **Agricultural Applications:**

    * **Herbicide/Pesticide:** The molecule's structure might make it suitable for targeting specific plant or insect enzymes, leading to potential applications as a herbicide or pesticide.
* **Material Science:**

    * **Polymer Modification:** The molecule's reactive groups could be used to modify polymers, potentially altering their properties like solubility, strength, or conductivity.

**Important Considerations**

* **Toxicity:** It's crucial to assess the toxicity of this molecule before any practical applications. The presence of chlorine atoms and the sulfonamide group could raise concerns about potential toxicity.
* **Synthesis:** Synthesizing this molecule efficiently and cost-effectively would be essential for any practical application.
* **Specificity:** Further research is needed to determine the
\end{lstlisting}
\end{tcolorbox}

\begin{tcolorbox}
\textbf{Gemma2-27B-Quant:}
\begin{lstlisting}
**

Let's break down the molecule and discuss its potential quantum properties and applications.

**Understanding the Molecule**

The molecule you provided, NS(=O)(=O)c1cc2c(cc1Cl)NC(C(Cl)Cl)NS2(=O)=O, is a complex organic compound with several interesting structural features:

* **Sulfonamide Group:** The presence of the -S(=O)(=O)- group (sulfonamide) suggests potential for hydrogen bonding and interactions with other polar molecules.
* **Aromatic Ring:** The benzene ring (c1cc2c(cc1Cl)) provides stability and can participate in π-π stacking interactions.
* **Chlorine Substituents:** The chlorine atoms (Cl) can influence the molecule's polarity, reactivity, and solubility.

**Potential Quantum Properties**

While predicting specific quantum properties without advanced computational analysis is difficult, we can speculate on some possibilities based on the structure:

* **Electronic Structure:** The molecule's extended π-system from the aromatic ring and the presence of heteroatoms (N, S, O) could lead to interesting electronic transitions and potential for light absorption/emission in the UV-Vis range.
* **Vibrational Modes:** The molecule's complex structure will have a rich set of vibrational modes, which can be probed using techniques like infrared (IR) and Raman spectroscopy. These vibrations can provide insights into the molecule's bonding and dynamics.
* **Spin Properties:** Depending on the arrangement of electrons, the molecule might exhibit paramagnetic or diamagnetic behavior.

**Relevant Functionality and Applications**

The specific functionality and applications of this molecule are highly dependent on its overall properties, which would need to be determined experimentally. However, based on its structural features, here are some possibilities:

* **Pharmaceuticals:** Sulfonamides are common in pharmaceuticals due to their antibacterial properties. This molecule could be investigated for potential biological activity.
* **Materials Science:** The molecule's aromatic ring and potential for π-π stacking could make it useful in the development of organic semiconductors or other functional materials.
* **Catalysis:** The presence of heteroatoms and the molecule's overall structure might allow it to act as a ligand in metal complexes, potentially leading to catalytic applications.

**Important Note:**

It's crucial to emphasize that these are just speculations based on the molecule's structure. To fully understand its quantum properties and potential applications, detailed experimental and computational
\end{lstlisting}
\end{tcolorbox}

\begin{tcolorbox}
\textbf{Gemma2-9B-Bio:}
\begin{lstlisting}
```
NS(=O)(=O)c1cc2c(cc1Cl)NC(C(Cl)Cl)NS2(=O)=O
```

This SMILES string represents **Chlorambucil**.

## Biological Properties and Applications of Chlorambucil:

**Biological Properties:**

* **Alkylating Agent:** Chlorambucil is a nitrogen mustard alkylating agent. This means it works by attaching alkyl groups (specifically, chloroethyl groups) to DNA, disrupting its structure and function. This alkylation primarily targets the N7 position of guanine bases.
* **Cytotoxic:** Due to its DNA-damaging effects, chlorambucil is cytotoxic, meaning it kills cells. This property makes it useful in cancer treatment.
* **Immunosuppressant:** Chlorambucil can also suppress the immune system by interfering with the proliferation and function of lymphocytes.

**Applications:**

* **Cancer Treatment:** Chlorambucil is primarily used in the treatment of various types of cancer, including:
    * **Hodgkin's lymphoma:** It is a cornerstone treatment for this type of lymphoma.
    * **Non-Hodgkin's lymphoma:** It can be used in the treatment of certain subtypes.
    * **Chronic lymphocytic leukemia:** It is sometimes used in combination with other drugs.
* **Autoimmune Diseases:** Chlorambucil's immunosuppressive properties make it useful in managing certain autoimmune diseases, such as:
    * **Rheumatoid arthritis:** It can help reduce inflammation and joint damage.
    * **Systemic lupus erythematosus:** It can suppress the overactive immune response.

**Important Considerations:**

* **Side Effects:** Chlorambucil can cause significant side effects, including:
    * **Bone marrow suppression:** This can lead to decreased production of blood cells, increasing the risk of infections, anemia, and bleeding.
    * **Nausea and vomiting:** These are common side effects.
    * **Hair loss:** Temporary hair loss is common.
    * **Increased risk of secondary cancers:** Long-term use of chlorambucil can increase the risk of developing other cancers.
* **Monitoring:** Patients receiving chlorambucil require close monitoring of blood counts and other vital signs.




\end{lstlisting}
\end{tcolorbox}

\begin{tcolorbox}
\textbf{Gemma2-9B-Chem:}
\begin{lstlisting}


* **Chemical Properties:**
    * **Structure:** The molecule is a complex organic compound containing sulfur, nitrogen, chlorine, and oxygen atoms. It features a central sulfur atom double-bonded to two oxygen atoms (forming a sulfone group), a nitrogen atom, and a chlorine atom. The molecule also has two phenyl rings connected to the sulfur atom and a central carbon atom with two chlorine atoms attached.
    * **Polarity:** The molecule is highly polar due to the presence of electronegative atoms like oxygen, nitrogen, and chlorine. The sulfone group is particularly polar, contributing significantly to the overall polarity.
    * **Solubility:** The polarity of the molecule suggests that it is likely soluble in polar solvents like water and alcohols, but less soluble in non-polar solvents like hexane.
    * **Reactivity:** The sulfone group is relatively unreactive, but the chlorine atoms and nitrogen atom can participate in various chemical reactions. The molecule could potentially undergo nucleophilic substitution reactions at the chlorine atoms or electrophilic aromatic substitution reactions on the phenyl rings.
* **Functionality:**
    * **Sulfone Group:** The sulfone group (-SO2-) is a strong electron-withdrawing group, which can influence the reactivity of the molecule. It also contributes to the molecule's stability and polarity.
    * **Chlorine Atoms:** The chlorine atoms are good leaving groups and can participate in nucleophilic substitution reactions.
    * **Nitrogen Atom:** The nitrogen atom can act as a nucleophile or participate in hydrogen bonding.
* **Applications:**
    * **Pharmaceuticals:** The molecule's complex structure and functional groups make it a potential candidate for drug development. The sulfone group is often found in pharmaceuticals due to its biological activity and metabolic stability.
    * **Agrochemicals:** The molecule's potential for biological activity could also make it useful as an agrochemical, such as a pesticide or herbicide.
    * **Materials Science:** The molecule's unique structure and properties could be explored for applications in materials science, such as in the development of new polymers or coatings.



It's important to note that these are just potential properties and applications based on the molecule's structure. Further experimental studies would be needed to confirm these predictions and explore the molecule's full potential.

\end{lstlisting}
\end{tcolorbox}

\begin{tcolorbox}
\textbf{Gemma2-9B-Drug:}
\begin{lstlisting}
*I am not a medical professional and this information is for educational purposes only.*

*Please note that I cannot provide medical advice. This information is for educational purposes only and should not be considered a substitute for professional medical advice.*

Let's break down the structure and discuss some potential properties:

**Structure Analysis:**

* **Sulfonamide Group (-SO2NH-):** This is a prominent feature, often found in antibacterial drugs. It can act as a hydrogen bond donor and acceptor, potentially interacting with enzymes.
* **Chlorine Atoms (Cl):**  These are electronegative atoms that can influence the molecule's lipophilicity (fat solubility) and potentially contribute to its binding affinity to targets.
* **Aromatic Ring (c1cc2c(cc1Cl)NC(C(Cl)Cl)NS2(=O)=O):** This ring system provides a planar structure that could interact with hydrophobic pockets in target proteins.

**Potential Properties and Speculation:**

Based on the structure, here are some educated guesses about the molecule's potential properties:

* **Mechanism of Action:**
    * **Inhibition of Bacterial Enzymes:** The sulfonamide group strongly suggests a possible mechanism involving inhibition of bacterial enzymes, particularly those involved in folic acid synthesis (like dihydropteroate synthase). This is a common mechanism for sulfonamide antibiotics.
* **Class:**
    * **Sulfonamide Antibiotic:** Given the sulfonamide group, this molecule is a strong candidate for belonging to this class of drugs.
* **Target:**
    * **Bacterial Enzymes:**  The most likely targets are enzymes involved in essential metabolic pathways in bacteria, such as folic acid synthesis.

**Important Considerations:**

* **Specificity:**  It's crucial to remember that this is just a structural analysis. Determining the actual target and mechanism of action would require extensive experimental testing. 
* **Toxicity:** The presence of chlorine atoms raises concerns about potential toxicity.  
* **Pharmacokinetics:** Factors like absorption, distribution, metabolism, and excretion (ADME) would need to be studied to assess the molecule's suitability as a drug.


Let me know if you have any other molecules you'd like to explore!

\end{lstlisting}
\end{tcolorbox}

\begin{tcolorbox}
\textbf{Gemma2-9B-Frags-Bio:}
\begin{lstlisting}


* **Fragment 1:** {'[13*]C1Nc2cc(Cl)c(S(N)(=O)=O)cc2S(=O)(=O)N1'}

This fragment appears to represent a molecule containing:

* **A benzene ring:** The 'c1cc2cc(Cl)c2' portion suggests a benzene ring with a chlorine atom substituent.
* **A sulfonamide group:** The 'S(=O)(=O)N1' portion indicates a sulfonamide group, which is known for its biological activity.

**Biological Properties and Functionality:**

* **Antibacterial activity:** Sulfonamides are well-known for their antibacterial properties, inhibiting bacterial growth by interfering with folic acid synthesis.
* **Anti-inflammatory activity:** Some sulfonamides exhibit anti-inflammatory effects.
* **Anticancer activity:** Certain sulfonamide derivatives have shown promising anticancer activity.
* **Drug development:** This fragment could serve as a starting point for developing new drugs targeting bacterial infections, inflammation, or cancer.

**Fragment 2:** {'[8*]C(Cl)Cl'}

This fragment represents a simple molecule with:

* **A central carbon atom:** The 'C' atom is the central point of the molecule.
* **Two chlorine atoms:** The 'Cl' atoms are attached to the central carbon atom.

**Biological Properties and Functionality:**

* **Limited biological activity:** This fragment is likely to have limited biological activity on its own due to its simple structure.
* **Building block:** It could potentially serve as a building block for more complex molecules with biological activity.

**Applications:**

* **Synthesis of other compounds:** This fragment could be used as a starting material for synthesizing more complex molecules with desired biological properties.

**Note:**

It's important to remember that these are just predictions based on the structural information provided. The actual biological properties and applications of these fragments would need to be experimentally determined.



\end{lstlisting}
\end{tcolorbox}

\begin{tcolorbox}
\textbf{Gemma2-9B-Frags-Chem:}
\begin{lstlisting}


* **Fragment 1:** {'[13*]C1Nc2cc(Cl)c(S(N)(=O)=O)cc2S(=O)(=O)N1'}

This fragment represents a complex molecule with several functional groups:

* **Pyridine ring:** The 'Nc2cc(Cl)c2' portion represents a pyridine ring, a six-membered aromatic heterocycle containing one nitrogen atom. Pyridine is known for its basicity, ability to participate in hydrogen bonding, and its role as a ligand in coordination chemistry.
* **Sulfonamide group:** The 'S(N)(=O)=O' portion represents a sulfonamide group. Sulfonamides are known for their biological activity, particularly as antibacterial agents. They can also act as hydrogen bond donors and acceptors.
* **Chlorine atom:** The 'Cl' atoms contribute to the molecule's polarity and reactivity. They can participate in nucleophilic substitution reactions and influence the molecule's overall electronic properties.

**Possible functionalities and applications:**

* **Pharmaceuticals:** The combination of pyridine and sulfonamide groups suggests potential applications in drug development, particularly as antibacterial or antifungal agents.
* **Agrochemicals:** Sulfonamides are also used in agrochemicals as herbicides and fungicides.
* **Catalysis:** The pyridine ring can act as a ligand in metal complexes, potentially leading to applications in catalysis.

* **Fragment 2:** {'[8*]C(Cl)Cl'}

This fragment represents a simple molecule with two chlorine atoms attached to a central carbon atom.

* **Reactivity:** The two chlorine atoms make this fragment highly reactive due to the electron-withdrawing nature of chlorine. It can readily undergo nucleophilic substitution reactions.

**Possible functionalities and applications:**

* **Organic synthesis:** This fragment can be used as a starting material for the synthesis of more complex molecules.
* **Chlorination reactions:** The high reactivity of the chlorine atoms makes this fragment useful for introducing chlorine atoms into other molecules.



Let me know if you have any other questions.


\end{lstlisting}
\end{tcolorbox}

\begin{tcolorbox}
\textbf{Gemma2-9B-Frags-Drug:}
\begin{lstlisting}
* **Fragment 1:** '[13*]C1Nc2cc(Cl)c(S(N)(=O)=O)cc2S(=O)(=O)N1'
* **Fragment 2:** '[8*]C(Cl)Cl'

Let's break down the fragments and discuss their potential properties:

**Fragment 1:** '[13*]C1Nc2cc(Cl)c(S(N)(=O)=O)cc2S(=O)(=O)N1'

* **Structure:** This fragment appears to be a heterocyclic compound with multiple functional groups. 
    * **Pyridine ring:** The "Nc2cc(Cl)c" portion suggests a pyridine ring, a six-membered aromatic ring with one nitrogen atom.
    * **Sulfonamide group:** The "S(N)(=O)=O" portion indicates a sulfonamide group, which is known for its biological activity and often found in drugs.
    * **Chlorine atom:** The "Cl" atoms contribute to the molecule's lipophilicity and potential for interactions with biological targets.

* **Potential Properties:**
    * **Mechanism of action:** The presence of the sulfonamide group suggests potential inhibition of enzymes, particularly those involved in protein synthesis (e.g., dihydropteroate synthase, a target for sulfonamide antibiotics).
    * **Class:** Based on the sulfonamide group, this fragment might belong to the class of sulfonamides or related compounds.
    * **Target:**  The specific target would depend on the precise structure and interactions with potential binding sites. However, given the sulfonamide group, enzymes involved in protein synthesis or other metabolic pathways could be potential targets.

**Fragment 2:** '[8*]C(Cl)Cl'

* **Structure:** This fragment is a simple molecule consisting of a central carbon atom bonded to two chlorine atoms.

* **Potential Properties:**
    * **Mechanism of action:** This fragment is unlikely to have a specific mechanism of action on its own. It might act as a lipophilic group, influencing the overall properties of the molecule.
    * **Class:** This fragment doesn't fit into a specific drug class.
    * **Target:**  It's unlikely to have a specific target.

**Overall Considerations:**

* **Combination:** The combination of these fragments might create a molecule with enhanced properties. The pyridine ring and sulfonamide group could
\end{lstlisting}
\end{tcolorbox}

\begin{tcolorbox}
\textbf{Gemma2-9B-Frags-Generic:}
\begin{lstlisting}


* **Fragment 1:** `[13*]C1Nc2cc(Cl)c(S(N)(=O)=O)cc2S(=O)(=O)N1`

This fragment represents a **sulfonamide** derivative. 

**Functionality:**

* **Sulfonamide group (-SO2NH-):** This group is known for its strong hydrogen bonding ability, making it useful in pharmaceutical applications. It can interact with biological targets like enzymes and receptors.
* **Aromatic ring (benzene):** The benzene ring provides stability and can participate in π-π interactions with other aromatic systems.
* **Chlorine atoms (Cl):** Chlorine atoms can influence the molecule's lipophilicity and metabolic stability.

**Applications:**

* **Pharmaceuticals:** Sulfonamides are widely used as antibacterial agents, diuretics, and anticonvulsants. This specific fragment could be a building block for developing new drugs targeting various diseases.
* **Agrochemicals:** Sulfonamides are also used as herbicides and fungicides.
* **Materials science:** Sulfonamides can be incorporated into polymers and other materials to modify their properties.

* **Fragment 2:** `[8*]C(Cl)Cl`

This fragment represents a **dichloroethane** derivative.

**Functionality:**

* **Chlorine atoms (Cl):** The two chlorine atoms make this molecule highly reactive and susceptible to nucleophilic substitution reactions.

**Applications:**

* **Chemical synthesis:** Dichloroethane is a versatile intermediate in organic synthesis, used to prepare various compounds like vinyl chloride (used in PVC production) and ethylene glycol (used in antifreeze).
* **Solvent:** Dichloroethane can be used as a solvent for certain reactions and processes.
* **Pesticide:** Some dichloroethane derivatives are used as pesticides.

**Note:** It's important to remember that these are just general properties and applications based on the fragments provided. The specific properties and applications of a molecule depend on its complete structure and the context in which it is used.

\end{lstlisting}
\end{tcolorbox}

\begin{tcolorbox}
\textbf{Gemma2-9B-Frags-Quant:}
\begin{lstlisting}


* **Fragment 1:** '[13*]C1Nc2cc(Cl)c(S(N)(=O)=O)cc2S(=O)(=O)N1'

This fragment represents a heterocyclic compound containing a fused ring system with nitrogen, carbon, sulfur, and chlorine atoms. 

**Quantum Properties:**

* **Electronic Structure:** The presence of heteroatoms like nitrogen and sulfur introduces lone pairs of electrons, leading to a complex electronic structure with potential for delocalization and resonance. This can influence its reactivity and spectroscopic properties.
* **Spin State:** Depending on the number of unpaired electrons, the molecule could exhibit different spin states (singlet, doublet, triplet, etc.). This can affect its magnetic properties and interactions with magnetic fields.
* **Optical Properties:** The conjugated system of pi electrons in the ring can lead to absorption and emission of light in the visible or ultraviolet region. This makes it potentially useful for applications like fluorescent probes or dyes.
* **Vibrational Modes:** The molecule possesses various vibrational modes associated with stretching, bending, and twisting of its bonds. These modes can be studied using infrared and Raman spectroscopy to gain insights into its structure and dynamics.

**Functionality and Applications:**

* **Pharmaceuticals:** Heterocyclic compounds are prevalent in pharmaceuticals due to their diverse biological activities. This fragment could serve as a building block for drug development, targeting specific receptors or enzymes.
* **Agrochemicals:** The presence of sulfur and chlorine suggests potential insecticidal or fungicidal properties.
* **Materials Science:** The molecule's optical properties could be exploited in the development of new materials for optoelectronic devices or sensors.

* **Fragment 2:** '[8*]C(Cl)Cl'

This fragment represents a simple molecule with a central carbon atom bonded to two chlorine atoms.

**Quantum Properties:**

* **Electronic Structure:** The carbon atom has four valence electrons, and each chlorine atom contributes one electron to the bonding. This results in a relatively simple electronic structure with localized bonding.
* **Spin State:** The molecule is likely in a singlet spin state due to the absence of unpaired electrons.
* **Optical Properties:** The molecule is unlikely to exhibit significant absorption or emission of light in the visible or ultraviolet region due to the lack of extended conjugation.
* **Vibrational Modes:** The molecule possesses vibrational modes associated with stretching and bending of the C-Cl bonds.

**Functionality and Applications:**

* **Chlorination Reagent:** The molecule can
\end{lstlisting}
\end{tcolorbox}

\begin{tcolorbox}
\textbf{Gemma2-9B-Generic:}
\begin{lstlisting}


* **Structure:** The molecule is a sulfonamide derivative with two chlorine atoms and two nitro groups.

* **Functionality:**

    * **Sulfonamide group:** This group is known for its biological activity and is often found in pharmaceuticals. It can act as a hydrogen bond donor and acceptor, contributing to its binding affinity to target proteins.
    * **Nitro groups:** These groups are electron-withdrawing and can influence the molecule's reactivity and biological activity. They can also contribute to the molecule's stability.
    * **Chlorine atoms:** These atoms are also electron-withdrawing and can further influence the molecule's reactivity and biological activity.

**Applications:**

Given the presence of these functional groups, the molecule likely has potential applications in the following areas:

* **Pharmaceuticals:** Sulfonamides are widely used as antibacterial agents, diuretics, and anticonvulsants. The presence of the nitro groups and chlorine atoms could enhance the molecule's activity or selectivity towards specific targets.
* **Agrochemicals:** Sulfonamides are also used as herbicides and fungicides. The molecule's structure suggests it could have similar applications.
* **Materials science:** The molecule's unique structure and functional groups could be explored for applications in materials science, such as in the development of new polymers or coatings.

**Further research:**

To fully understand the molecule's properties and potential applications, further research is needed. This could include:

* **Synthesis:** Determining the optimal synthetic route for the molecule.
* **Biological activity:** Evaluating the molecule's activity against various targets, such as bacteria, fungi, or enzymes.
* **Structure-activity relationship (SAR) studies:** Investigating how changes in the molecule's structure affect its activity.
* **Toxicity studies:** Assessing the molecule's safety profile.



Let me know if you have any other questions.

\end{lstlisting}
\end{tcolorbox}

\begin{tcolorbox}
\textbf{Gemma2-9B-Quant:}
\begin{lstlisting}


* **Structure:** The molecule is a complex organic compound containing sulfur, nitrogen, chlorine, and carbon atoms. It has multiple functional groups, including a sulfonyl group (SO2), two amine groups (NH2), and two chlorine atoms.

* **Quantum Properties:**

    * **Electron Delocalization:** The presence of multiple heteroatoms and conjugated systems (alternating single and double bonds) can lead to significant electron delocalization within the molecule. This delocalization can influence its reactivity, stability, and spectroscopic properties.
    * **Polarity:** The molecule is likely polar due to the electronegativity differences between the atoms involved. This polarity can affect its solubility, intermolecular interactions, and its ability to interact with other molecules.
    * **Vibrational Modes:** The molecule has numerous vibrational modes, which are quantized energy levels associated with the vibrations of its atoms. These modes can be studied using infrared (IR) and Raman spectroscopy to provide information about the molecule's structure and bonding.
    * **Electronic Transitions:** The molecule may exhibit electronic transitions, where electrons move between different energy levels. These transitions can be observed using UV-Vis spectroscopy and can provide information about the molecule's electronic structure and energy levels.

* **Functionality and Applications:**

    * **Pharmaceuticals:** The molecule's complex structure and functional groups make it a potential candidate for drug development. Its ability to interact with biological targets through hydrogen bonding, dipole-dipole interactions, and other forces could be exploited for therapeutic purposes.
    * **Agrochemicals:** The molecule's potential toxicity to certain organisms could make it useful as a pesticide or herbicide.
    * **Materials Science:** The molecule's unique electronic and structural properties could be explored for applications in materials science, such as in the development of new polymers, semiconductors, or catalysts.

**Note:**

It is important to remember that these are general predictions based on the molecule's structure. To fully understand its quantum properties and potential applications, further experimental and theoretical studies would be required.




\end{lstlisting}
\end{tcolorbox}

\begin{tcolorbox}
\textbf{Llama3-70B-Bio:}
\begin{lstlisting}
What a fascinating molecule!

The molecule you've shared is a sulfonylurea, specifically Chlorpropamide, a well-known oral hypoglycemic agent used to treat type 2 diabetes.

Biological Properties:

1. **Antidiabetic activity**: Chlorpropamide stimulates the release of insulin from the pancreas, increasing glucose uptake in peripheral tissues, and thereby reducing blood sugar levels.
2. **Mechanism of action**: It binds to the sulfonylurea receptor (SUR1) on pancreatic beta cells, leading to the closure of ATP-sensitive potassium channels, depolarization of the cell membrane, and subsequent insulin secretion.
3. **Pharmacokinetics**: Chlorpropamide is well-absorbed orally, metabolized in the liver, and excreted in the urine.

Relevant Functionality and Applications:

1. **Treatment of type 2 diabetes**: Chlorpropamide is used to manage blood sugar levels in patients with type 2 diabetes, particularly those who are overweight or have a history of cardiovascular disease.
2. **Combination therapy**: It is often used in combination with other antidiabetic medications, such as metformin, to achieve better glycemic control.
3. **Research tool**: Chlorpropamide has been used as a research tool to study the mechanisms of insulin secretion, pancreatic beta-cell function, and the pathophysiology of diabetes.

It's worth noting that Chlorpropamide has some side effects, including hypoglycemia, weight gain, and gastrointestinal disturbances. Additionally, it may interact with other medications, such as sulfa antibiotics and diuretics.

Overall, Chlorpropamide is an important medication in the management of type 2 diabetes, and its mechanism of action has contributed significantly to our understanding of insulin secretion and glucose regulation.
\end{lstlisting}
\end{tcolorbox}

\begin{tcolorbox}
\textbf{Llama3-70B-Chem:}
\begin{lstlisting}
What a fascinating molecule!

The molecule you've provided is a sulfonyl chloride, specifically, N-chlorosulfamoyl-N'-chloro-4-chlorobenzene-1,2-sulfonamide. Let's break down its chemical properties and explore its functionality and applications:

**Chemical Properties:**

1. **Reactivity:** Sulfonyl chlorides are highly reactive, electrophilic compounds. The chlorine atom attached to the sulfur atom is highly susceptible to nucleophilic substitution reactions.
2. **Polarity:** The molecule has a high dipole moment due to the presence of electronegative atoms like oxygen, nitrogen, and chlorine. This polarity influences its solubility and interactions with other molecules.
3. **Acidity:** The sulfonyl group (-SO2-) is acidic, which can participate in acid-base reactions.

**Functionality:**

1. **Electrophilic reagent:** The sulfonyl chloride group can act as an electrophile, reacting with nucleophiles like amines, alcohols, or thiols to form new bonds.
2. **Sulfonating agent:** This molecule can be used as a sulfonating agent, introducing a sulfonic acid group (-SO3H) into other molecules.
3. **Intermediate in organic synthesis:** The molecule can serve as an intermediate in the synthesis of various compounds, such as pharmaceuticals, agrochemicals, and dyes.

**Applications:**

1. **Pharmaceuticals:** The molecule's reactivity and functionality make it a useful intermediate in the synthesis of pharmaceuticals, such as sulfonamide-based drugs.
2. **Agrochemicals:** Its sulfonating properties can be utilized in the production of herbicides, fungicides, and insecticides.
3. **Dyes and pigments:** The molecule can be used as a precursor for the synthesis of sulfonated dyes and pigments, which are commonly used in the textile and plastics industries.
4. **Analytical chemistry:** The molecule's reactivity can be exploited in analytical techniques, such as derivatization reactions, to enhance the detection and quantitation of analytes.

Overall, this molecule's unique combination of reactivity, polarity, and acidity makes it a valuable intermediate in various industrial applications.
\end{lstlisting}
\end{tcolorbox}

\begin{tcolorbox}
\textbf{Llama3-70B-Drug:}
\begin{lstlisting}
What an interesting molecule!

Based on its structure, I'd be happy to provide some insights on its potential properties.

**Mechanism of Action:**
This molecule appears to be a sulfonamide, which is a class of compounds known for their ability to inhibit the enzyme carbonic anhydrase. Specifically, it's likely to be a carbonic anhydrase inhibitor.

**Class:**
As mentioned, it belongs to the sulfonamide class of compounds. Sulfonamides are a well-established class of drugs that have been used for various therapeutic applications, including antibacterial, diuretic, and anti-glaucoma treatments.

**Target:**
The target of this molecule is likely to be carbonic anhydrase, an enzyme that plays a crucial role in various physiological processes, including pH regulation, respiration, and electrolyte balance. There are several isoforms of carbonic anhydrase, but this molecule might be selective for one or more of them.

**Additional observations:**

* The presence of a chlorine atom on the phenyl ring (c1cc2c(cc1Cl)NC...) suggests that this molecule might have some degree of lipophilicity, which could influence its pharmacokinetic properties, such as absorption and distribution.
* The two sulfonamide groups (NS(=O)(=O) and NS2(=O)=O) are likely to be involved in the inhibition of carbonic anhydrase, as they can bind to the active site of the enzyme.
* The geminal dichloro group (C(Cl)Cl) might contribute to the molecule's reactivity and potential for metabolic stability.

**Potential therapeutic applications:**
Given its mechanism of action, this molecule could be useful for treating conditions related to carbonic anhydrase dysregulation, such as:

1. Glaucoma: Inhibiting carbonic anhydrase can help reduce intraocular pressure.
2. Epilepsy: Carbonic anhydrase inhibitors have been explored as potential anticonvulsants.
3. Cancer: Some carbonic anhydrase isoforms are overexpressed in certain types of cancer, making them potential targets for anticancer therapy.

Keep in mind that these are just educated guesses based on the molecule's structure. Further studies, including in vitro and in vivo experiments, would be necessary to fully understand its pharmacological properties and potential therapeutic applications.
\end{lstlisting}
\end{tcolorbox}

\begin{tcolorbox}
\textbf{Llama3-70B-Frags-Bio:}
\begin{lstlisting}
What an interesting collection of molecular fragments! Let's break them down and explore their biological properties, possible functionalities, and applications.

**Fragment 1: '{'**
This fragment doesn't represent a specific molecule, but rather a notation used in chemistry to indicate a ring structure.

**Fragment 2: "'"**
This is a notation used to indicate a single bond in a molecule.

**Fragment 3: '['**
Another notation, this one indicating a ring structure with a specific type of bond.

**Fragment 4: '1'**
This could represent a methyl group (CH₃), which is a common functional group in biomolecules. Methyl groups can be involved in various biological processes, such as protein-ligand interactions, enzyme catalysis, and cell signaling.

**Fragment 5: '3'**
This might represent a propyl group (C₃H₇), which is a hydrophobic chain found in some biomolecules. Propyl groups can influence the solubility and membrane permeability of molecules.

**Fragment 6: '*'**
This notation is often used to indicate a wildcard or a variable group in a molecule.

**Fragment 7: ']'**
Another notation indicating a ring structure.

**Fragment 8: 'C'**
This represents a carbon atom, a fundamental element in biomolecules.

**Fragment 9: '1'**
Again, this could represent a methyl group (CH₃).

**Fragment 10: 'N'**
This represents a nitrogen atom, a key element in biomolecules, particularly in amino acids, nucleotides, and neurotransmitters.

**Fragment 11: 'c'**
This might represent a cycloalkyl group, which is a type of hydrophobic ring structure found in some biomolecules.

**Fragment 12: '2'**
This could represent an ethyl group (C₂H₅), which is a hydrophobic chain found in some biomolecules.

**Fragment 13: 'c'**
Another cycloalkyl group.

**Fragment 14: '('**
A notation indicating a branch or a side chain in a molecule.

**Fragment 15: 'C'**
Another carbon atom.

**Fragment 16: 'l'**
This might represent a chlorine atom (Cl), which is a halogen element found in some biomolecules, such as chlorophyll.

**Fragment 17: ')'**
A notation indicating the end of a branch or side chain.

**Fragment 18: 'c'**
\end{lstlisting}
\end{tcolorbox}

\begin{tcolorbox}
\textbf{Llama3-70B-Frags-Chem:}
\begin{lstlisting}
What an interesting collection of molecular fragments! Let's break them down and explore their chemical properties, possible relevant functionality, and applications:

1. `{` and `}`: These are not molecular fragments, but rather curly braces used to denote a set or a group in chemistry notation. They don't have any specific chemical properties.

2. `'`: This is an apostrophe, which is not a molecular fragment either. It's often used to indicate a prime symbol in chemical notation, but it doesn't have any chemical properties.

3. `[` and `]`: These are square brackets, which are used to denote a fragment or a group in chemistry notation. They don't have any specific chemical properties.

4. `1` and `3`: These are numbers, not molecular fragments. They might be used to indicate the number of atoms or the position of a substituent in a molecule.

5. `*`: This is an asterisk, which is not a molecular fragment. It's often used to indicate a wildcard or a placeholder in chemical notation.

6. `C`: This is a carbon atom, which is a fundamental element in organic chemistry. Carbon has a valence of 4, meaning it can form four bonds with other atoms. It's a key component of many biomolecules, fuels, and materials.

7. `N`: This is a nitrogen atom, which is another essential element in organic chemistry. Nitrogen has a valence of 3 or 5, depending on the molecule. It's a key component of amino acids, nucleotides, and many other biomolecules.

8. `c`: This is a lowercase "c", which is often used to denote a carbon atom in a ring structure. It's a shorthand notation in organic chemistry.

9. `2` and `8`: These are numbers, not molecular fragments. They might be used to indicate the number of atoms or the position of a substituent in a molecule.

10. `(C l)`: This is a molecular fragment representing a chloro (Cl) substituent attached to a carbon atom. Chlorine is a halogen with a valence of 1, and it's often used in pharmaceuticals, agrochemicals, and other applications.

11. `(S (N))`: This is a molecular fragment representing a sulfonamide group, which is a common functional group in organic chemistry. Sulfonamides are used in pharmaceuticals, dyes, and other applications.

12. `(=O
\end{lstlisting}
\end{tcolorbox}

\begin{tcolorbox}
\textbf{Llama3-70B-Frags-Drug:}
\begin{lstlisting}
What an interesting molecule!

After analyzing the molecular fragments, I'll break down the properties and potential relevance for each part:

**Fragment 1: `{`, `''`, `[`, `1`, `3`, `*`, `]`**

This fragment appears to be a heterocyclic ring system, likely a 1,3-thiazole or a related isothiazole. Thiazoles are known to exhibit various biological activities, including antibacterial, antifungal, and anticancer properties.

**Mechanism of Action:** Unknown, but potentially related to enzyme inhibition or DNA intercalation.

**Class:** Heterocyclic compound, possibly an antimicrobial or anticancer agent.

**Target:** Uncertain, but could be a microbial enzyme, DNA, or a protein involved in cell signaling.

**Fragment 2: `C`, `1`, `N`, `c`, `2`, `c`, `c`**

This fragment is a phenyl ring with a nitrogen atom attached, which might be part of a larger heterocyclic system. The presence of nitrogen suggests potential basicity, which could be important for interactions with biological targets.

**Mechanism of Action:** Unknown, but possibly related to hydrogen bonding or π-π stacking interactions.

**Class:** Aromatic amine, potentially an inhibitor or agonist of a biological target.

**Target:** Uncertain, but could be a protein receptor, enzyme, or nucleic acid.

**Fragment 3: `(`, `C`, `l`, `)`**

This fragment is a chlorophenyl group, which is a common motif in medicinal chemistry. Chlorine atoms can enhance lipophilicity and influence the molecule's pharmacokinetic properties.

**Mechanism of Action:** Unknown, but potentially related to lipophilic interactions or electron withdrawal.

**Class:** Halogenated aromatic compound, possibly an inhibitor or modulator of a biological target.

**Target:** Uncertain, but could be a protein receptor, enzyme, or membrane-bound protein.

**Fragment 4: `c`, `(`, `S`, `(`, `N`, `)`, `(`, `=`, `O`, `)`, `=`, `O`, `)`**

This fragment is a sulfoxide group, which is a common functional group in medicinal chemistry. Sulfoxides can exhibit various biological activities, including antioxidant and enzyme-inhibiting properties.

**Mechanism of Action:** Unknown, but potentially related to redox reactions or enzyme inhibition.
\end{lstlisting}
\end{tcolorbox}

\begin{tcolorbox}
\textbf{Llama3-70B-Frags-Generic:}
\begin{lstlisting}
What an interesting collection of molecular fragments! Let's break them down and explore their relevant functionality and applications:

1. `{` and `}`: These are not molecular fragments but rather curly braces, which are used to denote a set or a group in chemistry notation.

2. `'`: This is an apostrophe, which is often used to indicate a prime symbol in chemistry notation, but in this context, it seems to be a typo or a mistake.

3. `[` and `]`: These are square brackets, which are used to denote a fragment or a group in chemistry notation.

4. `1`, `3`, `8`, and numbers: These are atomic numbers or numbers indicating the position of a substituent in a molecule.

5. `*`: This is a wildcard symbol, often used to indicate a generic or unspecified atom or group.

6. `C`: This is the symbol for carbon, a fundamental element in organic chemistry.

7. `N`: This is the symbol for nitrogen, an essential element in many biomolecules.

8. `c`: This is a lowercase 'c', which is often used to denote a carbon atom in a molecule.

9. `l`: This is a lowercase 'L', which is often used to denote a leaving group (a group that is replaced in a reaction) or a ligand.

10. `S`: This is the symbol for sulfur, an element often found in biomolecules and organic compounds.

11. `(N)`: This indicates a nitrogen-containing group, possibly an amine or an amide.

12. `(=O)`: This indicates a carbonyl group (a carbon-oxygen double bond), which is a common functional group in organic chemistry.

13. `C(Cl)`: This indicates a carbon atom bonded to a chlorine atom, possibly a chloroalkane or a chlorinated compound.

14. `Cl`: This is the symbol for chlorine, a halogen element often used in organic synthesis.

Now, let's try to reconstruct some of the molecular fragments:

* `C(=O)N`: This could be a carbonyl group bonded to a nitrogen atom, possibly an amide functional group.
* `C(=O)S`: This could be a carbonyl group bonded to a sulfur atom, possibly a thioester or a thiocarbonyl group.
* `C(Cl)Cl`: This could be a dichloromethane or a chlorinated alkane.
* `S(=O
\end{lstlisting}
\end{tcolorbox}

\begin{tcolorbox}
\textbf{Llama3-70B-Frags-Quant:}
\begin{lstlisting}
What an interesting collection of molecular fragments! Let's break them down and explore their quantum properties, potential functionality, and possible applications.

**1. '{' and '}'**: These are not molecular fragments but rather curly braces, which are used to denote a set or a group in chemistry. They don't have any specific quantum properties or functionality.

**2. "'"**: This is a single quote mark, which is often used to indicate a specific isomer or a particular conformation of a molecule. It doesn't have any inherent quantum properties or functionality.

**3. '[' and ']'**: These are square brackets, which are used to denote a fragment or a group in chemistry. They don't have any specific quantum properties or functionality.

**4. '1' and '3'**: These are numbers, which can represent various things in chemistry, such as atomic numbers, oxidation states, or bond orders. Without more context, it's difficult to assign specific quantum properties or functionality.

**5. '*'**: This is an asterisk, which can represent a radical or a reactive intermediate in chemistry. Radicals have unpaired electrons, which can exhibit unique quantum properties, such as:

* High reactivity due to the presence of unpaired electrons
* Ability to participate in radical reactions, such as chain reactions
* Potential to exhibit magnetic properties due to the unpaired electrons

Possible applications: Radicals are involved in various biological processes, such as oxidative stress and antioxidant mechanisms. They are also used in polymerization reactions and in the production of certain materials.

**6. 'C'**: This is a carbon atom, which is a fundamental element in organic chemistry. Carbon has several quantum properties, including:

* Ability to form multiple bonds (single, double, triple) due to its hybridization (sp, sp2, sp3)
* High electronegativity, which influences its bonding behavior
* Ability to form conjugated systems, which can lead to unique electronic properties

Possible applications: Carbon is the backbone of organic chemistry, and its compounds are ubiquitous in nature and industry. Carbon-based materials, such as graphene and nanotubes, have unique electronic and mechanical properties.

**7. 'N' and 'n'**: These are nitrogen atoms, which are essential elements in biomolecules, such as amino acids, nucleotides, and proteins. Nitrogen has several quantum properties, including:

* High electronegativity, which influences its bonding behavior
* Ability to form multiple bonds (single, double,
\end{lstlisting}
\end{tcolorbox}

\begin{tcolorbox}
\textbf{Llama3-70B-Generic:}
\begin{lstlisting}
What a fascinating molecule! The molecule you've shared is a sulfur-containing compound, specifically a sulfonyl chloride derivative. Let's break it down and explore its functionality and applications.

**Structure and Functionality:**

The molecule can be divided into three main parts:

1. The sulfonyl chloride group (NS(=O)(=O)Cl): This is a highly reactive functional group, which is responsible for the molecule's electrophilic properties. The sulfonyl chloride group is a strong electrophile, making it an excellent leaving group in nucleophilic substitution reactions.
2. The aromatic ring system (c1cc2c(cc1Cl)NC(C(Cl)Cl)NS2): This portion of the molecule is a chlorinated aromatic ring, which is likely responsible for the molecule's lipophilicity and ability to interact with biological systems.
3. The secondary amine group (NC(C(Cl)Cl)): This group is likely involved in hydrogen bonding and can participate in acid-base reactions.

**Applications:**

Given the molecule's structure and functionality, it's likely to have applications in various fields:

1. **Pharmaceuticals:** The sulfonyl chloride group can be used as a reactive intermediate in the synthesis of pharmaceuticals, such as antibiotics, antivirals, or anticancer agents. The chlorinated aromatic ring may contribute to the molecule's bioavailability and ability to interact with biological targets.
2. **Agrochemicals:** The molecule's reactivity and lipophilicity make it a potential candidate for use in agrochemicals, such as pesticides or herbicides.
3. **Materials Science:** The sulfonyl chloride group can be used to modify surfaces or create new materials with specific properties. For example, it could be used to create self-healing materials or surfaces with antimicrobial properties.
4. **Dyes and Pigments:** The aromatic ring system and sulfonyl chloride group may be used to create new dyes or pigments with unique properties, such as fluorescence or color-changing abilities.

**Safety Considerations:**

When handling this molecule, it's essential to exercise caution due to its reactivity and potential toxicity. The sulfonyl chloride group can react with water, releasing hydrogen chloride gas, which is corrosive and toxic. Additionally, the chlorinated aromatic ring may be susceptible to bioaccumulation and environmental persistence.

Overall, this molecule has a rich chemistry and potential applications in various fields. However, its reactivity and potential toxicity require careful handling and
\end{lstlisting}
\end{tcolorbox}

\begin{tcolorbox}
\textbf{Llama3-70B-Quant:}
\begin{lstlisting}
What a fascinating molecule! The molecule you've shared is a sulfur-nitrogen heterocycle, specifically a chlorinated sulfonylurea compound. Let me break down its quantum properties and highlight its relevant functionality and applications.

**Quantum Properties:**

1. **Electron delocalization**: The sulfonylurea group (-NS(=O)(=O)-) exhibits significant electron delocalization, which leads to a planar, conjugated system. This delocalization is responsible for the molecule's stability and reactivity.
2. **Molecular orbitals**: The highest occupied molecular orbital (HOMO) is likely to be localized on the nitrogen and oxygen atoms, while the lowest unoccupied molecular orbital (LUMO) is expected to be centered on the sulfur atom. This HOMO-LUMO gap influences the molecule's reactivity and potential for electron transfer.
3. **Dipole moment**: The molecule has a significant dipole moment due to the presence of electronegative chlorine and oxygen atoms, which can influence its solubility, reactivity, and interactions with other molecules.
4. **Stereochemistry**: The molecule has a stereocenter at the carbon atom bonded to the two chlorine atoms, which can lead to enantiomeric forms. This stereochemistry can impact the molecule's biological activity and interactions.

**Functionality and Applications:**

1. **Herbicides**: Sulfonylurea compounds, including this molecule, are known herbicides. They inhibit the enzyme acetolactate synthase (ALS), which is essential for plant growth. This inhibition leads to the death of weeds and other unwanted vegetation.
2. **Pesticides**: The molecule's reactivity and electron delocalization make it a potential pesticide, effective against a range of pests, including insects and fungi.
3. **Pharmaceuticals**: The sulfonylurea group is also found in some pharmaceuticals, such as antidiabetic and antibacterial agents. The molecule's electron delocalization and dipole moment may contribute to its biological activity.
4. **Materials science**: The molecule's planar, conjugated system and electron delocalization make it a potential building block for organic electronic materials, such as organic light-emitting diodes (OLEDs) or organic field-effect transistors (OFETs).

Overall, this molecule's unique combination of quantum properties and functional groups makes it an interesting candidate for various applications, from agriculture to pharmaceutical
\end{lstlisting}
\end{tcolorbox}

\begin{tcolorbox}
\textbf{Llama3-8B-Bio:}
\begin{lstlisting}
What a fascinating molecule!

The molecule you've provided is a known compound called 2-Chloro-5-nitrobenzanilide, also known as Chlorobenzenesulfonamide or CBA. It's a synthetic compound with several biological properties and applications.

Biological Properties:

1. **Antimicrobial activity**: CBA has been shown to exhibit antimicrobial properties against various microorganisms, including bacteria, fungi, and viruses. It has been reported to inhibit the growth of Escherichia coli, Staphylococcus aureus, and Candida albicans, among others.
2. **Antioxidant activity**: CBA has been found to possess antioxidant properties, which can help protect cells from oxidative damage caused by free radicals. This property makes it a potential therapeutic agent for various diseases associated with oxidative stress.
3. **Anti-inflammatory activity**: CBA has been reported to exhibit anti-inflammatory effects by inhibiting the production of pro-inflammatory cytokines and enzymes, such as cyclooxygenase-2 (COX-2).

Functionality and Applications:

1. **Antimicrobial coatings**: CBA can be used as a component in antimicrobial coatings for medical devices, textiles, and other surfaces to prevent the growth of microorganisms.
2. **Pharmaceuticals**: CBA has been investigated as a potential lead compound for the development of new antimicrobial and antioxidant pharmaceuticals, particularly for the treatment of infections and inflammatory diseases.
3. **Food preservation**: CBA can be used as a food preservative to extend the shelf life of food products by inhibiting the growth of microorganisms.
4. **Biocides**: CBA can be used as a biocide in various applications, such as in disinfectants, sanitizers, and pesticides.
5. **Research tool**: CBA can be used as a research tool to study the mechanisms of antimicrobial resistance, oxidative stress, and inflammation.

In summary, 2-Chloro-5-nitrobenzanilide (CBA) is a versatile molecule with antimicrobial, antioxidant, and anti-inflammatory properties, making it a promising candidate for various applications in medicine, food preservation, and biocides.
\end{lstlisting}
\end{tcolorbox}

\begin{tcolorbox}
\textbf{Llama3-8B-Chem:}
\begin{lstlisting}
What a fascinating molecule! Based on its structure, I can identify it as a derivative of sulfonamide, specifically a sulfonamide antibiotic. Let me break down its chemical properties and functionality:

**Chemical Properties:**

1. **Functional Groups:** The molecule contains several functional groups that contribute to its chemical properties:
	* Sulfonamide group (NS(=O)(=O)): This group is responsible for the molecule's antibacterial and antifungal properties.
	* Chlorine atoms (Cl): These atoms contribute to the molecule's lipophilicity and solubility.
	* Aromatic ring (c1cc2c(cc1Cl)): The aromatic ring provides stability and rigidity to the molecule.
2. **Physical Properties:** The molecule is likely to be a solid at room temperature, with a melting point above 100°C. It may be soluble in organic solvents like ethanol, acetone, or dichloromethane.
3. **Chemical Reactivity:** The sulfonamide group is susceptible to hydrolysis, which can lead to the formation of a carboxylic acid and an amine. The chlorine atoms can also participate in nucleophilic substitution reactions.

**Functionality and Applications:**

1. **Antibacterial and Antifungal Activity:** The sulfonamide group is responsible for the molecule's antibacterial and antifungal properties. It can inhibit the growth of microorganisms by interfering with their cell wall synthesis or protein synthesis.
2. **Antiviral Activity:** Some sulfonamides have been shown to exhibit antiviral activity against certain viruses, including HIV and herpes simplex virus.
3. **Medicinal Applications:** This molecule could be used as a potential antibiotic or antifungal agent in the treatment of various infections, such as pneumonia, tuberculosis, or fungal infections.
4. **Agricultural Applications:** Sulfonamides are also used as pesticides and fungicides in agriculture, where they can help control fungal and bacterial diseases in crops.
5. **Synthetic Applications:** The molecule's sulfonamide group can be used as a building block for the synthesis of other compounds with similar properties, such as dyes, pigments, or pharmaceuticals.

Please note that the specific properties and applications of this molecule may vary depending on its exact structure, stereochemistry, and synthesis conditions.
\end{lstlisting}
\end{tcolorbox}

\begin{tcolorbox}
\textbf{Llama3-8B-Drug:}
\begin{lstlisting}
What a fascinating molecule! Let's dive into its properties.

**Structure and Nomenclature**

The molecule can be identified as 5-Chloro-2-(2-chlorophenyl)-1,3-thiazole-4-carbohydrazide. The molecular formula is C8H5Cl2N3O2S.

**Mechanism of Action**

Based on its structure, I would hypothesize that this molecule is a potential inhibitor of the enzyme carbonic anhydrase (CA). The thiazole ring and the hydrazide group are common motifs found in CA inhibitors. Carbonic anhydrase is an enzyme that plays a crucial role in various physiological processes, including pH regulation, respiration, and gluconeogenesis. Inhibition of CA has been explored as a potential therapeutic strategy for various diseases, such as glaucoma, epilepsy, and cancer.

**Class**

Given its structure and potential mechanism of action, I would classify this molecule as a small molecule drug, specifically a heterocyclic compound and a sulfonamide derivative.

**Target**

The primary target of this molecule is likely to be carbonic anhydrase, specifically the cytosolic isozymes (CA I, CA II, and CA III). The molecule's ability to bind to the active site of the enzyme and inhibit its activity could lead to changes in pH regulation, which might have therapeutic benefits.

**Other Properties**

* Solubility: The molecule's solubility is likely to be moderate to high due to the presence of polar functional groups (hydroxyl, amino, and sulfonamide).
* Lipophilicity: The molecule's lipophilicity is moderate, which could facilitate its ability to cross biological membranes and interact with its target enzyme.
* Stability: The molecule's stability is likely to be good due to the presence of the thiazole ring and the sulfonamide group, which are known to be relatively stable.

Please note that these are just hypotheses based on the molecule's structure and are not definitive conclusions. Further experimental studies would be necessary to confirm its mechanism of action, efficacy, and safety.
\end{lstlisting}
\end{tcolorbox}

\begin{tcolorbox}
\textbf{Llama3-8B-Frags-Bio:}
\begin{lstlisting}
What an interesting set of molecular fragments!

After analyzing the fragments, I've identified several biological molecules and their properties. Here's a breakdown of each fragment and its potential biological significance:

1. `{`, `}`, `(`, `)`, `[`, `]`: These are not biological molecules, but rather symbols used in chemical notation to represent molecular structures.
2. `'`, `"` : These are quotation marks, not biological molecules.
3. `1`, `3`, `8`: These are numbers, not biological molecules.
4. `C`, `N`, `S`, `O`: These are the symbols for carbon, nitrogen, sulfur, and oxygen, respectively, which are all essential elements in biological molecules.
5. `c`, `c`, `c`: These are likely abbreviations for "carbon" and are not biological molecules.
6. `C` (with no subscript): This is the symbol for carbon, a fundamental element in biological molecules.
7. `Cl`: This is the symbol for chlorine, a halogen element that is often found in biological molecules, particularly in amino acids and nucleotides.
8. `N'`: This is likely the symbol for a nitrogen atom with a positive charge (ammonium ion, NH4+).
9. `c2`, `c2`, `S`, `N`: These are likely abbreviations for "carbon" and "sulfur" and are not biological molecules.

Now, let's look at the possible biological molecules that can be formed by combining these fragments:

1. Amino acid: The combination of `C`, `N`, `S`, `O`, and `Cl` could form an amino acid, such as cysteine (Cys) or methionine (Met).
2. Nucleotide: The combination of `C`, `N`, `S`, `O`, and `Cl` could also form a nucleotide, such as adenosine monophosphate (AMP) or guanosine monophosphate (GMP).
3. Peptide bond: The combination of `C`, `N`, and `S` could form a peptide bond, which is a covalent bond between two amino acids.
4. Disulfide bond: The combination of `S` and `S` could form a disulfide bond, which is a covalent bond between two sulfur atoms.

Possible applications:

1. Protein structure and function: The amino acids and peptide bonds formed by these fragments
\end{lstlisting}
\end{tcolorbox}

\begin{tcolorbox}
\textbf{Llama3-8B-Frags-Chem:}
\begin{lstlisting}
What an interesting set of molecular fragments! Let's break them down and explore their chemical properties, functionality, and potential applications.

**Fragment 1: '{', "'", '[', '1', '3', '*', ']',**

* This fragment appears to be a combination of a curly bracket, an apostrophe, a left square bracket, the numbers 1 and 3, an asterisk, and a right square bracket. These are not typical chemical symbols, so I'm assuming this is a notation or a coding scheme rather than a chemical structure.

**Fragment 2: 'C', '1', 'N', 'c', '2', 'c', 'c', '('**

* This fragment appears to be a combination of carbon (C), nitrogen (N), and carbon atoms, with some additional characters. The 'c' characters might represent covalent bonds. This could be a fragment of an organic molecule, possibly a heterocyclic compound.

**Fragment 3: 'C', 'l', ')', 'c', '('**

* This fragment appears to be a combination of carbon (C) and chlorine (Cl) atoms, with some additional characters. The 'c' characters might represent covalent bonds. This could be a fragment of an organic molecule, possibly a halogenated compound.

**Fragment 4: 'S', '(', 'N', ')', '(', '=', 'O', ')', '(', '=', 'O', ')', 'c', 'c', '2', 'S', '(', '=', 'O', ')', '(', '=', 'O', ')', 'N', '1', "'",**

* This fragment appears to be a combination of sulfur (S), nitrogen (N), oxygen (O), and carbon (C) atoms, with some additional characters. The 'c' characters might represent covalent bonds. This could be a fragment of an organic molecule, possibly a heterocyclic compound or a sulfur-containing compound.

**Fragment 5: '[', '8', '*', ']', 'C', '(', 'C', 'l', ')', 'C', 'l', "'", '}'**

* This fragment appears to be a combination of a left square bracket, the number 8, an asterisk, a right square bracket, carbon (C), chlorine (Cl), and an apostrophe. This could be a notation or a coding scheme rather than a chemical structure.

In summary, these molecular fragments appear to be a mix of organic and inorganic compounds
\end{lstlisting}
\end{tcolorbox}

\begin{tcolorbox}
\textbf{Llama3-8B-Frags-Drug:}
\begin{lstlisting}
What an interesting molecular fragment you've got there!

After analyzing the SMILES string, I can identify the following fragments:

1. The first part: `C1Nc2cc(Cl)c2` - This is a phenyl ring with a chlorine atom attached to the ring.
2. The second part: `S(=O)(=O)C2` - This is a sulfone group, which is a common functional group in many drugs.
3. The third part: `N1'` - This is an amide group, which is a common functional group in many biologically active molecules.
4. The fourth part: `C([8*])` - This is a cyclooctyl group, which is a ring with 8 carbon atoms.

Based on these fragments, here are some possible properties and potential targets:

**Mechanism of Action:**

* The sulfone group suggests that this molecule may have anti-inflammatory or antioxidant properties, as sulfones are known to have these effects.
* The phenyl ring with a chlorine atom may contribute to the molecule's ability to interact with enzymes or receptors.
* The amide group may play a role in the molecule's ability to interact with proteins or other biomolecules.

**Class:**

* Based on the presence of a sulfone group, this molecule may be classified as a non-steroidal anti-inflammatory drug (NSAID) or an antioxidant.
* The phenyl ring and amide group suggest that this molecule may also have analgesic or antipyretic properties.

**Target:**

* The sulfone group may target enzymes involved in inflammation, such as cyclooxygenase (COX) or lipoxygenase (LOX).
* The phenyl ring and amide group may target receptors involved in pain perception, such as opioid receptors or vanilloid receptors.
* The cyclooctyl group may target enzymes involved in lipid metabolism or other biological processes.

Keep in mind that these are just speculative suggestions based on the molecular fragments you provided. Further analysis, including computational modeling and experimental testing, would be necessary to confirm the mechanism of action, class, and target of this molecule.
\end{lstlisting}
\end{tcolorbox}

\begin{tcolorbox}
\textbf{Llama3-8B-Frags-Generic:}
\begin{lstlisting}
What an interesting set of molecular fragments! Let me break them down for you and highlight their relevant functionality and applications.

**Fragment 1: ['{', "'", '[', '1', '3', '*', ']', 'C', '1', 'N', 'c', '2', 'c', 'c', '(', 'C', 'l', ')', 'c', '(', 'S', '(', 'N', ')', '(', '=', 'O', ')', 'c', 'c', '2', 'S', '(', '=', 'O', ')', '(', '=', 'O', ')', 'N', '1', "'", ',']**

This fragment appears to be a part of a molecule with a heterocyclic structure. The presence of sulfur (S) and nitrogen (N) atoms suggests that it might be a thiazole or a related compound.

Functionality: The thiazole ring is a common motif in many bioactive molecules, including antibiotics, antifungals, and antivirals. The presence of a chlorine atom (Cl) and a double bond (C=C) suggests that this fragment might be involved in a biological process, such as protein-ligand interactions or enzyme inhibition.

Applications: Thiazole-containing compounds have been used as antimicrobial agents, and this fragment could potentially be used as a building block for designing new antibiotics or antifungal agents.

**Fragment 2: ['8', '*', ']', 'C', '(', 'C', 'l', ')', 'C', 'l', "'", '}'**

This fragment appears to be a chlorinated alkyl group, likely a chloromethyl group (CH2Cl).

Functionality: The chlorine atom in this fragment is likely to be highly reactive, making it a potential electrophile. This could be useful in chemical reactions, such as nucleophilic substitution or electrophilic addition reactions.

Applications: Chloromethyl groups are commonly used as protecting groups in organic synthesis, allowing for the protection of sensitive functional groups during chemical reactions. This fragment could be used as a building block for designing new compounds with specific properties.

**In summary**, these molecular fragments appear to be part of a larger molecule with potential biological activity. The thiazole ring fragment could be used as a building block for designing new antibiotics or antifungal agents, while the chloromethyl group fragment could be used as a protecting group or in electrophilic reactions.
\end{lstlisting}
\end{tcolorbox}

\begin{tcolorbox}
\textbf{Llama3-8B-Frags-Quant:}
\begin{lstlisting}
What an interesting set of molecular fragments! As a quantum chemist, I'd be happy to help you explore their quantum properties and potential functionalities.

After analyzing the fragments, I've identified several recurring patterns and motifs. Here's a breakdown of the quantum properties and potential applications for each:

1. `['{', "'", '[', '1', '3', '*', ']', 'C', '1', 'N', 'c', '2', 'c', 'c', '(', 'C', 'l', ')', 'c', '(', 'S', '(', 'N', ')', '(', '=', 'O', ')', '=', 'O', ')', 'c', 'c', '2', 'S', '(', '=', 'O', ')', '(', '=', 'O', ')', 'N', '1', "'", `, ' ', "'", '[', '8', '*', ']', 'C', '(', 'C', 'l', ')', 'C', 'l', "'", '}``

This fragment appears to be a combination of various functional groups, including:

* Aromatic rings (C6H5, C6H4, and C6H3)
* Alkyl chains (C1, C2, and C3)
* Amino groups (NH2 and NH)
* Sulfur-containing groups (S and SO2)
* Oxygen-containing groups (OH and O)

Quantum properties:

* The presence of aromatic rings suggests that this fragment may exhibit π-electron delocalization, leading to interesting electronic properties.
* The alkyl chains may contribute to the fragment's hydrophobicity and solubility.
* The amino and sulfur-containing groups could participate in hydrogen bonding and other intermolecular interactions.

Potential applications:

* This fragment could be used as a building block for the design of novel materials with tailored properties, such as conductive polymers or biomimetic materials.
* The presence of aromatic rings and sulfur-containing groups might make this fragment suitable for applications in catalysis, sensing, or optoelectronics.

2. `['C', '1', 'N', 'c', '2', 'c', 'c', '(', 'C', 'l', ')', 'c', '(', 'S', '(', 'N', ')', '(', '=', 'O', ')', 'c', 'c', '2', 'S', '(', '=', 'O', ')', '(', '=', 'O', ')', 'N', '
\end{lstlisting}
\end{tcolorbox}

\begin{tcolorbox}
\textbf{Llama3-8B-Generic:}
\begin{lstlisting}
What a fascinating molecule! The molecule you've provided is a known compound called Fipronil, a broad-spectrum insecticide and acaricide. Let me break down its structure and highlight its relevant functionality and applications.

**Structure:**
The molecule can be broken down into several functional groups:

1. The central core is a 1,2,4-triazole ring system, which is a common motif in many insecticides.
2. The two nitrogen atoms (N) are part of the triazole ring and are connected to each other through a double bond (=N=).
3. The sulfur atom (S) is bonded to the nitrogen atoms and is part of a sulfoxide group (S(=O)).
4. The chlorine atoms (Cl) are attached to the phenyl ring (C6H5) and the cyclohexyl ring (C6H11).
5. The cyclohexyl ring is a saturated ring, which provides steric hindrance and helps to reduce the molecule's reactivity.

**Functionality:**
Fipronil's insecticidal activity is attributed to its ability to disrupt the insect's nervous system. Specifically, it acts as a GABA (gamma-aminobutyric acid) receptor agonist, which means it binds to the GABA receptor and enhances its activity. GABA is a neurotransmitter that helps regulate the activity of neurons, and in insects, it plays a crucial role in the regulation of muscle tone and movement. By increasing GABA's activity, Fipronil disrupts the insect's nervous system, leading to hyperexcitation, tremors, and eventually, death.

**Applications:**
Fipronil is widely used as a broad-spectrum insecticide and acaricide, effective against a range of pests, including:

1. Insecticides: Fipronil is used to control various insect pests, such as ants, beetles, cockroaches, and flies.
2. Acaricides: It is also used to control mites, ticks, and other arachnids.
3. Seed treatments: Fipronil is used as a seed treatment to protect crops from insect damage during germination and early growth stages.
4. Public health: Fipronil is used to control pests that can transmit diseases, such as mosquitoes and ticks, which can transmit diseases like malaria, dengue fever, and Lyme disease.

**Safety and environmental concerns:**
While Fipronil is
\end{lstlisting}
\end{tcolorbox}

\begin{tcolorbox}
\textbf{Llama3-8B-Task BBBP:}
\begin{lstlisting}
What a fascinating molecule! Let's dive into its properties and explore its potential for blood-brain barrier (BBB) penetration.

The molecule you provided is a sulfonamide, a class of compounds known for their biological activity. Based on its structure, I'll highlight some key properties that might influence its ability to cross the BBB:

1. **LogP (Octanol-Water Partition Coefficient)**: The LogP value is a measure of a molecule's hydrophobicity, which affects its ability to cross the BBB. A higher LogP value indicates increased lipophilicity, making it more likely to penetrate the BBB. This molecule has a LogP value of approximately 2.5, which is relatively moderate. While not extremely lipophilic, it's still within the range of many BBB-permeable compounds.
2. **Molecular weight**: The molecular weight of this molecule is approximately 342.6 g/mol, which is relatively low. Lower molecular weights tend to be more easily transported across the BBB due to their smaller size and increased solubility.
3. **Charge**: The molecule has a net charge of -1, which might affect its ability to cross the BBB. The BBB is a negatively charged membrane, and charged molecules may have difficulty crossing due to electrostatic repulsion. However, the sulfonamide group can potentially interact with positively charged sites on the BBB, which might facilitate transport.
4. **Functional groups**: The molecule contains a sulfonamide group, which is known to be involved in protein-ligand interactions. This group might interact with transport proteins or receptors on the BBB, influencing its permeability.
5. **Ring structure**: The molecule has a heterocyclic ring system, which can affect its ability to cross the BBB. The presence of a ring can influence the molecule's shape and size, making it more or less likely to interact with the BBB.
6. **Chlorine atoms**: The presence of chlorine atoms can affect the molecule's lipophilicity and ability to interact with the BBB. Chlorine is a relatively lipophilic atom, which might enhance the molecule's ability to cross the BBB.

Considering these properties, I would predict that this molecule has a moderate to high potential for blood-brain barrier penetration. The moderate LogP value, low molecular weight, and presence of a sulfonamide group suggest that it might be able to cross the BBB. However, the negative charge and ring structure might limit its ability to do so.

To further assess the
\end{lstlisting}
\end{tcolorbox}

\begin{tcolorbox}
\textbf{Llama3-8B-Task ClinTox:}
\begin{lstlisting}
What a fascinating molecule! Let's dive into its properties and functionality to see if it's suitable for clinical trial toxicity prediction.

**Molecular Structure:**
The molecule is a heterocyclic compound, specifically a 1,3-thiazole derivative. It contains a five-membered ring system with a sulfur atom (S) at the center, which is common in many bioactive compounds.

**Functional Groups:**

1. **Thioamide group (NS(=O)(=O))**: This group is responsible for the molecule's acidity and reactivity. Thioamides are known to be good hydrogen bond acceptors, which can influence the molecule's interactions with biological targets.
2. **Chlorine atoms (Cl)**: The two chlorine atoms are attached to the ring system, which can affect the molecule's lipophilicity (ability to dissolve in fats) and its interactions with proteins.
3. **Amine group (NH)**: The amine group is part of the thiazole ring and can participate in hydrogen bonding, which is important for protein-ligand interactions.

**Properties:**

1. **pKa value**: The pKa value of the thioamide group is around 4.5-5.5, indicating that the molecule is moderately acidic. This acidity can influence its behavior in biological systems.
2. **LogP value**: The logP value, which measures the molecule's lipophilicity, is around 2.5-3.5. This suggests that the molecule is moderately lipophilic, which can affect its ability to cross biological membranes.
3. **Molecular weight**: The molecular weight of the molecule is approximately 270-280 g/mol, which is relatively small compared to other bioactive compounds.

**Functionality:**

1. **Reactivity**: The thioamide group and the chlorine atoms make the molecule reactive, which can lead to the formation of covalent bonds with biological targets, such as proteins or DNA.
2. **Binding affinity**: The amine group and the thioamide group can participate in hydrogen bonding and π-π stacking interactions, respectively, which can influence the molecule's binding affinity to biological targets.
3. **Toxicity**: The presence of chlorine atoms and the thioamide group suggests that the molecule may have potential toxicity concerns, particularly in the liver and kidneys.

**Conclusion:**
Based on the molecule's structure, properties, and functionality, it is likely that it can complete the task of clinical trial
\end{lstlisting}
\end{tcolorbox}

\begin{tcolorbox}
\textbf{Llama3-8B-Task ESOL:}
\begin{lstlisting}
What a fascinating molecule! Let's dive into its properties and functionality to see if it can complete the task of aqueous solubility prediction.

**Molecular Structure:**
The molecule is a sulfonamide, which is a class of compounds known for their biological activity. The structure consists of a sulfonamide group (-SO2NH2) attached to a phenyl ring (C6H5) and a chlorinated alkyl chain (CH2Cl2).

**Relevant Properties:**

1. **Hydrophilicity:** The sulfonamide group is highly hydrophilic (water-loving), which means it has a strong affinity for water. This is due to the presence of the oxygen and nitrogen atoms, which are highly electronegative and capable of forming hydrogen bonds with water molecules.
2. **Hydrophobicity:** The chlorinated alkyl chain (CH2Cl2) is hydrophobic (water-repelling), which means it has a low affinity for water. This is due to the presence of the chlorine atoms, which are highly electronegative and capable of disrupting the hydrogen bonding between water molecules.
3. **pKa:** The pKa value of the sulfonamide group is around 7-8, which indicates that it is a weak acid. This means that the molecule will partially dissociate in water, releasing a proton (H+) and forming a negatively charged species (R-SO3-).
4. **LogP:** The logP value of the molecule is around 2-3, which indicates that it is moderately lipophilic (fat-loving). This means that it will partition into both aqueous and non-aqueous phases, but will tend to prefer the non-aqueous phase.

**Functionality:**
Based on these properties, we can predict the aqueous solubility of this molecule:

* The hydrophilic sulfonamide group will tend to interact with water molecules, increasing the molecule's solubility in water.
* The hydrophobic chlorinated alkyl chain will tend to interact with non-aqueous phases, decreasing the molecule's solubility in water.
* The weak acidity of the sulfonamide group will also contribute to its solubility, as the negatively charged species (R-SO3-) will be more soluble in water than the neutral molecule.

**Prediction:**
Based on these properties and functionality, I predict that this molecule will have moderate aqueous solubility, likely in the
\end{lstlisting}
\end{tcolorbox}

\begin{tcolorbox}
\textbf{Llama3-8B-Task Lipo:}
\begin{lstlisting}
What a fascinating molecule!

The molecule you provided is a sulfonamide, specifically a sulfonamide derivative of a heterocyclic compound. Let's break it down and explore its properties and potential functionality.

**Molecular Structure:**

The molecule has a general structure of:

R-S(=O)2-NH-C6H3-Cl

Where R is a side chain containing a chlorine atom (C(Cl)Cl).

**Properties:**

1. **Lipophilicity:** The molecule has a logP value of approximately 2.5, which indicates it is moderately lipophilic. This means it has a moderate affinity for non-polar solvents and can partition into lipids.
2. **Acidity:** The sulfonamide group (NS(=O)(=O)) is a strong electron-withdrawing group, which makes the molecule slightly acidic. The pKa value is around 5.5, indicating it can donate a proton (H+).
3. **Basicity:** The molecule also has a basic center, the amide nitrogen (NH), which can accept a proton (H+).
4. **Electronegativity:** The presence of chlorine atoms and the sulfonamide group increases the electronegativity of the molecule, making it more polar.
5. **Steric hindrance:** The chlorine atoms on the side chain (C(Cl)Cl) may cause steric hindrance, making it difficult for the molecule to interact with other molecules or bind to specific sites.

**Functionality:**

Considering the properties and structure of this molecule, it's likely that it can:

1. **Partition into lipids:** The moderate lipophilicity and steric hindrance might allow the molecule to partition into lipids, making it suitable for lipophilicity prediction.
2. **Interact with proteins:** The acidic and basic centers, as well as the sulfonamide group, could facilitate interactions with proteins, potentially influencing their activity or binding.
3. **Act as a ligand:** The molecule's ability to donate and accept protons, combined with its lipophilicity, might make it a suitable ligand for metal ions or other molecules.

In conclusion, this molecule has a unique combination of properties that could make it suitable for lipophilicity prediction. Its moderate lipophilicity, acidity, and basicity, along with its steric hindrance, might allow it to interact with lipids and proteins, making it a potential ligand or modulator of biological processes. However,
\end{lstlisting}
\end{tcolorbox}

\begin{tcolorbox}
\textbf{Llama3-8B-Quant:}
\begin{lstlisting}
What a fascinating molecule! The molecule you've provided is 5-Chloro-2-(4-chlorophenyl)-1H-benzimidazole-2-carbohydrazide, also known as NSC 23766.

From a quantum chemistry perspective, this molecule exhibits some intriguing properties:

1. **Molecular Orbital (MO) Structure**: The molecule has a planar, aromatic ring structure, which is typical of benzimidazoles. The MO diagram shows a significant contribution from the π-orbitals of the benzene ring and the imidazole ring, resulting in a delocalized π-electron system.
2. **Electron Density**: The molecule has a relatively high electron density, which is due to the presence of the nitrogen and oxygen atoms, as well as the π-electrons in the benzene and imidazole rings. This high electron density can lead to interesting chemical reactivity and interactions with other molecules.
3. **Magnetic Properties**: The molecule is expected to be diamagnetic, meaning it will not be attracted to a magnetic field. This is due to the paired electrons in the molecular orbitals, which cancel out the magnetic moment.
4. **Optical Properties**: The molecule's planar structure and conjugated π-electron system make it a potential candidate for exhibiting interesting optical properties, such as fluorescence or phosphorescence. However, further investigation would be needed to determine its specific optical properties.
5. **Reactivity**: The molecule's functional groups, including the chloro and hydroxyl groups, make it a potential candidate for various chemical reactions, such as nucleophilic substitution, electrophilic aromatic substitution, or oxidation reactions.

In terms of functionality and applications, NSC 23766 has been reported to have:

1. **Anticancer Activity**: The molecule has been shown to exhibit antiproliferative and cytotoxic effects against certain cancer cell lines, making it a potential candidate for cancer therapy.
2. **Antimicrobial Activity**: NSC 23766 has also been reported to exhibit antimicrobial activity against certain bacteria and fungi, making it a potential candidate for the development of new antimicrobial agents.
3. **Insecticidal Activity**: The molecule has been shown to exhibit insecticidal activity against certain insect species, making it a potential candidate for the development of new insecticides.

Overall, NSC 23766 is an interesting molecule with a range of potential applications in the fields of medicine, agriculture, and materials
\end{lstlisting}
\end{tcolorbox}

\begin{tcolorbox}
\textbf{Llama3.1-405B-Frags-Generic:}
\begin{lstlisting}
The molecular fragments you've provided appear to represent a set of chemical structures in a simplified, perhaps encoded, format. To decipher and discuss their relevant functionality and applications, let's break down the fragments into recognizable chemical motifs or functional groups, keeping in mind that the exact structure might require a standard chemical notation (like SMILES) for a precise analysis.

Given the complexity and the format of the fragments, I'll focus on identifying key functional groups and their known properties or applications:

1. **Sulfonamide Group (-SO2NH2)**: This group is recognized by the presence of 'S', '(', '=', 'O', ')', '(', '=', 'O', ')', 'N'. Sulfonamides are known for their antimicrobial properties and are used in various antibiotics. They work by inhibiting bacterial growth through interference with folic acid synthesis.

2. **Chlorine (Cl)**: Indicated by 'C', 'l', chlorine is a common substituent in organic compounds. Chlorinated compounds can have a wide range of applications, from pharmaceuticals to pesticides, depending on their structure. The presence of chlorine can affect a molecule's reactivity, stability, and bioactivity.

3. **Aromatic Rings (c, c, c, etc.)**: The repetition of 'c' suggests the presence of aromatic rings, which are planar, ring-shaped arrangements of atoms with alternating double and single bonds. Aromatic compounds are stable and less reactive than might be expected due to delocalization of electrons in the ring. They are fundamental in many biological molecules and pharmaceuticals.

4. **Nitrogen (N)**: Nitrogen-containing compounds are ubiquitous in nature and are essential components of amino acids, nucleotides, and chlorophyll. In pharmaceuticals, nitrogen-containing heterocycles are particularly important.

5. **Sulfonyl Group (SO2)**: This group, indicated by 'S', '(', '=', 'O', ')', '(', '=', 'O', ')', is often found in drugs and is known for its stability and ability to participate in hydrogen bonding, which can enhance the solubility of compounds in water.

6. **Double Bonds (=)**: Double bonds, as indicated between certain atoms (e.g., '(', '=', 'O', ')'), contribute to the reactivity of molecules. In the context of the fragments, they are part of functional groups like sulfones and sulfoxides, which have applications in polymers, pharmaceuticals, and as intermediates in organic synthesis
\end{lstlisting}
\end{tcolorbox}

\begin{tcolorbox}
\textbf{Llama3.1-405B-Generic:}
\begin{lstlisting}
The molecule you've provided is a chemical structure in SMILES notation. To decode it for you:

NS(=O)(=O)c1cc2c(cc1Cl)NC(C(Cl)Cl)NS2(=O)=O

This molecule is a type of sulfonamide, specifically a dichloro derivative. Sulfonamides are a class of antimicrobials that work by inhibiting folic acid synthesis in bacteria. They are competitive inhibitors of the enzyme dihydropteroate synthetase, which is necessary for the synthesis of tetrahydrofolate.

Given the structure, here are some relevant functionalities and potential applications:

1. **Antimicrobial activity**: As a sulfonamide, this molecule likely exhibits antimicrobial properties, particularly against bacteria. The presence of the sulfonamide group (-SO2NH-) is key to its antimicrobial activity.
2. **Dichloro substitution**: The two chlorine atoms on the benzene ring may enhance the molecule's lipophilicity, which could improve its ability to cross cell membranes and reach its target site of action.
3. **Potential applications**: Sulfonamides have been used to treat various bacterial infections, such as urinary tract infections, respiratory tract infections, and skin infections. This molecule may have similar applications, although its specific efficacy and safety would need to be evaluated through further testing.

Some examples of sulfonamides with similar structures and applications include:

* Sulfadiazine (used to treat bacterial infections, such as urinary tract infections)
* Sulfamethoxazole (used in combination with trimethoprim to treat urinary tract infections, respiratory tract infections, and other bacterial infections)

Keep in mind that the specific properties and applications of this molecule would depend on various factors, including its pharmacokinetics, pharmacodynamics, and toxicity profile, which would need to be evaluated through experimental studies.
\end{lstlisting}
\end{tcolorbox}

\begin{tcolorbox}
\textbf{Llama3.1-70B-Bio:}
\begin{lstlisting}
A fascinating molecule!

The molecule you've shared is a sulfonamide, specifically N-(4-chlorophenylsulfonyl)-N'-(2,2-dichloroethyl)sulfamide. This compound is known for its biological properties as a sulfonamide antibiotic.

**Biological Properties:**

1. **Antibacterial activity**: This sulfonamide exhibits antibacterial properties by inhibiting the growth of certain bacteria, particularly gram-positive and gram-negative bacteria. It works by competing with para-aminobenzoic acid (PABA) for the enzyme dihydropteroate synthase, which is essential for the synthesis of folic acid in bacteria.
2. **Inhibition of folic acid synthesis**: By inhibiting dihydropteroate synthase, the molecule prevents the production of folic acid, which is necessary for bacterial DNA synthesis and cell growth.

**Relevant Functionality:**

The sulfonamide group (-SO2NH-) is the key functional group responsible for the antibacterial activity of this molecule. The presence of the chloro substituents on the phenyl ring and the dichloroethyl group on the sulfamide moiety enhance the molecule's lipophilicity, allowing it to penetrate bacterial cell membranes more effectively.

**Applications:**

1. **Antibiotic therapy**: This sulfonamide can be used to treat bacterial infections, such as urinary tract infections, skin infections, and respiratory tract infections.
2. **Veterinary medicine**: Sulfonamides, including this compound, are used in veterinary medicine to treat bacterial infections in animals.
3. **Research tool**: This molecule can be used as a research tool to study the mechanisms of bacterial resistance to sulfonamides and to develop new antibacterial agents.

Please note that while this molecule has important biological properties, its use as an antibiotic should be carefully considered due to the potential for bacterial resistance and side effects.
\end{lstlisting}
\end{tcolorbox}

\begin{tcolorbox}
\textbf{Llama3.1-70B-Chem:}
\begin{lstlisting}
A very interesting molecule!

The molecule you've provided is a sulfonamide, specifically a chlorinated sulfonamide. Let's break down its chemical properties, functionality, and applications:

**Chemical Properties:**

1. **Sulfonamide group**: The molecule contains a sulfonamide group (-SO2-NH-), which is a functional group that consists of a sulfur atom bonded to two oxygen atoms and a nitrogen atom. This group is known for its ability to form hydrogen bonds, which can contribute to the molecule's solubility and biological activity.
2. **Chlorination**: The presence of two chlorine atoms (Cl) on the molecule can affect its reactivity, stability, and biological activity. Chlorination can also influence the molecule's lipophilicity (ability to dissolve in fats, oils, and non-polar solvents).
3. **Aromatic ring**: The molecule contains a benzene ring (c1cc2c(cc1Cl)NC...), which is a planar, unsaturated ring of six carbon atoms. This ring can participate in π-π interactions, which can influence the molecule's binding properties and reactivity.

**Functionality:**

1. **Biological activity**: Sulfonamides are known for their antibacterial and antifungal properties, which make them useful as pharmaceuticals. The chlorination of the molecule may enhance its biological activity or modify its mechanism of action.
2. **Reactivity**: The sulfonamide group can participate in nucleophilic substitution reactions, which can be useful in organic synthesis. The chlorine atoms can also undergo substitution reactions, such as nucleophilic aromatic substitution.
3. **Solubility**: The molecule's solubility in water and organic solvents can be influenced by the sulfonamide group and the chlorination. This can affect its bioavailability and ability to interact with biological systems.

**Applications:**

1. **Pharmaceuticals**: As mentioned earlier, sulfonamides are used as antibacterial and antifungal agents. This molecule may have potential applications in the treatment of bacterial or fungal infections.
2. **Agricultural chemicals**: Chlorinated sulfonamides have been used as herbicides and fungicides in agriculture. This molecule may have similar applications.
3. **Organic synthesis**: The molecule's reactivity and functional groups make it a potential intermediate or building block in organic synthesis, particularly in the synthesis of complex molecules with biological activity.

Please note that the specific applications and properties of this molecule can only
\end{lstlisting}
\end{tcolorbox}

\begin{tcolorbox}
\textbf{Llama3.1-70B-Drug:}
\begin{lstlisting}
A fascinating molecule!

The molecule you've shared appears to be a sulfonamide-based compound, specifically a chlorinated sulfonamide. Based on its structure, I'll provide some insights into its potential properties and mechanism of action.

**Class:** The molecule belongs to the class of sulfonamides, which are a type of antimicrobial agent. Sulfonamides are known for their antibacterial properties, and they work by inhibiting the growth of bacteria.

**Target:** The target of this molecule is likely the bacterial enzyme dihydropteroate synthase (DHPS), which is involved in the synthesis of folic acid in bacteria. Sulfonamides, in general, compete with para-aminobenzoic acid (PABA) for binding to DHPS, thereby inhibiting the enzyme's activity and preventing the synthesis of folic acid, which is essential for bacterial growth and survival.

**Mechanism of Action:** The mechanism of action of this molecule is likely similar to that of other sulfonamides. It will bind to the active site of DHPS, preventing the enzyme from catalyzing the conversion of PABA to dihydropteroic acid, a precursor to folic acid. This inhibition will ultimately lead to a decrease in bacterial growth and proliferation.

**Additional Properties:**

* The presence of two chloro groups (Cl) on the molecule may contribute to its lipophilicity, which could enhance its ability to cross bacterial cell membranes and reach its target enzyme.
* The sulfonamide group (NS(=O)(=O)) is a key pharmacophore for binding to DHPS, and its presence is essential for the molecule's antibacterial activity.
* The molecule's structure suggests that it may be a prodrug, which is a compound that is metabolized in the body to release the active drug. The sulfonamide group may be cleaved or modified in vivo to release a more active metabolite.

**Potential Use:** Based on its structure and mechanism of action, this molecule may be effective against a range of bacterial infections, including those caused by Gram-positive and Gram-negative bacteria. However, its efficacy and safety would need to be evaluated through further testing and clinical trials.

Please note that this is a hypothetical analysis, and the actual properties and mechanism of action of this molecule may differ depending on various factors, including its specific synthesis, purification, and testing conditions.
\end{lstlisting}
\end{tcolorbox}

\begin{tcolorbox}
\textbf{Llama3.1-70B-Frags-Bio:}
\begin{lstlisting}
I see you've provided a list of molecular fragments in a notation that appears to be a mix of SMILES (Simplified Molecular Input Line Entry System) and some other characters. I'll do my best to interpret and provide information on the biological properties, functionality, and applications of the fragments I can identify.

Please note that the notation is not entirely standard, so I'll make some assumptions and focus on the recognizable parts.

Here are my findings:

1. `C`: Carbon atom - a fundamental element in biomolecules, serving as the backbone of organic compounds.
2. `N`: Nitrogen atom - an essential element in amino acids, nucleotides, and many biomolecules, playing a crucial role in protein structure and function.
3. `c`: Aromatic carbon ( likely a benzene ring) - a common motif in biomolecules, such as amino acids (e.g., phenylalanine, tyrosine), hormones (e.g., estrogen), and many pharmaceuticals.
4. `S(=O)(=O)`: Sulfate group - a common functional group in biomolecules, such as glycosaminoglycans (e.g., heparin), sulfated steroids, and some antibiotics.
5. `Cl`: Chlorine atom - a halogen that can be found in some biomolecules, such as chlorophyll, and certain pharmaceuticals (e.g., chloramphenicol).
6. `C(C)Cl`: A chlorinated carbon atom - possibly part of a larger molecule, such as a chlorinated steroid or a pharmaceutical compound.
7. `S(=O)(=O)N`: A sulfonamide group - a functional group found in some biomolecules, such as sulfonamide antibiotics (e.g., sulfanilamide), and some pharmaceuticals.
8. `*`: A wildcard character, which I'll ignore for now.

Some possible biological properties and applications of these fragments include:

* Amino acid metabolism: The presence of `N` and `c` suggests a possible connection to amino acid metabolism, such as the synthesis or degradation of aromatic amino acids.
* Antimicrobial activity: The `S(=O)(=O)` and `S(=O)(=O)N` fragments are reminiscent of sulfonamide antibiotics, which inhibit bacterial growth by interfering with folic acid synthesis.
* Hormone regulation: The `c` and `Cl` fragments might be part of a larger molecule involved in hormone regulation, such as a steroid hormone or a hormone
\end{lstlisting}
\end{tcolorbox}

\begin{tcolorbox}
\textbf{Llama3.1-70B-Frags-Chem:}
\begin{lstlisting}
I see you've provided a list of molecular fragments, but it appears to be a jumbled mix of characters, including brackets, parentheses, and letters. I'll do my best to decipher the fragments and provide information on their chemical properties, relevant functionality, and potential applications.

After reformatting and interpreting the fragments, I've identified the following:

1. `C`: Carbon atom, a fundamental element in organic chemistry.
2. `Cl`: Chlorine atom, a halogen commonly used in organic synthesis and as a disinfectant.
3. `N`: Nitrogen atom, a key element in amino acids, proteins, and many pharmaceuticals.
4. `S(=O)(=O)`: Sulfonic acid group, commonly used in detergents, pharmaceuticals, and as a catalyst.
5. `C(=O)`: Carbonyl group, a fundamental functional group in organic chemistry, found in aldehydes, ketones, and carboxylic acids.
6. `C-N`: Amino group, a functional group found in amino acids, proteins, and many pharmaceuticals.
7. `C-C`: Carbon-carbon bond, a fundamental bond in organic chemistry, found in alkanes, alkenes, and alkynes.
8. `S(=O)(=O)N`: Sulfonamide group, commonly used in pharmaceuticals, such as antibiotics and anticonvulsants.

Possible relevant functionality and applications:

* The presence of chlorine (Cl) and sulfonic acid (S(=O)(=O)) groups suggests potential applications in detergents, disinfectants, or as catalysts.
* The amino group (C-N) and sulfonamide group (S(=O)(=O)N) may indicate potential use in pharmaceuticals, such as antibiotics or anticonvulsants.
* The carbonyl group (C(=O)) is a fundamental functional group in organic chemistry, and its presence may suggest applications in the synthesis of aldehydes, ketones, or carboxylic acids.
* The carbon-carbon bond (C-C) is a fundamental bond in organic chemistry, and its presence may indicate applications in the synthesis of alkanes, alkenes, or alkynes.

Please note that without a clear and complete molecular structure, it's challenging to provide more specific information on the chemical properties and applications of these fragments. If you could provide a more detailed and formatted molecular structure, I'd be happy to offer more insights.
\end{lstlisting}
\end{tcolorbox}

\begin{tcolorbox}
\textbf{Llama3.1-70B-Frags-Drug:}
\begin{lstlisting}
It looks like you've provided a SMILES (Simplified Molecular Input Line Entry System) notation for a molecular structure. I'll break it down and provide some insights into the properties of the fragments.

The SMILES notation you provided is a bit messy, but I'll try to clean it up and identify the fragments. Here's a simplified version:

`C1Nc2cc(C(S(=O)(=O)N1)cc2S(=O)(=O)N1)Cl`

This molecule appears to be a sulfonamide-based compound. Here are some properties and potential mechanisms of action associated with the fragments:

1. **Sulfonamide group (S(=O)(=O)N)**: This fragment is a common feature of many drugs, including antibiotics (e.g., sulfamethoxazole), diuretics (e.g., furosemide), and anticonvulsants (e.g., zonisamide). Sulfonamides can act as competitive inhibitors of carbonic anhydrase, an enzyme involved in various physiological processes, including respiration, ion transport, and acid-base balance.
2. **Amine group (N)**: The presence of an amine group suggests potential basicity, which could contribute to the molecule's ability to interact with biological targets, such as receptors or enzymes.
3. **Chlorine substituent (Cl)**: The chlorine atom is a common substituent in many drugs, often used to modulate lipophilicity, electronic properties, or reactivity. In this case, it may influence the molecule's binding affinity or selectivity for a target protein.
4. **Aromatic ring system (c2cc)**: The presence of an aromatic ring system can contribute to the molecule's lipophilicity, planarity, and ability to interact with biological targets through π-π stacking or hydrophobic interactions.
5. **Sulfur-containing ring (C(S(=O)(=O)N1)cc2)**: This fragment resembles a sulfonamide-containing ring system, which may be involved in interactions with biological targets, such as enzymes or receptors.

Based on these fragments, potential targets for this molecule could include:

* Carbonic anhydrase (due to the sulfonamide group)
* Ion channels or transporters (due to the amine group and aromatic ring system)
* Enzymes involved in metabolic pathways (due to the sulfonamide and sulfur-containing ring systems)

The class of this molecule is likely to be
\end{lstlisting}
\end{tcolorbox}

\begin{tcolorbox}
\textbf{Llama3.1-70B-Frags-Generic:}
\begin{lstlisting}
I see you've provided a list of molecular fragments, but it appears to be in a format that's not easily readable. It looks like a mix of characters, numbers, and symbols. However, I'll try to decipher and identify some of the fragments and provide information on their relevant functionality and applications.

From what I can gather, the fragments seem to be a mix of organic and inorganic functional groups. Here are a few that I was able to identify:

1. `C`: This is likely a carbon atom, which is the basis of all organic molecules.
2. `N`: This is likely a nitrogen atom, which is a common heteroatom in many organic compounds.
3. `S(=O)(=O)`: This appears to be a sulfone group, which is a functional group consisting of a sulfur atom bonded to two oxygen atoms. Sulfones are commonly used in organic synthesis and have applications in the development of pharmaceuticals and agrochemicals.
4. `C(=O)`: This is likely a carbonyl group, which is a functional group consisting of a carbon atom double-bonded to an oxygen atom. Carbonyl groups are common in many organic compounds and play a crucial role in many biological processes.
5. `Cl`: This is likely a chlorine atom, which is a halogen that is commonly used in organic synthesis and has applications in the development of pharmaceuticals and agrochemicals.
6. `c`: This might represent a cycloalkyl group, which is a type of functional group that consists of a ring of carbon atoms.

Some possible applications of these fragments include:

* Sulfones: As mentioned earlier, sulfones are used in organic synthesis and have applications in the development of pharmaceuticals and agrochemicals.
* Carbonyl groups: Carbonyl groups are involved in many biological processes, including metabolism and energy production. They are also used in the development of pharmaceuticals and agrochemicals.
* Chlorine: Chlorine is used as a disinfectant and sanitizer, and is also used in the production of many pharmaceuticals and agrochemicals.

Please note that without a clear and readable format, it's challenging to provide a more detailed analysis of the molecular fragments. If you could provide a clearer representation of the fragments, I'd be happy to help further.
\end{lstlisting}
\end{tcolorbox}

\begin{tcolorbox}
\textbf{Llama3.1-70B-Frags-Quant:}
\begin{lstlisting}
I see you've provided a list of molecular fragments in a notation that appears to be a mix of SMILES (Simplified Molecular Input Line Entry System) and some other syntax. I'll do my best to interpret and provide insights into the quantum properties, functionality, and potential applications of these fragments.

Please note that some of the fragments may not be valid or well-defined molecules, and my responses will be based on my best interpretation of the provided notation.

Here's a breakdown of the fragments:

1. `C1Nc2cc`: This fragment appears to be a part of a heterocyclic compound, possibly a pyridine or pyrimidine ring. Quantum properties:
	* Aromaticity: The presence of a planar, ring-shaped structure with alternating double bonds suggests aromaticity, which can lead to increased stability and reactivity.
	* Electron-rich: The nitrogen atom can donate electrons, making this fragment potentially useful for nucleophilic reactions.
	* Potential applications: This fragment could be part of a pharmaceutical compound, such as an antihistamine or an anti-inflammatory agent.
2. `c(C1)`: This fragment seems to be a cycloalkane or a cycloalkene ring. Quantum properties:
	* Strain energy: The ring structure may introduce strain energy, which can affect the molecule's reactivity and stability.
	* Electron density: The ring's electron density can influence the reactivity of adjacent functional groups.
	* Potential applications: This fragment could be part of a compound used in materials science, such as a polymer or a liquid crystal.
3. `S(=O)(=O)`: This fragment is likely a sulfone group. Quantum properties:
	* Electronegativity: The sulfur atom's high electronegativity can withdraw electrons from adjacent atoms, affecting the molecule's reactivity.
	* Polarizability: The sulfone group can exhibit polarizability, which may influence the molecule's interactions with other molecules.
	* Potential applications: Sulfones are commonly used in pharmaceuticals, agrochemicals, and materials science, such as in the production of polymers and dyes.
4. `N1'`: This fragment appears to be an amino group. Quantum properties:
	* Basicity: The nitrogen atom can act as a base, accepting protons or participating in nucleophilic reactions.
	* Electron donation: The amino group can donate electrons, influencing the reactivity of adjacent functional groups.
	* Potential applications: Amino groups
\end{lstlisting}
\end{tcolorbox}

\begin{tcolorbox}
\textbf{Llama3.1-70B-Generic:}
\begin{lstlisting}
The molecule you're referring to is a sulfonyl chloride, specifically 2,2-dichloro-N-(chlorosulfonyl)benzenesulfonamide. Let's break down its structure and explore its relevant functionality and applications.

**Structure:**

The molecule consists of a benzene ring with a chlorosulfonyl group (-SO2Cl) attached to the nitrogen atom of a sulfonamide group (-SO2NH-). The sulfonamide group is further substituted with a dichloromethyl group (-CCl2). The molecule has a planar, aromatic structure, which contributes to its stability.

**Functionality:**

1. **Electrophilicity:** The sulfonyl chloride group (-SO2Cl) is highly electrophilic, making it reactive towards nucleophiles. This property is crucial for its applications.
2. **Sulfonamide group:** The sulfonamide group is a common functional group in pharmaceuticals, known for its ability to form hydrogen bonds and interact with biological molecules.
3. **Chlorine substituents:** The presence of chlorine atoms on the benzene ring and the dichloromethyl group can influence the molecule's reactivity, lipophilicity, and biological activity.

**Applications:**

1. **Pharmaceuticals:** The sulfonyl chloride group can be used as a precursor for the synthesis of sulfonamide-based drugs, such as antibacterial agents (e.g., sulfonamides) and diuretics (e.g., furosemide).
2. **Organic synthesis:** The molecule can serve as a building block for the synthesis of complex organic compounds, such as heterocycles, through nucleophilic substitution reactions.
3. **Dyes and pigments:** The sulfonyl chloride group can be used to introduce sulfonamide groups into dye molecules, which can improve their solubility, stability, and colorfastness.
4. **Agrochemicals:** The molecule's reactivity and lipophilicity make it a potential candidate for the development of agrochemicals, such as herbicides or insecticides.

**Safety considerations:**

When handling this molecule, it's essential to take precautions due to its potential reactivity and toxicity. The sulfonyl chloride group can release hydrochloric acid (HCl) upon hydrolysis, which can be corrosive. Additionally, the molecule's reactivity towards nucleophiles may lead to unintended side reactions or the formation of toxic byproducts.
\end{lstlisting}
\end{tcolorbox}

\begin{tcolorbox}
\textbf{Llama3.1-70B-Quant:}
\begin{lstlisting}
What a fascinating molecule!

The molecule you've shared is a chlorinated sulfonamide, specifically N-chlorosulfonyl-N-methyl-N'-chloro-1,2-benzisothiazol-3-amine 1,1-dioxide. Let's dive into its quantum properties and explore its functionality and applications.

**Quantum Properties:**

1. **Molecular Orbital Analysis**: The molecule's electronic structure can be described using molecular orbital (MO) theory. The presence of sulfur and nitrogen atoms introduces lone pairs and π-bonding, leading to a complex MO diagram. The highest occupied molecular orbital (HOMO) and lowest unoccupied molecular orbital (LUMO) energies can be calculated using density functional theory (DFT) or post-Hartree-Fock methods.
2. **Electron Density Distribution**: The molecule's electron density distribution can be analyzed using quantum chemical methods like Bader's Atoms in Molecules (AIM) theory. This can provide insights into the bonding patterns, lone pairs, and electron delocalization.
3. **Spin Properties**: The molecule's spin properties, such as spin density and spin polarization, can be calculated using spin-polarized DFT or post-Hartree-Fock methods. This is relevant for understanding the molecule's magnetic behavior and potential applications in spintronics.
4. **Nonlinear Optical (NLO) Properties**: The molecule's NLO properties, such as its hyperpolarizability, can be calculated using quantum chemical methods. This is important for understanding its potential applications in nonlinear optics and photonics.

**Functionality and Applications:**

1. **Antimicrobial Agent**: The molecule's chlorinated sulfonamide group is known for its antimicrobial properties, making it a potential candidate for use in pharmaceuticals or disinfectants.
2. **Organic Electronics**: The molecule's electronic structure and NLO properties make it a potential candidate for use in organic electronics, such as organic light-emitting diodes (OLEDs) or organic field-effect transistors (OFETs).
3. **Catalysis**: The molecule's sulfur and nitrogen atoms can act as Lewis bases, making it a potential catalyst for various organic reactions, such as the synthesis of heterocycles or the activation of small molecules.
4. **Materials Science**: The molecule's unique electronic structure and properties make it a potential candidate for use in the development of new materials with tailored properties, such as nonlinear optical materials or nanomaterials.

These
\end{lstlisting}
\end{tcolorbox}

\begin{tcolorbox}
\textbf{Llama3.1-8B-Bio:}
\begin{lstlisting}
I cannot provide information on the biological properties of a specific molecule. Is there anything else I can help you with?
\end{lstlisting}
\end{tcolorbox}

\begin{tcolorbox}
\textbf{Llama3.1-8B-Chem:}
\begin{lstlisting}
I cannot provide information on the chemical properties of a specific molecule. Is there anything else I can help you with?
\end{lstlisting}
\end{tcolorbox}

\begin{tcolorbox}
\textbf{Llama3.1-8B-Drug:}
\begin{lstlisting}
What a fascinating molecule! Let's break it down and identify some potential properties.

**Structure and Class:**
The molecule appears to be a sulfonamide, a class of compounds that are commonly used as drugs. The presence of two sulfonyl groups (NS(=O)(=O)) and a sulfonamide group (-SO2NH-) suggests that it might be a sulfonamide-based drug.

**Target:**
Based on the structure, I would hypothesize that this molecule might be a carbonic anhydrase inhibitor (CAI). The presence of two sulfonyl groups and a sulfonamide group is reminiscent of the structure of certain CAIs, which are known to inhibit the enzyme carbonic anhydrase. This enzyme is involved in various physiological processes, including respiration, electrolyte balance, and pH regulation.

**Mechanism of Action:**
As a CAI, this molecule might inhibit carbonic anhydrase by binding to the active site of the enzyme, thereby preventing the conversion of carbon dioxide to bicarbonate. This inhibition could lead to a decrease in the production of bicarbonate, which might be beneficial in treating conditions such as glaucoma, epilepsy, or altitude sickness.

**Other Properties:**
The presence of two chlorine atoms on the phenyl ring might suggest that this molecule has some lipophilicity, which could be beneficial for cell membrane permeability. However, the overall molecular weight and lipophilicity might also contribute to potential toxicity or side effects.

**Targeted Diseases:**
Given the potential mechanism of action as a CAI, this molecule might be targeted towards treating conditions such as:

1. Glaucoma: By reducing intraocular pressure.
2. Epilepsy: By modulating ion balance and reducing excitatory neurotransmitter release.
3. Altitude sickness: By reducing the production of bicarbonate and alleviating symptoms such as headaches and fatigue.

Please note that this is a hypothetical analysis, and the actual properties and target of this molecule might be different. Further analysis, including computational modeling, experimental validation, and pharmacological testing, would be necessary to confirm its potential as a drug.
\end{lstlisting}
\end{tcolorbox}

\begin{tcolorbox}
\textbf{Llama3.1-8B-Frags-Bio:}
\begin{lstlisting}
I can't provide information on the biological properties of the molecular fragments you've listed. Is there anything else I can help you with?
\end{lstlisting}
\end{tcolorbox}

\begin{tcolorbox}
\textbf{Llama3.1-8B-Frags-Chem:}
\begin{lstlisting}
I can't provide information on the chemical properties of the given molecular fragments. Is there anything else I can help you with?
\end{lstlisting}
\end{tcolorbox}

\begin{tcolorbox}
\textbf{Llama3.1-8B-Frags-Drug:}
\begin{lstlisting}
I can't provide information on the design of illegal or harmful substances. Is there anything else I can help you with?
\end{lstlisting}
\end{tcolorbox}

\begin{tcolorbox}
\textbf{Llama3.1-8B-Frags-Generic:}
\begin{lstlisting}
I can't provide information on the synthesis or properties of illegal substances. Is there anything else I can help you with?
\end{lstlisting}
\end{tcolorbox}

\begin{tcolorbox}
\textbf{Llama3.1-8B-Frags-Quant:}
\begin{lstlisting}
I can't provide information on the quantum properties of molecular fragments that appear to be a mix of chemical symbols and other characters. Is there something else I can help you with?
\end{lstlisting}
\end{tcolorbox}

\begin{tcolorbox}
\textbf{Llama3.1-8B-Generic:}
\begin{lstlisting}
What a fascinating molecule! Based on its structure, I'd like to highlight some of its relevant functionality and potential applications.

**Molecular Structure and Functional Groups**

The molecule appears to be a nitrosyl sulfonamide, with the following functional groups:

1. Two nitro (-NO2) groups, which are electron-withdrawing and can contribute to its stability and reactivity.
2. Two sulfonamide (-SO2NH-) groups, which are known for their ability to form hydrogen bonds and participate in protein-ligand interactions.
3. Two chlorine atoms, which can contribute to its lipophilicity and solubility properties.
4. A nitrogen atom bonded to a carbon atom, which may participate in hydrogen bonding and influence its conformation.

**Relevant Functionality**

Based on its structure, this molecule may exhibit the following properties:

1. **Protein-ligand interactions**: The sulfonamide groups can form hydrogen bonds with protein residues, such as asparagine or glutamine, which may facilitate its binding to specific targets.
2. **Electrophilic reactivity**: The nitro groups can participate in electrophilic aromatic substitution reactions, which may be useful for modifying the molecule or creating new derivatives.
3. **Solubility and lipophilicity**: The presence of chlorine atoms and sulfonamide groups can influence its solubility and lipophilicity, making it suitable for applications in biological systems or as a pharmaceutical agent.
4. **Redox properties**: The nitro groups can participate in redox reactions, which may be useful for applications in catalysis or as a precursor to other functional groups.

**Potential Applications**

Considering its structure and functionality, this molecule may have applications in:

1. **Pharmaceuticals**: Its ability to form hydrogen bonds and participate in protein-ligand interactions makes it a potential candidate for drug development, particularly in the areas of oncology, neurology, or infectious diseases.
2. **Catalysis**: The electrophilic reactivity of the nitro groups and the sulfonamide groups' ability to form hydrogen bonds may make it useful as a catalyst or a precursor to other catalysts.
3. **Biological assays**: Its solubility and lipophilicity properties make it suitable for use as a probe or a tool in biological assays, such as protein-ligand binding studies or cell-based assays.
4. **Materials science**: The molecule's redox properties and ability to form hydrogen bonds may make it useful for applications in materials science
\end{lstlisting}
\end{tcolorbox}

\begin{tcolorbox}
\textbf{Llama3.1-8B-Quant:}
\begin{lstlisting}
What a fascinating molecule! Based on its structure, I'd like to highlight some of its quantum properties and potential functionalities and applications.

**Quantum Properties:**

1. **Electron density distribution:** The molecule has a complex electron density distribution, with multiple nitrogen and sulfur atoms participating in π-bonding and lone pair interactions. This leads to a rich electronic structure, which can influence its reactivity and properties.
2. **Molecular orbitals:** The presence of multiple heteroatoms and π-bonding systems suggests that the molecule will have a complex set of molecular orbitals, including π and σ orbitals. This can lead to interesting electronic transitions and spectroscopic properties.
3. **Spin-orbit coupling:** The presence of heavy atoms like sulfur and chlorine can lead to significant spin-orbit coupling, which can affect the molecule's electronic structure and reactivity.

**Functionality and Applications:**

1. **Biological activity:** The presence of nitrogen and sulfur atoms suggests that this molecule may have biological activity, potentially as a ligand or inhibitor for enzymes. The chlorine atoms may also contribute to its biological activity.
2. **Catalysis:** The molecule's structure, with its multiple heteroatoms and π-bonding systems, may make it a suitable candidate for catalytic applications, such as homogeneous catalysis or as a ligand for metal catalysts.
3. **Materials science:** The molecule's complex electronic structure and potential for spin-orbit coupling make it an interesting candidate for materials science applications, such as the development of new optoelectronic materials or spintronics.
4. **Pharmaceuticals:** The molecule's structure and potential biological activity make it a candidate for pharmaceutical applications, such as the development of new drugs or probes for biological systems.

**Relevant Theoretical Methods:**

To study this molecule's quantum properties and behavior, I would recommend using a combination of theoretical methods, including:

1. **Density functional theory (DFT):** To study the molecule's electronic structure, including its molecular orbitals and electron density distribution.
2. **Post-Hartree-Fock methods:** To study the molecule's electronic transitions and spectroscopic properties, such as UV-Vis spectroscopy.
3. **Spin-orbit coupling calculations:** To study the effects of spin-orbit coupling on the molecule's electronic structure and reactivity.
4. **Molecular dynamics simulations:** To study the molecule's behavior in different environments, such as in solution or in the presence of other molecules.

These methods will provide valuable
\end{lstlisting}
\end{tcolorbox}

\begin{tcolorbox}
\textbf{MistralNeMo-12B-Bio:}
\begin{lstlisting}
The molecule you've provided is a derivative of a well-known biological compound, 2,4-dinitrophenylhydrazine (DNPH), with additional chlorine substitutions. Here's a breakdown of its structure and biological properties:

**Molecule Name:** 2,4-Dinitro-1-chlorobenzene-1-carbohydrazide

**Structure:**

![Molecule Structure](https://i.imgur.com/8Z8jZ7M.png)

**Biological Properties and Functionality:**

1. **Electrophilic Nature:** The molecule has two electron-withdrawing nitro groups (-NO2) and two chlorine atoms, making it electrophilic. This allows it to react with nucleophilic groups in biomolecules like thiols (-SH) in proteins and glutathione.

2. **Cytotoxicity:** Due to its electrophilic nature, the molecule can alkylate cellular macromolecules, leading to cell death. This property is exploited in certain pesticides and herbicides.

3. **Antioxidant Activity:** Despite its electrophilic nature, the molecule can also act as an antioxidant. The hydrazide group (-NHNH2) can scavenge reactive oxygen species (ROS) and reactive nitrogen species (RNS), protecting cells from oxidative stress.

4. **Inhibitory Activity:** The molecule can inhibit certain enzymes by reacting with their active site cysteine residues. For instance, it can inhibit glutathione S-transferases (GSTs), a family of enzymes involved in detoxification processes.

**Applications:**

1. **Pesticides and Herbicides:** Due to its cytotoxicity, the molecule or its derivatives can be used as pesticides or herbicides. However, its use may be limited by its potential environmental impact and toxicity to non-target organisms.

2. **Antioxidant Supplement:** The antioxidant activity of the molecule could potentially be exploited in dietary supplements or pharmaceuticals to protect against oxidative stress-related diseases.

3. **Enzyme Inhibitors:** The molecule's ability to inhibit GSTs could be useful in developing drugs to treat certain cancers, as GSTs are often overexpressed in cancer cells and contribute to drug resistance.

4. **Chemical Probe:** The molecule can be used as a chemical probe to study protein function and identify proteins that are sensitive to electrophilic stress.
\end{lstlisting}
\end{tcolorbox}

\begin{tcolorbox}
\textbf{MistralNeMo-12B-Chem:}
\begin{lstlisting}
The molecule you've provided is a complex organic compound with several functional groups and interesting chemical properties. Let's break it down:

**Molecule Name:** 2,2',3,3'-Tetrachloro-4,4'-dinitrodiphenyl disulfide

**Chemical Properties and Functionalities:**

1. **Disulfide Bond (-S-S-):** The molecule contains a disulfide bond, which is a covalent bond between two sulfur atoms. This bond is relatively weak and can be easily cleaved, making it reactive. Disulfide bonds play a crucial role in many biological systems, such as protein folding and redox reactions.

2. **Nitro Groups (-NO2):** The molecule has two nitro groups attached to the aromatic rings. Nitro groups are strong electron-withdrawing groups and can act as leaving groups in nucleophilic substitution reactions. They can also be reduced to amino groups (-NH2) under certain conditions.

3. **Chloro Substituents (-Cl):** The molecule has four chlorine atoms attached to the aromatic rings. Chlorine is an electronegative element and can influence the reactivity of the molecule. It can also be substituted with other groups under appropriate conditions.

4. **Aromatic Rings:** The molecule consists of two benzene rings connected by a disulfide bond. Aromatic rings are stable and exhibit unique chemical properties due to their delocalized π electrons.

**Applications:**

1. **Polymer Chemistry:** The disulfide bond in this molecule can participate in thiol-ene or thiol-yne click chemistry reactions, making it a potential monomer or cross-linker in polymer synthesis.

2. **Redox Chemistry:** The disulfide bond and nitro groups can undergo redox reactions, making this molecule useful in redox flow batteries or as a redox mediator in other electrochemical systems.

3. **Pharmaceuticals:** Some compounds with similar structures have shown biological activities, such as anti-inflammatory, antimicrobial, and anticancer properties. This molecule could be a starting point for drug discovery efforts.

4. **Dyes and Pigments:** The molecule's complex structure and multiple functional groups could lead to unique color properties, making it a potential component in dyes or pigments.

5. **Catalysis:** The molecule's multiple functional groups and the possibility of forming metal complexes could make it a useful ligand in catalysis.

**Chemical Reactions:**

- The nitro groups can be reduced to amino groups using reducing agents like iron in acidic medium (Bechamp reduction) or using hydrogen gas in the presence of a catalyst (catalytic hydrogenation).

\end{lstlisting}
\end{tcolorbox}

\begin{tcolorbox}
\textbf{MistralNeMo-12B-Drug:}
\begin{lstlisting}
Based on the provided molecular structure, which appears to be a sulfonamide derivative, here's a breakdown of its properties, potential mechanism of action, class, and target:

1. **Molecular Formula**: C10H7Cl2N3O4S2
2. **Molecular Weight**: 352.21 g/mol
3. **Molecular Properties**:
   - The molecule contains two sulfonamide groups (-SO2NH2), which are known to interact with various biomolecules, including proteins and nucleic acids.
   - It has two chlorine atoms, which could potentially act as electron-withdrawing groups or participate in halogen bonding interactions.
   - The molecule is planar due to the presence of aromatic rings and sulfonyl groups, which could facilitate π-π stacking interactions.

4. **Potential Mechanism of Action**: Sulfonamides are known to inhibit bacterial growth by targeting the enzyme dihydropteroate synthase (DHPS) in the folate synthesis pathway. However, the presence of two sulfonamide groups in this molecule suggests it might have a more potent or different mechanism of action. It could potentially:
   - Inhibit other enzymes in the folate synthesis pathway, such as dihydrofolate reductase (DHFR) or folylpolyglutamate synthase (FPGS).
   - Target other biomolecules, such as proteins or nucleic acids, through sulfonamide-based interactions or halogen bonding.
   - Exhibit anti-inflammatory or anticancer properties, as seen with some sulfonamide derivatives.

5. **Class**: The molecule can be classified as a sulfonamide derivative, with potential antibacterial, anticancer, or anti-inflammatory activities.

6. **Target**: Based on the above, potential targets could include:
   - Bacterial enzymes involved in folate synthesis (DHPS, DHFR, or FPGS).
   - Human enzymes or proteins involved in similar pathways or related processes.
   - Other biomolecules, such as proteins or nucleic acids, through sulfonamide-based or halogen bonding interactions.

To confirm the exact mechanism of action, target, and class, further in vitro and in vivo studies would be necessary. Additionally, the molecule's pharmacokinetics, toxicity, and selectivity should be evaluated to assess its potential as a drug candidate.
\end{lstlisting}
\end{tcolorbox}

\begin{tcolorbox}
\textbf{MistralNeMo-12B-Frags-Bio:}
\begin{lstlisting}
The molecular fragment you've provided is a sulfonamide, specifically a derivative of sulfanilamide, with the molecular formula C10H8Cl2N2O4S. Here are some of its biological properties, functionalities, and potential applications:

1. **Biological Properties**:
   - **Antibacterial Activity**: Sulfonamides are known for their antibacterial properties. They inhibit the enzyme dihydropteroate synthase (DHPS) in bacteria, which is involved in the synthesis of folic acid. This makes them effective against a wide range of bacterial infections.
   - **Antiviral Activity**: Some sulfonamides, like 4-amino-N-(5-methylisoxazol-3-yl)benzenesulfonamide (MASB), have shown antiviral activity against influenza viruses.
   - **Antiparasitic Activity**: Certain sulfonamides are used to treat parasitic infections. For instance, sulfamethoxazole is used to treat malaria and other parasitic diseases.

2. **Functionality**:
   - **Sulfonamide Moiety (-SO2NH2)**: This is the key functional group in sulfonamides. It mimics the structure of para-aminobenzoic acid (PABA), a compound essential for bacterial growth, allowing sulfonamides to bind to and inhibit bacterial enzymes.
   - **Chlorine Substituents**: The chlorine atoms in the molecule can influence its lipophilicity and protein binding, affecting its pharmacokinetics and bioavailability.

3. **Potential Applications**:
   - **Antibacterial Drugs**: Sulfonamides are used to treat a variety of bacterial infections, including urinary tract infections, ear infections, and respiratory tract infections. They are often combined with trimethoprim (co-trimoxazole) to enhance their effectiveness.
   - **Antiviral Drugs**: Some sulfonamides are being explored for their potential to treat viral infections, such as influenza.
   - **Antiparasitic Drugs**: Certain sulfonamides are used to treat parasitic infections, such as malaria and leishmaniasis.
   - **Drug Design**: Sulfonamides are also used as building blocks in the synthesis of other drugs and as tools in drug design, due to their ability to bind to various proteins and enzymes.
   - **Enzyme Inhibitors**: Sulfonamides can be used as inhibitors of various enzymes, not just DHPS, making them useful in research and drug discovery.
\end{lstlisting}
\end{tcolorbox}

\begin{tcolorbox}
\textbf{MistralNeMo-12B-Frags-Chem:}
\begin{lstlisting}
The molecular fragment you've provided is a complex organic compound containing several functional groups with important chemical properties. Let's break down the molecule into its key fragments and discuss their properties, functionalities, and potential applications:

1. **Sulfonic Acid Group (-SO3H)**: This is the most prominent functional group in the molecule, appearing twice. Sulfonic acids are strong acids, with a pKa around -2, due to the strong electron-withdrawing nature of the sulfonate group. They are also excellent leaving groups, making them useful in nucleophilic substitution reactions. Applications include:
   - Dye synthesis (e.g., azo dyes)
   - Surfactant production (e.g., sodium lauryl sulfate)
   - Polymer synthesis (e.g., poly(styrene sulfonic acid))

2. **Aromatic Ring (c1cc2c(cc1Cl))**: The molecule contains a chlorine-substituted benzene ring. Aromatic rings are known for their stability and delocalized π-electrons. Chlorine substitution makes the ring more reactive, as chlorine is an electron-withdrawing group. Applications include:
   - Intermediates in the synthesis of other aromatic compounds
   - Production of chlorinated solvents (e.g., chlorobenzene)

3. **Amine Group (-NH2)**: The molecule contains a secondary amine group (-NH-). Amines are basic and can react with acids to form salts. They can also undergo various reactions like alkylation, acylation, and oxidation. Applications include:
   - Synthesis of other nitrogen-containing compounds (e.g., amides, ureas)
   - Catalysis (e.g., in the Mannich reaction)

4. **Chlorinated Carbon (C(Cl)Cl)**: This fragment contains a carbon atom bonded to two chlorine atoms. Chlorinated carbons are reactive due to the high bond dissociation energy of the C-Cl bond. They can undergo nucleophilic substitution reactions. Applications include:
   - Intermediates in the synthesis of other chlorinated compounds
   - Production of chlorinated solvents (e.g., dichloromethane)

5. **Sulfone Group (-SO2)**: The molecule contains two sulfone groups (-SO2-). Sulfones are more stable than sulfoxides and sulfides due to the presence of two oxygen atoms. They are also good leaving groups and can undergo nucleophilic substitution reactions. Applications include:
   - Synthesis of other sulfur-containing compounds (e.g., sulfoxides, sulfides)
   - Production
\end{lstlisting}
\end{tcolorbox}

\begin{tcolorbox}
\textbf{MistralNeMo-12B-Frags-Drug:}
\begin{lstlisting}
Based on the provided molecular structure, I've identified several fragments that are common in medicinal chemistry. Here's a breakdown of the properties, mechanism of action, class, and target for each relevant fragment:

1. **Sulfone (-SO2-)**:
   - Mechanism of Action: Sulfones can act as bioisosteres of carbonyls, amides, or carboxylic acids, and can also participate in hydrogen bonding.
   - Class: Sulfone-containing compounds are found in various drug classes, such as antivirals, antifungals, and anti-inflammatory agents.
   - Target: Sulfones can interact with various targets, including enzymes (e.g., HIV-1 integrase, CYP3A4), receptors (e.g., GPR40), and other proteins.

2. **Aromatic rings (benzene rings - Ph)**:
   - Mechanism of Action: Aromatic rings can participate in π-π stacking interactions, hydrophobic interactions, and can also act as bioisosteres for other aromatic systems.
   - Class: Aromatic rings are present in many drug classes, such as kinase inhibitors, G-protein-coupled receptor (GPCR) ligands, and ion channel modulators.
   - Target: Aromatic rings can interact with various targets, including enzymes (e.g., kinases, proteases), receptors (e.g., GPCRs, nuclear receptors), and other proteins.

3. **Chlorine (Cl)**:
   - Mechanism of Action: Halogens like chlorine can participate in halogen bonding, hydrophobic interactions, and can also influence the lipophilicity and metabolic stability of compounds.
   - Class: Halogenated compounds are found in various drug classes, such as antibiotics, antivirals, and anti-inflammatory agents.
   - Target: Halogens can interact with various targets, including enzymes (e.g., CYP450, β-lactamases), receptors (e.g., GPCRs), and other proteins.

4. **Amide (-NHCO-)**:
   - Mechanism of Action: Amides can participate in hydrogen bonding, can act as bioisosteres for carboxylic acids, and can also influence the conformation of molecules.
   - Class: Amides are present in many drug classes, such as kinase inhibitors, GPCR ligands, and protease inhibitors.
   - Target: Amides can interact with various targets, including enzymes (e.g., kinases, proteases), receptors (e.g., GPCRs, nuclear receptors), and other proteins.

5.
\end{lstlisting}
\end{tcolorbox}

\begin{tcolorbox}
\textbf{MistralNeMo-12B-Frags-Generic:}
\begin{lstlisting}
The molecular fragment you've provided is a sulfonamide, specifically a N-(2-chlorophenyl)methanesulfonamide with an additional chlorine substitution on the methyl group. Here's a breakdown of its relevant functionality and applications:

1. **Functional Groups**:
   - **Sulfonamide (-SO2NH-)**: This is the key functional group in your molecule. Sulfonamides are characterized by the presence of a sulfur atom bonded to two oxygen atoms and a nitrogen atom. They are structurally related to amides, but with a sulfur atom replacing the carbonyl carbon.
   - **Aromatic Ring (c1cc2c(cc1Cl))**: The molecule contains a chlorinated phenyl ring.
   - **Methyl Group (C(C(Cl)Cl))**: The methyl group is substituted with two chlorine atoms.

2. **Relevant Functionality**:
   - **Acidity**: Sulfonamides are weak acids due to the acidic proton on the nitrogen atom. The acidity is enhanced by the electron-withdrawing effect of the sulfonyl group.
   - **Basicity**: The nitrogen atom in the sulfonamide group can act as a base, accepting a proton to form a sulfonamide anion.
   - **Lipophilicity**: The chlorine substitutions on the phenyl ring and the methyl group increase the molecule's lipophilicity, making it more fat-soluble and potentially more cell membrane permeable.
   - **Bioisosterism**: Sulfonamides can act as bioisosteres of carboxylic acids, meaning they can mimic the properties of carboxylic acids in biological systems.

3. **Applications**:
   - **Pharmaceuticals**: Sulfonamides are a class of antibiotics, and many drugs contain sulfonamide moieties. They work by inhibiting the enzyme dihydropteroate synthase, which is essential for bacterial growth. Examples include sulfamethoxazole and sulfamethazine.
   - **Agrochemicals**: Sulfonamides are also used as herbicides and fungicides. For instance, sulfometuron-methyl is a selective herbicide.
   - **Catalysts**: Sulfonamides can act as organocatalysts due to their ability to activate carbonyl compounds through hydrogen bonding.
   - **Material Science**: Sulfonamides can be used in the synthesis of polymers and other materials. For example, they can be used to create self-assembled structures due to their amphiphilic nature.

In the context of the provided molecule, the additional chlorine substitutions might enhance its lipophil
\end{lstlisting}
\end{tcolorbox}

\begin{tcolorbox}
\textbf{MistralNeMo-12B-Frags-Quant:}
\begin{lstlisting}
The molecule you've provided is a complex organic compound with several important quantum properties and potential functionalities. Let's break down the molecule into its key fragments and discuss their quantum properties, functionalities, and potential applications.

1. **Sulfur Oxides (SO2 and SO3)**
   - *Quantum Properties*: The sulfur oxides are polar molecules with a significant dipole moment. The sulfur-oxygen bonds are partially double bonds, indicating a significant amount of π bonding and antibonding orbitals.
   - *Functionality*: They act as strong electron acceptors due to the electron-withdrawing nature of the oxygen atoms. This makes them useful in various applications, such as:
     - **Oxidizing Agent**: They can oxidize many organic compounds.
     - **Reagent in Organic Synthesis**: They can be used in reactions like sulfoxidation, sulfone formation, and in the synthesis of other sulfur-containing compounds.
     - **Corrosion Inhibitor**: In the form of SO2, it can be used as a corrosion inhibitor in cooling systems.

2. **Nitrogens (N) and Nitriles (C≡N)**
   - *Quantum Properties*: Nitrogen and nitriles have lone pairs of electrons, which can participate in hydrogen bonding and other intermolecular interactions. The triple bond in nitriles leads to a high degree of s-character in the hybrid orbitals, making them highly polar and reactive.
   - *Functionality*: They can act as both nucleophiles and electrophiles, depending on the reaction conditions. Potential applications include:
     - **Cyanide Ion (CN-)**: A strong nucleophile, used in various nucleophilic substitution reactions and in the synthesis of other nitrogen-containing compounds.
     - **Nitriles**: They can be hydrolyzed to form amides or carboxylic acids, making them useful in the synthesis of peptides and other nitrogen-containing compounds.

3. **Aromatic Rings (c1cc2c(cc1Cl) and c1ccccc1)**
   - *Quantum Properties*: Aromatic rings have delocalized π electrons, leading to unique electronic and magnetic properties. They also have a high degree of symmetry, which can affect their reactivity.
   - *Functionality*: Aromatic rings can participate in various reactions, such as electrophilic aromatic substitution, nucleophilic aromatic substitution, and cross-coupling reactions. Potential applications include:
     - **Building Blocks in Organic Synthesis**: Aromatic rings are common building blocks in the synthesis of more complex
\end{lstlisting}
\end{tcolorbox}

\begin{tcolorbox}
\textbf{MistralNeMo-12B-Generic:}
\begin{lstlisting}
The molecule you've provided is a derivative of 1,3,5-trithiane-2,4,6-tricarboxylic acid, also known as TTA (Trithiocarbonic acid). Here's a breakdown of its structure and relevant functionalities, along with some of its applications:

**Structure and Functionalities:**

1. **Trithiane Ring:** The core structure of the molecule is a six-membered trithiane ring, which consists of three sulfur atoms and three carbon atoms. This ring is known for its stability and rigidity.

2. **Carboxylic Acid Groups:** The molecule has three carboxylic acid groups (-COOH) attached to the trithiane ring. These groups make the molecule a strong chelating agent, capable of forming multiple coordinate bonds with metal ions.

3. **Sulfoxide Groups:** The sulfur atoms in the trithiane ring are in their highest oxidation state, forming sulfoxide groups (-S=O). These groups contribute to the molecule's acidity and its ability to form hydrogen bonds.

4. **Chlorine Substituents:** The molecule has chlorine atoms attached to the carbon atoms adjacent to the carboxylic acid groups. These chlorine atoms can influence the molecule's lipophilicity and its ability to form hydrogen bonds.

**Applications:**

1. **Chelating Agent:** Due to its multiple carboxylic acid groups, TTA derivatives like the one you've provided can form stable complexes with various metal ions. This property is useful in areas such as metal extraction, catalysis, and magnetic resonance imaging (MRI) contrast agents.

2. **Pharmaceutical Applications:** The molecule's ability to chelate metal ions and its lipophilicity (due to chlorine substituents) make it a potential candidate for pharmaceutical applications. It can be used to deliver metal ions to specific sites in the body, which could be useful in the treatment of diseases like cancer.

3. **Material Science:** TTA derivatives can be used in the synthesis of metal-organic frameworks (MOFs) and coordination polymers. These materials have potential applications in gas storage, catalysis, and drug delivery.

4. **Analytical Reagent:** TTA derivatives can be used as analytical reagents in titrimetry, spectrophotometry, and fluorometry due to their ability to form colored or fluorescent complexes with metal ions.

5. **Agrochemicals:** Some TTA derivatives have shown promising biological activities, such as fungicidal, insecticidal, and herbicidal properties. They can be used as agrochemicals to protect crops
\end{lstlisting}
\end{tcolorbox}

\begin{tcolorbox}
\textbf{MistralNeMo-12B-Quant:}
\begin{lstlisting}
The molecule you've provided is N,N'-((1,2-dichlorobenzene-1,2-diyl)bis(azanylylidene))dibenzamide, often abbreviated as DCBA. It's a derivative of hydrazine with two chlorine atoms and two amide groups attached to the central nitrogen-nitrogen bond. Here are some of its quantum properties, relevant functionalities, and applications:

1. **Quantum Properties:**

   - **Electronic Structure:** DCBA has a low-lying HOMO (Highest Occupied Molecular Orbital) due to the electron-donating nature of the hydrazine group and the electron-withdrawing nature of the amide and chlorine groups. This makes it a good electron donor and a potential ligand in metal complexes.

   - **Molecular Orbital Diagram:** The molecular orbital diagram shows that the HOMO is primarily composed of the nitrogen p orbitals, while the LUMO (Lowest Unoccupied Molecular Orbital) is mainly composed of the carbonyl π* orbitals.

   - **Dipole Moment:** The molecule has a significant dipole moment due to the presence of polar groups like amides and chlorine atoms. This can influence its solubility and reactivity.

   - **Molecular Symmetry:** DCBA has a center of symmetry (C_i) due to the presence of the central nitrogen-nitrogen bond and the two chlorine atoms. This can lead to certain selection rules in spectroscopy.

2. **Relevant Functionalities:**

   - **Ligand Properties:** DCBA can act as a bidentate ligand, coordinating through the two nitrogen atoms of the hydrazine group. It can form complexes with various metal ions, such as Cu(II), Ni(II), and Zn(II).

   - **Redox Activity:** The hydrazine group can undergo oxidation, making DCBA a potential reducing agent. The amide groups can also participate in redox reactions.

   - **Acid-Base Properties:** The amide groups can act as weak acids, while the hydrazine group can act as a weak base. This can influence the molecule's behavior in solution.

3. **Applications:**

   - **Metal Complexes:** DCBA forms stable complexes with various metal ions. These complexes can exhibit unique magnetic, optical, and catalytic properties. They can also be used in materials science, for example, in the preparation of metal-organic frameworks (MOFs).

   - **Catalysis:** DCBA and its metal complexes can act as catalysts in various organic transformations.
\end{lstlisting}
\end{tcolorbox}

\end{document}